\documentclass{article}

\PassOptionsToPackage{numbers, sort&compress}{natbib}   %
\usepackage[preprint]{neurips}
\makeatletter
\renewcommand{\@noticestring}{Technical report.}
\makeatother
\usepackage[export]{adjustbox}
\usepackage[ruled]{algorithm2e}
\usepackage[inline, shortlabels]{enumitem}
\usepackage[T1]{fontenc}
\usepackage{hyperref}
\usepackage{microtype}
\usepackage{pifont}
\usepackage{xcolor}
\usepackage{xurl}
\usepackage{graphicx}
\usepackage{booktabs}
\usepackage{tabularray}
\usepackage{listings}
\usepackage{amsmath, amsfonts}
\usepackage{nicefrac}

\UseTblrLibrary{booktabs}

\makeatletter
\newcommand{\ssymbol}[1]{\@fnsymbol{#1}}
\newcommand{\romanNumeral}[1]{\expandafter\@slowromancap\romannumeral #1@}
\makeatother

\usepackage{booktabs}
\usepackage{multirow}
\usepackage{array}
\usepackage{makecell}
\usepackage{graphicx}
\usepackage{xcolor}
\usepackage{placeins}
\usepackage{fontspec}          %
\newfontfamily\tamilfont[
  Path         = fonts/,
  Extension    = .ttf,
  UprightFont  = *-Regular,
  BoldFont     = *-Bold,
  Scale        = MatchLowercase
]{NotoSansTamil}
\providecommand{\tamilfont}{}
\ifdefined\XeTeXgenerateactualtext \XeTeXgenerateactualtext=1 \fi
\newcommand{\ta}[1]{{\tamilfont #1}}
\newcommand{\agaram}{\ta{அகரம்}\,/\,Agaram}

\newcommand{\src}[1]{}          %
\newcolumntype{L}[1]{>{\raggedright\arraybackslash}p{#1}}
\usepackage{tikz}
\usetikzlibrary{positioning, arrows.meta, calc, fit, backgrounds}
\definecolor{agIndigo}{HTML}{2B3A8C}   %
\definecolor{agSaffron}{HTML}{E08E0B}
\definecolor{agTeal}{HTML}{0F7C7A}
\definecolor{agGrey}{HTML}{4A4A4A}
\graphicspath{{figures/}}
\hypersetup{colorlinks=true, linkcolor={red!60!black}, citecolor={blue!60!black}, urlcolor={blue!60!black}}
\title{\agaram{} 2.5B: The First Tamil-First Language Model\\Trained from Scratch with Tool Calling\\
\large A playbook for under-represented languages}

\author{
  Ramchand Kumaresan \\
  Murai Labs \\
  \texttt{ramchand@murailabs.com} \\
  \And
  Madhusudhanan G \\
  Independent researcher \\
  \texttt{smkp84@gmail.com} \\
  \And
  Kaushik Kezendirane \\
  Independent researcher \\
  \texttt{0xkaushik.eth@gmail.com} \\
}

\begin{document}
\maketitle

\begin{abstract}
We present \agaram{} 2.5B, a Tamil-first language model trained from scratch that converses in Tamil script,
romanized Tanglish and English, and calls tools. Our first model, \agaram{} 1.5B, guided the new base, tokenizer and
fine-tuning recipe: we pretrained a new 2.53B-parameter base on 50 billion tokens, 58\% of them Tamil, with a tokenizer
that needs 1.55 tokens per Tamil word where Qwen2.5's needs 9.44. A four-day public preview of our first 2.5B chat
model showed where it fell short with real users, and we rebuilt supervised fine-tuning around what we saw, on 222,662
records. The released model scores above Sarvam-1 in all six translation directions we measured (FLORES chrF 56.9
into Tamil and 55.4 into English), chooses the expected tool on all 25 prompts of a tool set that was used during
checkpoint selection and whose schemas appear in training, with exact arguments on 16, and was preferred over the preview model by an LLM judge on 100 sealed human-written prompts (net preference $+30$
points, 95\% interval $+14.5$ to $+45.0$). On FLORES it also scores above Google's Gemma 4 E4B-it and our fine-tuned
variant of it in both directions, with 2.53B parameters against E4B's 4.5B effective parameters. Factual knowledge and
reliable refusal of harmful requests remain open problems, which we set out as future work. The base and chat weights
and GGUF files are released under Apache-2.0, and this report gives a full account of the data, training and
evaluation as a playbook for building language models for under-represented languages, using approximately
USD 13,900 in compute credits and paid services across both models.
\end{abstract}

\section{Introduction}\label{sec:intro}

Tamil has tens of millions of speakers and a literature more than two thousand years old, yet open language models
serve it poorly, both in its own script and in the romanized form that much everyday typing uses. This report describes
\agaram{} 2.5B, a Tamil--English model we trained from scratch to be a conversation partner in Tamil, and how we built
it: the corpus, a tokenizer designed for Tamil, two pretraining runs, a fine-tuning process shaped by a public preview,
and an evaluation designed for a language with few benchmarks. It is, to our knowledge, the first Tamil-first model
trained from scratch that calls tools (Section~\ref{sec:related}). The name \ta{அகரம்} (Agaram) refers to \ta{அ}, the first letter of the Tamil
alphabet, and was chosen by public vote.

\subsection{Tamil script, Tanglish and the intended use}\label{sec:intro-lang}

\paragraph{Script and register.} Tamil is written in its own script, in which a consonant and a following vowel form
one written unit: the vowel is a sign attached to the consonant, and Unicode stores it as a separate combining mark.
A single visible letter can therefore be more than one code point, and a tokenizer that splits at combining marks cuts
words into many small pieces. Written Tamil also spans several \emph{registers}, by which we mean varieties of the
language used for different purposes: classical poetry such as the Sangam anthologies and the Thirukkural, formal
modern prose of newspapers, textbooks and government text, and the colloquial Tamil of speech and messages, which
is far from both. We evaluate the base model separately by register
(Section~\ref{sec:res-register}).

\paragraph{Tanglish.} Many Tamil speakers type Tamil in Latin letters, often mixed with English words, for example
\textit{``Kalvi pathi oru Thirukkural sollu''} (``tell me a Thirukkural couplet about education''). We call this romanized,
code-mixed Tamil \emph{Tanglish}. It has no standard spelling. In the first two days of our public preview, an AI
annotator labelled 45\% of visitors' turns as Tanglish, 37\% as English and 13\% as Tamil script, so a usable Tamil
chat model has to read and answer all three.

\paragraph{Token cost.} A model reads text as tokens, pieces of words from a fixed vocabulary, so tokenization affects
how much Tamil fits within a fixed token budget. On our samples, Qwen2.5's tokenizer uses 9.44 tokens per Tamil word
and 1.35 per English word (Table~\ref{tab:tok}): equal word counts need about seven times as many tokens in Tamil and
fill a token-limited context correspondingly faster. Our tokenizer uses 1.55 tokens per Tamil prose word. These
measurements describe token counts; they do not estimate runtime or financial cost. We trained from scratch to use this
vocabulary from the start and to keep a record of every training document for overlap checks.

\paragraph{Intended use.} \agaram{} 2.5B is built for everyday conversation in Tamil script, Tanglish and English:
greetings, everyday questions, explanations, translation between Tamil and English, and simple tool calls such as a
lookup or a calculation. Translation is its strongest skill (Section~\ref{sec:res-bench}). The 8-bit GGUF file (GGUF
is the model-weight format used by llama.cpp) is 2.69\,GB and runs locally with llama.cpp without a GPU, so students,
developers and researchers can use and adapt a Tamil model without a cloud service. It is also a starting point for
further work: fine-tuning for a particular domain, building Tamil applications, and research on Tamil language
modelling. For applications that depend on factual accuracy, its answers need checking, and it should not be relied on
for medical, legal, financial or crisis advice (Section~\ref{sec:limits}).

\subsection{Model overview and contributions}\label{sec:intro-overview}

\agaram{} 2.5B is a dense decoder-only Transformer with 2.53B parameters and a 64,000-token vocabulary. It was
pretrained from random initialisation on 50B tokens: 43.5B in a stable stage at 4,096 tokens of context and 6.5B in
a decay stage at 8,192, on one node of eight H100 GPUs, in a single run without restarts (Section~\ref{sec:pretrain}).
Tamil is 58\% of the training tokens; English, math, code, parallel text and instruction data make up the rest
(Section~\ref{sec:data}). The released chat model was fine-tuned from this base on 222,662 training records, of which
about a fifth of the supervised tokens are translation, mostly from human-translated sentence pairs
(Section~\ref{sec:sft-data}).

\begin{figure}[t]
\centering
\includegraphics[width=\linewidth]{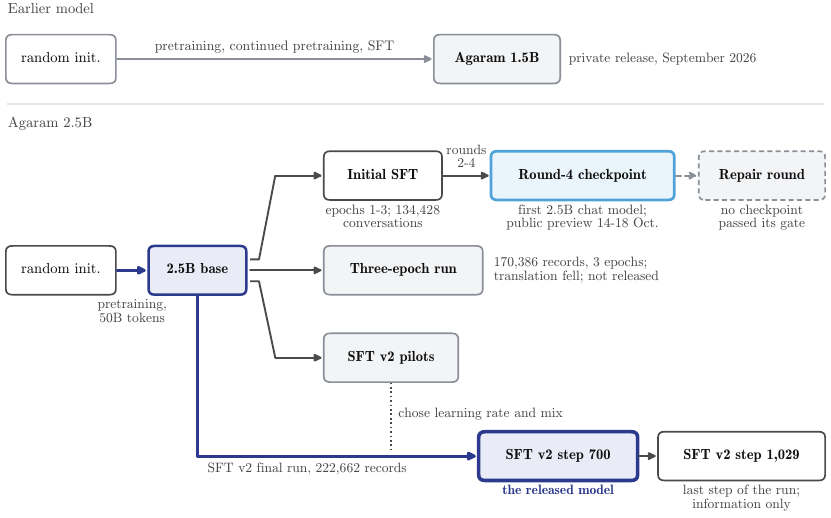}
\caption{How the checkpoints in Table~\ref{tab:ledger} descend from one another. Solid arrows are training runs; the
arrow from step 700 to step 1,029 is the rest of the same run, and the dotted line marks that the pilots chose its
learning rate and mix. Every 2.5B SFT run starts from the 2.5B base except targeted rounds 2--4, which continue from
initial SFT epoch 3, and the repair round (dashed), which continues from the round-4 checkpoint and produced no
checkpoint that met its acceptance criterion.}
\label{fig:history}
\end{figure}

Table~\ref{tab:ledger} identifies the checkpoints and their starting weights. The round-4 checkpoint served the public
preview; SFT v2 step 700 is the model selected for release, and ``the released model'' means this checkpoint.
Figure~\ref{fig:history} shows the separate training branches from the pretrained base.

\begin{table}[t]
\centering
\small
\caption{Checkpoints named in this report. Every 2.5B SFT run starts either from the 2.5B base or from a checkpoint in
an earlier row, as given in the third column.}
\label{tab:ledger}
\setlength{\tabcolsep}{4pt}
\begin{tabular}{L{2.8cm}L{5.0cm}L{2.1cm}L{2.9cm}}
\toprule
Name in this report & What it is & Starts from & Role \\
\midrule
\agaram{} 1.5B & Our first model: 1.56B-parameter dense Transformer, pretrained and then continued to about 66B
tokens, fine-tuned and released privately in September 2026 & random init. & Weak at open chat; motivated the restart
(Section~\ref{sec:motivation}) \\
\addlinespace
2.5B base & Fresh pretraining on 50B tokens, stable and decay stages & random init. & Base of every 2.5B SFT run \\
\addlinespace
Initial SFT, epochs 1--3 & General SFT on 134,428 conversations for three epochs on one H200 & 2.5B base & Epoch 3
started the targeted rounds \\
\addlinespace
Round-4 checkpoint & Initial SFT epoch 3, then targeted rounds 2--4; epoch 1 of round 4 & initial SFT epoch 3 & First
2.5B chat model; public preview, 14--18 October 2026 \\
\addlinespace
Three-epoch run & One SFT run on all SFT data then available (170,386 conversations), three epochs & 2.5B base & Better
chat, lower English-to-Tamil translation; not released \\
\addlinespace
SFT v2 step 700 & Step 700 of the final SFT v2 run (revised mix, seed 19) & 2.5B base & \textbf{The released model} \\
\addlinespace
SFT v2 step 1,029 & Last step of the same run & 2.5B base & For information only \\
\addlinespace
Agaram-Gemma & Low-rank adapter (LoRA) SFT of Gemma 4 E4B-it on our initial SFT and targeted-round data & Gemma 4 E4B-it & Comparison only
(Section~\ref{sec:gemma}) \\
\bottomrule
\end{tabular}
\end{table}

The released model scores higher than the preview model on English-to-Tamil translation and higher than Sarvam-1 in
all six benchmark directions we measured (FLORES chrF 56.9 into Tamil, against 54.3 for the preview model and 49.2 for
Sarvam-1), chooses the expected
tool on all 25 prompts of a development tool set used during checkpoint selection and fills in exact arguments on 16,
and was preferred over the preview model on a sealed
set of human-written prompts (Section~\ref{sec:results}). Its main gap is factual knowledge, which we discuss with the
other limitations in Section~\ref{sec:limits}. The contributions of this report are:
\begin{itemize}
\item \textbf{An open Tamil-first model.} The base and chat weights of a Tamil-first model trained from scratch that
calls tools, with GGUF files that run locally with llama.cpp, released under Apache-2.0
(Section~\ref{sec:release-artifacts}).
\item \textbf{A tokenizer built for Tamil.} 1.55 tokens per Tamil word against 9.44 for Qwen2.5's tokenizer, with
line breaks and indentation kept, so the same Tamil text needs about a sixth of the tokens Qwen2.5 needs (Section~\ref{sec:data-tok}).
\item \textbf{A recipe for Tamil training data.} How we assembled 50 billion training tokens from web corpora,
digitised books and our own OCR of scanned books, and the practice that made the filters trustworthy: judging a filter
by reading what it removes (Sections~\ref{sec:model1} and~\ref{sec:data}).
\item \textbf{Fine-tuning shaped by real users.} A post-training process in which each run answered failures observed
in the previous model or in a public preview, and the findings it produced, including what did not help: more
fine-tuning on a weak base, and training on mixtures that give translation too little weight (Section~\ref{sec:sft}).
\item \textbf{An evaluation design for a language with few benchmarks.} Prompts written by a native speaker after the
training data was fixed, a primary sealed comparison with the finalist frozen before any reply was generated and later
exploratory uses disclosed, repeated judge passes, and overlap scans that are possible because we keep an inventory of
every document the model saw in pretraining (Section~\ref{sec:eval-protocol}).
\item \textbf{A cost record.} The whole project, two models and both pretraining runs, used about USD 13,900
in compute credits and paid services, a reference point for other small teams (Section~\ref{sec:cost}).
\end{itemize}

Sections~\ref{sec:motivation} to~\ref{sec:sft} follow the work in order: the earlier model, the pretraining data and
run, and post-training. Section~\ref{sec:eval-protocol} defines the evaluation, and Sections~\ref{sec:results}
to~\ref{sec:gemma} report the results, a look inside the model and a comparison with a fine-tuned Gemma.
Section~\ref{sec:release} covers cost and release, Section~\ref{sec:limits} the limitations, and
Section~\ref{sec:conclusion} what we learned and the next steps. The appendices hold the inventories, recipes and
intermediate results.

\subsection{Related Tamil and multilingual models}\label{sec:related}

By a \emph{Tamil-first} model we mean one for which Tamil is the majority of its pretraining tokens; \agaram{} 2.5B's
are 58\% Tamil. Most open Tamil models adapt an existing model; Tamil-LLaMA, for example, continues Llama 2 with 16,000
added Tamil tokens \citep{balachandran2023tamilllama}.

\paragraph{Multilingual models trained from scratch.} Sarvam-1 (about 2B parameters, ten Indic languages and English,
about 4 trillion tokens) is a base model \citep{sarvam2024sarvam1}; we compare with it in Section~\ref{sec:results}.
Sarvam 30B and 105B are trained from scratch and support tool calling; their tokenizer covers all 22 scheduled Indian
languages, and their model cards report performance across 22 Indian languages \citep{sarvam2026sarvam30b105b}. Both are
mixture-of-experts models, and the 30B model activates 2.4B parameters per token, close to our 2.5B. BharatGen's Param2,
also a mixture of experts (17B total, 2.4B active), covers English and the 22 scheduled languages and supports tool
calling \citep{bharatgen2026param2}; its predecessor PARAM-1 (2.9B) was trained from scratch on Hindi and English
\citep{pundalik2025param1}. Agaram-Gemma, our LoRA fine-tune of Gemma 4 E4B-it \citep{gemmateam2026gemma4}, is compared
in Section~\ref{sec:gemma}.

\paragraph{Tamil-first models trained from scratch.} Paramanu trains monolingual models for five Indian languages
including Tamil, at 108M to 367M parameters \citep{niyogi2026paramanu}; smaller ones include GPT2-Tamil
\citep{abinayam2022gpt2tamil} and a 23M-parameter model \citep{agdatascientist2026tamilllm23m}. Tamil-800M, released in
July 2026, is a Tamil-only Llama-style model trained from scratch on 4B Tamil tokens with a 32K Tamil tokenizer (653M
parameters in its card's table), with an experimental instruction-tuned version and no tool use described
\citep{skyblanket2026tamil800m}. The closest work to ours
is Balaji's 1111AI-Tamil1B, released on 9 October 2026 \citep{balaji2026tamil1b}: 1.24B parameters (1.12B without
embeddings), pretrained on Tamil, Tanglish and English text (57.6B tokens seen) and instruction-tuned on 11,261
question--answer pairs; its card states that it has no tool use.

\paragraph{Scope of the claim.} In web and model-card searches on 12 and 27 October 2026, we found no other Tamil-first
model trained from scratch that does tool calling, and no larger publicly available Tamil-first model trained from
scratch. To our knowledge as of 27 October 2026, \agaram{} 2.5B is therefore the first Tamil-first model trained from scratch with
tool calling. Its weights are available at the addresses in Section~\ref{sec:release-artifacts}.

\section{Project motivation and earlier model}\label{sec:motivation}

\subsection{Motivation and community participation}\label{sec:motivation-why}

This subsection is the first author's account of why the project exists.

The first books I remember are copies of \ta{அம்புலிமாமா} (Ambulimama) my father brought home, and I still connect them
with the smell of coffee, idli and sambar. My favourite part was \ta{வேதாளம் சொல்லும் கதை}, the Vikram and Vetal
stories. Every story ends with a question the king has to answer, and he cannot stay silent if he knows the answer.
Through stories like these, and the other Tamil books my father kept bringing, I came to love the language and to feel I
owed something to the generations who had kept it alive for more than two thousand years.

When I started working with language models, a question followed me: what would a model learn if Tamil guided it from
the first token, with a tokenizer built for Tamil, data chosen for Tamil, and training from scratch rather than borrowed
from an English model? The question became personal when I thought of my nephews and nieces. They will grow up asking
machines questions, writing with them and learning from them. I want them to be able to do that in Tamil, as naturally
as I read those books. I cannot accept a future in which those tools work well only in English.

So I tried. It started as a personal project: I began collecting data in December 2025 and wrote the first code in May
2026, and it kept growing. My earlier experiments had taught me how to build a language model, but Tamil was a different
problem. Open Tamil datasets are few and small. Older writing is scarcer still: Project Madurai's whole collection of
digitised Tamil texts comes to about 45 million tokens, under a tenth of a percent of what \agaram{} 2.5B was trained
on. The language has also changed: the Tamil of those texts is far from the Tamil people speak and type today, which
mixes in English and is often written in Latin script.

In June 2026 two volunteers, Madhusudhanan G and Kaushik Kezendirane, joined, and from then on it was our project. I
built in public and posted weekly updates, and many people wrote in to support the work or offer help. That is when I
decided the model's name should come from the community: in a poll on X, people proposed their favourite names in 226
replies and voted, and the majority chose \ta{அகரம்} (Agaram). In October we opened the first chat model to the public
for four days. People greeted it with \ta{வணக்கம்}, asked it about Thiruvalluvar and the Sangam poems, and talked to it in
Tanglish. It often failed them, and the model described here is our answer to what they found. The books my father gave me
made Tamil part of my life. I hope this work helps Tamil remain part of the tools the next generation grows up with.

The volunteers' work runs through both models. They tested and graded \agaram{} 1.5B, and one of their tests traced
its run-on free chat to training documents packed without separators. One volunteer built a validator for SFT data
that found mixed-script rows and decomposed Tamil vowel signs our own checks had missed; for \agaram{} 2.5B a volunteer
wrote the distress probe used to test crisis replies. The public preview stored 534 turns from visitors who ticked a
consent box (Section~\ref{sec:sft-preview}). The Contributions section gives each person's part.

\subsection{Agaram 1.5B: the first model and what it taught us}\label{sec:model1}\label{sec:road-model1}

\begin{figure}[t]
\centering
\includegraphics[width=0.85\linewidth]{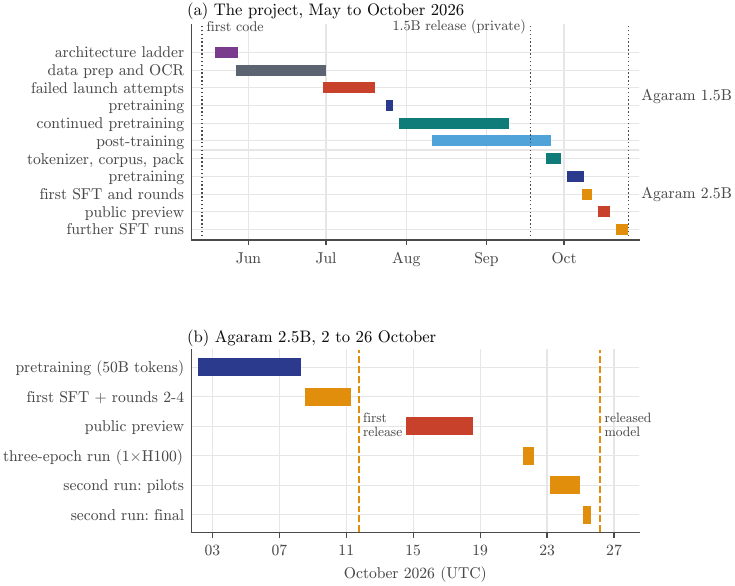}
\caption{(a) The project from its first code to the released chat model (26 October); data collection began earlier, in
December 2025. (b) The 2.5B work in detail. Bars show elapsed windows: the first SFT bar combines general SFT and the
targeted rounds, and the pilot bar combines separate rentals. Later runs span machine launch to termination, including
setup, on-node evaluation and copying. The preview bar shows when the chat demo accepted messages.}
\label{fig:timeline}
\end{figure}

\agaram{} 1.5B was built between May and September 2026 (Figure~\ref{fig:timeline}). Most of the project's human effort
went into its data, and the lessons from that work shaped the second corpus (Section~\ref{sec:data}).
Appendix~\ref{app:corpus-first} gives the fuller account of this corpus with all of its counts.

\paragraph{Architecture comparison.} We first planned a hybrid of Mamba and attention layers, because Mamba's cost
grows linearly with sequence length, whereas attention's cost grows quadratically. Before training anything we set a rule: a
hybrid would replace the standard Transformer only if it matched or beat it on held-out Tamil text and was also more
efficient on long contexts. On a ladder of models of about 91M parameters, three seeds each, a Mamba-2 hybrid matched
the dense Transformer at 25,000 steps but was clearly behind at 75,000 (held-out loss $7.278 \pm 0.078$ against
$7.003 \pm 0.055$); a Mamba-3 hybrid was unstable across seeds, and a Gated DeltaNet hybrid trailed. Both \agaram{}
models therefore use the dense Transformer block. The ladder stopped at about 100M parameters, so it shows that these
hybrids lost under these conditions, not that hybrids cannot win at larger scale (Appendix~\ref{app:model1}).

\paragraph{Collecting and sorting the first corpus.} The first sources were public: Project Madurai's e-texts and the
Tamil parts of web corpora such as CC100, FineWeb-2, HPLT and Sangraha. A search of 178 Hugging Face datasets and 136
Kaggle datasets found no large, clean Tamil prose source to add, so the rest of the book text came from folders of
Tamil PDFs that the first author downloaded from the TamilBooksPDF site and the Tamil Digital Library, and from scanned
books on archive.org. Not everything collected was Tamil text: one 225-page teaching book for spoken Tamil yielded
320,902 characters and not one Tamil character, because its Tamil was romanized or in a legacy encoding. A script
therefore sorted every PDF by what could be extracted from it. In the second Tamil Digital Library pull, 5,006 of
8,053 PDFs had only page images or an empty text layer (Table~\ref{tab:books1}).

\paragraph{OCR.} Much of the printed Tamil material available to us was accessible only as scans, so OCR was the only
way to bring most of these books into the corpus. We kept Tesseract's fast Tamil model for speed, and because one worker would have needed weeks for
thousands of books, OCR ran as four parallel workers under a supervisor that could stop and resume and recorded books
that produced no text so they were not retried forever. The decision to OCR the Tamil Digital Library before the full
pretraining run came from a measurement: on a pilot model, OCR-like book text was the source family with the highest
held-out loss. By mid-June the workers had turned 4,438 PDFs, 625,048 pages, into text.

\paragraph{Filter validation.} OCR output went through page-level filters, and several of their defects were found
only by reading what they removed. When archive.org's own OCR text was filtered for continued pretraining, a
Tamil-share threshold dropped 1,462 works, most of them picture-heavy scans of illustrated books whose page furniture
lowers the Tamil share; 506 held more than 50,000 Tamil characters, and we stopped using the threshold for scanned
books. Separately, an early labelling script put nearly all of the large Tamil corpora into one class called ``web
general''; the first author pointed out that corpora such as OSCAR, HPLT and Sangraha hold mostly modern formal prose,
and about 35.0B tokens were relabelled.

In August, the first author's review of question-answering SFT data for \agaram{} 1.5B flagged garbled OCR in its
source passages, including a passage collapsed onto a single repeated letter (236 of its 242 consonants were
\ta{ப}) that an LLM judge had accepted with a score of 0.97, because the answer faithfully copied the only readable
string. A lexicon-based filter did not separate damaged from clean text, because OCR often turns one real word into
another, such as \ta{நாடியும்} into \ta{தாடியும்}.

The first author and a volunteer then marked 194 of the filter's decisions in a review sheet. Of 41 passages
rejected by one rule, 37 were usable Tamil. That rule caused about half of all rejections; it had looked safe because
we had measured only how rarely it fired on clean text, not how often its rejections were wrong. In the opposite direction, 37 of 60 passages added from archive.org scans were not usable.
From then on a filter's calibration had to report how many of its rejections were wrong, measured by reading them, and
evaluation passages and verbatim-span SFT rows came from typed text, not scans.

\paragraph{What the corpus still lacked.} Exact and near-duplicate removal across the web corpora followed
(Appendix~\ref{app:corpus-first}), but neither the OCR work nor the clean-up changed the shape of the corpus. When the
pool was first assembled, 98\% of its 37.4B tokens were web text derived from Common Crawl. The kinds of Tamil a
conversational and literary model needs most were the scarcest: 0.48B unique tokens of modern literary books, 57M of
classical and cultural text and 0.19B of colloquial dialogue, which pretraining repeated 3.7, 5.3 and 2.0 times. Of the
30B training tokens, 75\% were Tamil web text, 13\% English, and 0.7\% instruction-shaped text.

\paragraph{Pretraining and post-training.} Getting a run to finish took longer than designing it: attempts on rented
8$\times$H200 nodes failed for about three weeks on communication timeouts between GPUs and on resuming from
checkpoints. The production run started in late July on Lambda (1.56B parameters, 30B tokens on 8$\times$H100); its
node shut itself down before the final checkpoint was copied off, so the last 4,221 steps had to be redone. Four rounds
of continued pretraining brought the total to about 66B tokens. Two months of post-training followed
(Table~\ref{tab:post1} in Appendix~\ref{app:model1}). Most training approaches improved their target metric but did not
improve open conversation; adding general public instruction corpora made the model copy passages from its input, and a
Qwen2.5-1.5B control trained on the same data copied too, which pointed to the data rather than to our model. We
released \agaram{} 1.5B privately to the volunteers on 18 September 2026; it was never made public.

\subsection{Why we pretrained a new base}\label{sec:restart}

\paragraph{Staged conversational SFT on the earlier base.} Before funding fresh pretraining, we tested whether better
conversational data could repair the existing 1.5B model. An open-weight teacher generated native Tamil and Tanglish
conversations, which the judge rated mostly usable. Increasing the SFT data from roughly 5K to 15K conversations still
left the best checkpoint usable on 8 of 100 evaluation prompts (Table~\ref{tab:staged} in
Appendix~\ref{app:model1}). We therefore chose fresh pretraining instead of the planned 50K stage: better conversation data alone did not repair
the earlier base, although the test does not separate base capacity from every other explanation.

\paragraph{Changes adopted for Agaram 2.5B.} Each problem found in \agaram{} 1.5B led to a choice for \agaram{} 2.5B
(Table~\ref{tab:lessons} in Appendix~\ref{app:model1} pairs them): a tokenizer that keeps whitespace, document
separators and masking, a measured mix with more English, math and code, OCR filters checked by reading their
rejections, checkpoints copied off the node during the run, overlap scans before numbers are reported, and judge scores
averaged over passes. The tokenizer defect is the clearest: its normaliser erased every line break, so the model never
saw a newline and could not represent lists, verse or indented code. It is one of several changes made together, so the
report does not attribute any later result to it alone.

\section{Pretraining data and tokenization}\label{sec:data}

\begin{figure}[b]
\centering
\includegraphics[width=\linewidth]{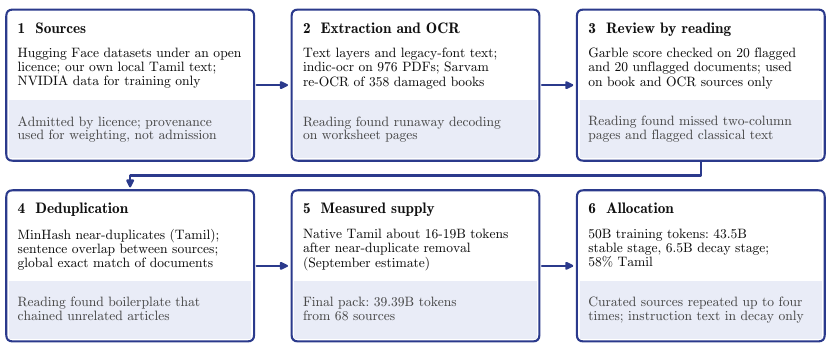}
\caption{Stages of corpus construction. Each box names what was done to the text; the shaded line gives what reading
the output found or what the stage produced. Measured supply, pack tokens and training exposure are different
quantities (Appendix~\ref{app:corpus-quantities}), so the figure is not a funnel, and the earlier model's OCR counts
(Section~\ref{sec:road-model1}) are a different population and are not shown.}
\label{fig:corpus}
\end{figure}

Building the pretraining corpus took more of the project's time than training the model. The work had five parts:
finding sources and deciding which could be used; turning scanned and badly encoded books into text; removing the text
that several collections share; measuring what was left; and dividing the 50 billion training tokens between the two
training stages (Figure~\ref{fig:corpus}). Three token quantities recur and are not interchangeable: the \emph{measured supply} of a source after
cleaning and near-duplicate removal, the \emph{pack tokens} written to the training pack after a final exact
deduplication (39.39B, which we also call the pack's unique tokens), and the \emph{training exposure} of 50B tokens,
repeats included (Appendix~\ref{app:corpus-quantities}). Appendix~\ref{app:corpus} lists every source, its token counts,
the exact processing rules and the full account of each step below.

\subsection{Source collection and classification}\label{sec:data-sources}

The first decision was which text could be used at all. Earlier in the project each dataset had passed through a
per-source review of how it had been produced. For the new corpus we replaced those reviews with one rule: a dataset is
admitted if it is published on Hugging Face under an open licence, and how the publisher obtained or generated the text
is not a criterion. Six NVIDIA datasets distributed under a data agreement were admitted for training only and are not
redistributed, and benchmark test sets are excluded (Section~\ref{sec:contam}). The rule left out much of the Tamil we
had collected ourselves, such as blogs, Project Madurai's e-texts \citep{projectmadurai2024etexts} and the text of
scanned books; after measuring it we admitted its readable part, about 0.9B tokens.

Provenance mattered for weighting rather than admission. On the first model, general instruction data translated into
Tamil had made the model worse, so machine translation was treated as a quality attribute: translated Tamil was capped
at about 1.5B training tokens and machine-translated romanized Tamil at 0.8B, out of 8.8B and 6.9B available. One
consequence is that most of the Tanglish the model saw was machine-produced. Of about 1.35B romanized-Tamil training
tokens, human-written Tanglish accounts for roughly 0.02B; the rest is machine translation and teacher-written
conversations. Reading samples also removed one admitted source, the Thamizhmalai corpus, whose numbers and English
words had been replaced by placeholder tokens (Appendix~\ref{app:corpus-steps}).

\subsection{Text extraction and OCR review}\label{sec:data-corpus}

Much of the Tamil book text available to us was scanned or stored in legacy font encodings. Besides the first model's
Tesseract layer, we added two OCR passes and read the output of each before writing any filter. The open indic-ocr
model ran on our two DGX Spark machines over 976 PDFs (142,271 pages) of state textbooks and library books; on worksheet
pages it decoded long runaway sequences, so that 12\% of the pages took about 70\% of the compute time, and we capped
generation and truncated flagged pages where repetitive output began. Sarvam's OCR service then re-read 358 damaged Tamil Digital Library
books (59,841 pages), chosen because on damaged scans its mean garble score was 0.23 where the text already in our
corpus scored 0.91.

The first model had taught us that a filter's cost is the good text it removes (Section~\ref{sec:road-model1}), so a
garble score built for the new corpus was checked by reading 20 flagged and 20 unflagged documents. The reading showed
errors in both directions. Of the 20 flagged documents, 12 were garbled, but three of ten unflagged OCR documents were
two-column pages whose lines had been interleaved, which the score cannot detect, and the score flagged 29\% of Project
Madurai's classical texts, which read fine: old vocabulary looks like damage to a vocabulary built from modern web text.
We therefore used the score only on book and OCR sources, at the stricter threshold, and not on the classical
e-texts. None of these rules detects an OCR error that turns one valid Tamil word into another. After all of this, OCR
output is about 2.0\% of stable-stage tokens and 0.9\% of decay-stage tokens.

\subsection{Cross-source deduplication and measured supply}\label{sec:data-dedup}

Most public Tamil web datasets are cut from the same few web crawls, so adding them up overstates how much Tamil text
exists. Before near-duplicate removal the Tamil sources held an estimated 65.5B tokens, and an earlier estimate based
on exact matching had put native Tamil at about 45B. MinHash near-duplicate detection over the 87.7M Tamil documents
(Unicode NFC \citep{unicode2026uax15}, newlines kept) left 61.3M. Reading the largest duplicate groups exposed a problem:
groups are formed transitively, and boilerplate shared by many news sites had chained unrelated articles into one group
of 259,644 OSCAR documents. In large groups we therefore dropped a document only when a better-ranked document shared
one of its MinHash bands directly, that is, a matching slice of the compact similarity signature that MinHash computes
for each document.

After this step, admitted native Tamil came to 18.6B tokens, or about 15.9B after discounting sentences already present
in other sources: roughly 16--19B, not 45B. Forty-four percent of CulturaX's characters, for example, were in groups
that also contained OSCAR documents (Appendix~\ref{app:corpus-overlap}). That measurement changed the plan. A 50B-token
run with about 60\% Tamil needs roughly 29B Tamil tokens, so the Tamil sources would have to be repeated, and how often
each could be repeated became a design question for the mixture (Section~\ref{sec:data-decay}). Further sentence-level
and exact passes followed, and the final pack holds 13.9B tokens of native Tamil (Table~\ref{tab:dedup-steps}).

Four review rounds of the built pack by an independent coding agent, reading the packed documents rather than the
build logs, each found defects that the build checks had passed, among them destroyed Python indentation and SFT
conversations in the instruction pool (Table~\ref{tab:pack-review}). The final pack holds 39.39B tokens in 69.17M
documents from 68 sources (Tables~\ref{tab:inventory-ta} and~\ref{tab:inventory-other}).

\subsection{Tokenizer design and validation}\label{sec:data-tok}

The first model's tokenizer, a 64,000-token SentencePiece BPE \citep{kudo2018sentencepiece}, applied a normaliser that
collapsed every run of whitespace into one space, so newlines and indentation never reached the model. We set four
requirements for a replacement: an exact round trip for every input, line breaks included; Tamil cost no worse than
the old tokenizer; lower cost on Tanglish, code, JSON and chat text; and special ids for chat and tool markers reserved
from the start. Two candidates with 64,000 ids were trained on a 930M-character sample of the new corpus
(Table~\ref{tab:tok-sample}): a byte-level variant of BPE \citep{sennrich2016bpe}, which starts from the 256 byte values and can
encode any input, and a SentencePiece BPE with whitespace preserved. Both round-tripped every test set exactly and cost
the same on Tamil prose (1.553 and 1.557 tokens per word); the byte-level BPE was cheaper on code, Tanglish, JSON and
chat and never needed a byte fallback, so we chose it (Table~\ref{tab:tok}).

\begin{table}[t]
\centering
\small
\caption{Tokens per whitespace-separated word (lower means fewer tokens per word) for the chosen tokenizer, the rejected candidate, the
first model's tokenizer and Qwen2.5's. The old tokenizer failed exact reconstruction of every multi-line code and verse test example
(code 0/320, verse 0/204).}
\label{tab:tok}
\begin{adjustbox}{max width=\linewidth}
\begin{tabular}{lrrrr}
\toprule
Sample & \makecell[r]{Byte-level BPE\\(chosen)} & \makecell[r]{SentencePiece,\\whitespace kept} &
\makecell[r]{Old tokenizer\\(\agaram{} 1.5B)} & Qwen2.5 \\
\midrule
Tamil prose          & 1.55 & 1.56 & 1.55 & 9.44 \\
Tamil verse          & 2.29 & 2.27 & 1.80 & 9.19 \\
English              & 1.41 & 1.42 & 1.67 & 1.35 \\
Tanglish             & 1.95 & 2.04 & 2.95 & 2.44 \\
Code (Stack-Edu)     & 3.54 & 3.86 & 4.54 & 3.01 \\
Tool-call JSON       & 2.52 & 2.56 & 3.46 & 2.35 \\
Plain-text chat      & 2.19 & 2.23 & 3.14 & 2.24 \\
\midrule
Exact round trip     & 100\% & 100\% & \makecell[r]{0\% on\\multi-line text} & -- \\
Tamil boundaries inside a grapheme & 14.3\% & 5.1\% & 4.7\% & 65.8\% \\
\bottomrule
\end{tabular}
\end{adjustbox}
\end{table}

The pre-tokenizer is Qwen2's split pattern with one change: in Qwen2 a word consists of Unicode letters only, and
Tamil vowel signs are combining marks, so every Tamil word is cut at each vowel sign; our pattern also admits combining
marks and the two zero-width joiners. We expect this split rule is one reason Qwen2.5 spends 9.4 tokens per Tamil word
on our samples, but we did not measure its share (Appendix~\ref{app:corpus-tok}).

The new tokenizer is no cheaper than the old one on Tamil prose. What it adds is the exact round trip and lower cost on
the other kinds of text: romanized Tamil costs 1.67 times as many tokens as the same sentence in Tamil script, against
2.71 times with the old tokenizer. It also has two measured costs. Classical verse became more expensive (2.29 against
1.80 tokens per word), and 14.3\% of Tamil token boundaries fall inside a grapheme cluster, against 4.7\% for the old
tokenizer; we did not measure what the grapheme splits cost the model. Retraining with up to 80M characters of added
classical text lowered the verse cost by 3.8\% at most, so the cost is not mainly a matter of sample share, and we kept
the original tokenizer.

\subsection{Stable-stage and decay-stage mixtures}\label{sec:data-decay}

The mixture had to answer three questions: how much of the 50B tokens should be Tamil, how often each source could be
repeated, and what the final, low-learning-rate stage of training should see. Table~\ref{tab:mix} gives the result by
category, and Figure~\ref{fig:mix} in Appendix~\ref{app:corpus-steps} by source.

\begin{table}[t]
\centering
\small
\caption{Training tokens by category (50B in total; Tamil includes machine-translated and romanized Tamil).}
\label{tab:mix}
\begin{tabular}{lrrrL{5.6cm}}
\toprule
Category & Stable & Decay & Share & Main sources \\
\midrule
Tamil        & 26.5B & 2.6B & 58.2\% & FineWeb-2, HPLT, Sangraha, CulturaX, OSCAR, Wikipedia, Project Madurai, books and OCR \\
English      &  8.9B & 1.0B & 19.8\% & FineWeb-Edu, Nemotron-CC (training only), DCLM \\
Math         &  4.2B & 0.8B & 10.0\% & FineMath, Nemotron-CC-Math (training only), OpenWebMath \\
Code         &  3.1B & 0.4B &  7.0\% & Stack-Edu, Nemotron code (training only) \\
Parallel     &  0.8B & 0.2B &  2.0\% & finetranslations, Pralekha, BPCC, small parallel sets \\
Instruction  &  --   & 1.5B &  3.0\% & decay stage only (23\% of its tokens) \\
\bottomrule
\end{tabular}
\end{table}

\paragraph{Tamil share.} We found no published measurement of the primary-language share for a model of this size, so
we measured it on a proxy: three 153M-parameter models trained on 2B tokens with 50\%, 65\% or 80\% Tamil and the rest
English. Each step up in Tamil share lowered Tamil loss, but the gains shrank while the English cost grew: increasing the Tamil
share from 65\% to 80\% reduced Tamil loss by 0.036 but increased English loss by 0.152 (Table~\ref{tab:proxy}). We set monolingual Tamil at 58.2\%
of training tokens, about 62\% with the Tamil instruction and parallel text. The proxy was small and short, so it
informs the mixture at this budget rather than showing a general effect.

\paragraph{Repetition.} With 13.9B tokens of native Tamil and about 29B Tamil training tokens, Tamil sources had to be
seen more than once. Curated sources were repeated up to four times and the Tamil web crawls about 2.2 times, while
OCR output was used about once and machine-translated text less than once. Four passes was the upper limit because
repeating data up to about four times has been reported to cost little in loss, at a fixed compute budget, compared
with fresh data \citep{muennighoff2023datacons}; we did not test these caps at this scale.

\paragraph{The decay stage.} Training ended with a decay stage of 6.5B tokens in which the learning rate fell to a
tenth of its peak. OLMo 2, for example, up-samples its highest-quality sources in a final stage that coincides with the
learning-rate anneal \citep{olmo2024olmo2}, and we also used this stage to change the mix: it drops the broad crawls and
raises the curated and filtered sources, so that FineWeb-2 Tamil goes from 9.2\% to 25.8\% of tokens
(Figure~\ref{fig:blendshift}). Five instruction sources appear only in the decay stage, together 23\% of its tokens,
following MiniCPM's report that SFT-style data in the decay stage improved results after the same fine-tuning
\citep{hu2024minicpm}. They are written as plain text without the chat markers, after a report that chat special tokens
added during mid-training made a base model emit them in its output \citep{olmo2025olmo3}, and the conversation families
chosen for SFT were kept out of them. One of them includes Tamil translations of SmolTalk's constraint-following
conversations, whose instruction templates come from IFEval; Section~\ref{sec:contam} reports what that does to one
benchmark.

\section{Model architecture and pretraining}\label{sec:pretrain}

\agaram{} 2.5B was pretrained once, in two stages, on one rented node of eight H100 GPUs. The budget allowed a single
run, so new components were tested on small models before launch, and the run was built to survive the loss of the
node. The full recipe is in Appendix~\ref{app:pretrain}.

\subsection{Architecture}\label{sec:pretrain-model}

\paragraph{Block type and size.} The model uses the dense Transformer decoder block of the first model, because no
hybrid met the rule set for \agaram{} 1.5B (Section~\ref{sec:road-model1}); the choice rests on that small-scale
evidence. The size followed from the budget. About USD 4,000 of the project's USD 5,000 was planned for pretraining,
and projected from the first model's throughput on the same hardware, a 2.5B-parameter model trained on 50B tokens came
to about USD 3,900. That is roughly $7\times10^{20}$ training FLOPs, which at about 20 training tokens per parameter,
the ratio implied by the compute-optimal estimates in Table 3 of \citet{hoffmann2022chinchilla}, gives about 2.4B
parameters on about 50B tokens. A
throughput test of the 2.5B shape was planned but did not run (Appendix~\ref{app:pretrain-ops}), so the cost was known
only once training had started.

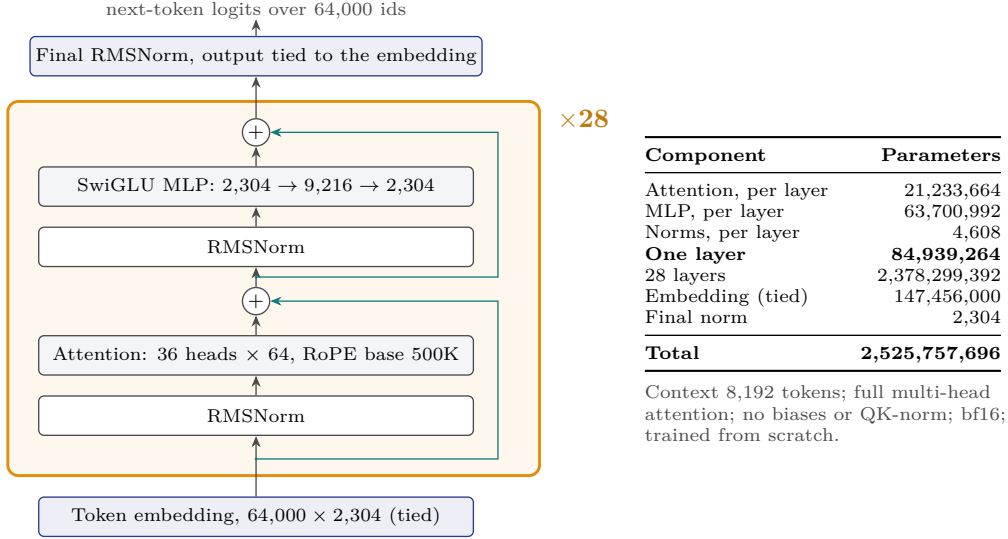
\begin{figure}[t]
\centering
\begin{tikzpicture}[
  font=\scriptsize,
  blk/.style={rectangle, rounded corners=2pt, draw=agGrey, line width=0.5pt, fill=white,
              minimum width=5.75cm, minimum height=5.2mm, align=center, inner sep=2pt},
  sum/.style={circle, draw=agGrey, line width=0.5pt, fill=white, inner sep=0pt, minimum size=3.6mm},
  flow/.style={-{Stealth[length=1.5mm, width=1.2mm]}, line width=0.5pt, draw=agGrey},
  res/.style={-{Stealth[length=1.5mm, width=1.2mm]}, line width=0.5pt, draw=agTeal},
  node distance=2.6mm,
]
\node[blk, fill=agIndigo!8, draw=agIndigo] (emb) {Token embedding, $64{,}000 \times 2{,}304$ (tied)};
\node[blk, above=8mm of emb] (n1) {RMSNorm};
\node[blk, above=of n1, fill=agIndigo!5] (att) {Attention: 36 heads $\times$ 64, RoPE base 500K};
\node[sum, above=of att] (a1) {$+$};
\node[blk, above=of a1] (n2) {RMSNorm};
\node[blk, above=of n2, fill=agIndigo!5] (mlp) {SwiGLU MLP: $2{,}304 \rightarrow 9{,}216 \rightarrow 2{,}304$};
\node[sum, above=of mlp] (a2) {$+$};
\node[blk, above=5.5mm of a2, fill=agIndigo!8, draw=agIndigo] (out) {Final RMSNorm, output tied to the embedding};
\node[above=2.2mm of out, inner sep=0pt, text=agGrey] (logits) {next-token logits over 64,000 ids};
\draw[flow] (emb) -- (n1);
\draw[flow] (n1) -- (att);
\draw[flow] (att) -- (a1);
\draw[flow] (a1) -- (n2);
\draw[flow] (n2) -- (mlp);
\draw[flow] (mlp) -- (a2);
\draw[flow] (a2) -- (out);
\draw[flow] (out) -- (logits);
\coordinate (r0) at ($(emb.north)!0.62!(n1.south)$);
\coordinate (r1) at ($(a1.north)!0.5!(n2.south)$);
\draw[res] (r0) -- ++(3.2cm,0) |- (a1.east);
\draw[res] (r1) -- ++(3.2cm,0) |- (a2.east);
\fill[agTeal] (r0) circle (0.5pt) (r1) circle (0.5pt);
\coordinate (rr) at ($(r0)+(3.35cm,0)$);
\begin{scope}[on background layer]
  \node[draw=agSaffron, line width=1.1pt, rounded corners=4pt, fill=agSaffron!7, inner xsep=4mm, inner ysep=2.2mm,
        fit=(n1) (a2) (rr)] (block) {};
\end{scope}
\node[anchor=north west, text=agSaffron!85!black, font=\normalsize\bfseries] at ($(block.north east)+(1mm,0)$) {$\times$28};
\node[anchor=west, font=\scriptsize, align=left] at ($(block.east)+(1.25cm,0)$) {%
  \begin{tabular}{@{}lr@{}}
  \toprule
  \textbf{Component} & \textbf{Parameters} \\
  \midrule
  Attention, per layer & 21,233,664 \\
  MLP, per layer & 63,700,992 \\
  Norms, per layer & 4,608 \\
  \textbf{One layer} & \textbf{84,939,264} \\
  28 layers & 2,378,299,392 \\
  Embedding (tied) & 147,456,000 \\
  Final norm & 2,304 \\
  \midrule
  \textbf{Total} & \textbf{2,525,757,696} \\
  \bottomrule
  \end{tabular}\\[4pt]
  \parbox{5.2cm}{\raggedright\color{agGrey}Context 8,192 tokens; full multi-head attention; no biases or QK-norm;
  bf16; trained from scratch.}};
\end{tikzpicture}
\caption{The decoder block, repeated 28 times (pre-norm; teal lines are the residual connections), and the parameter
count read from the released weights' tensor shapes.}
\label{fig:arch}
\end{figure}

\paragraph{Configuration.} The model has 2,525,757,696 parameters: 28 layers, hidden size 2,304, 36 attention heads of
dimension 64, a SwiGLU feed-forward layer of width 9,216 \citep{shazeer2020glu}, RMSNorm \citep{zhang2019rmsnorm},
rotary position embeddings \citep{su2021roformer} and input and output embeddings tied over the 64,000-token vocabulary
(Figure~\ref{fig:arch}). The rotary base was set to 500,000 from the first step so that extending the context from
4,096 to 8,192 tokens in the decay stage needed no change of base.

\paragraph{Document masking.} Training samples pack several documents end to end, and with ordinary causal attention
each document is predicted partly from the unrelated one before it. The first model's continued pretraining packed
documents with no separator, and its free chat ran on into memorised instruction text; we did not isolate how much the
packing contributed, but removed the possibility. Every document ends with an end-of-text token, and attention and
rotary positions are restricted to each document. Before launch, the logits of a document scored alone and after another
document in the same sample differed by 0.00000, against 1.89 with plain causal attention.

\subsection{Training schedule and optimisation}\label{sec:pretrain-schedule}

We used a warmup--stable--decay schedule \citep{hu2024minicpm}: the learning rate rises over a short warmup, stays
constant for most of training and falls only at the end. Unlike a cosine schedule, it does not fix the length of the
run in advance, since any stable-stage checkpoint can be finished with a short decay, and the decay stage gave a natural
point to change the data mix (Section~\ref{sec:data-decay}). The split was 43.5B tokens of stable training and 6.5B of
decay. Because only 20.0\% of stable-stage tokens lie in documents longer than 4,096 tokens, and training at 8,192
throughout would have cost an estimated 9--17\% more compute, we trained at 4,096 tokens in the stable stage and 8,192
in the decay stage, which cost about 1--2\%. Training used Adam in bf16 mixed precision (16-bit floating-point arithmetic) on NeMo 2 and Megatron-Core
\citep{kuchaiev2019nemo,shoeybi2019megatron}; Table~\ref{tab:schedule} summarises the two stages.

\begin{table}[t]
\centering
\small
\caption{The two pretraining stages on one node of 8$\times$H100. Wall time is from the node's own log; throughput is
the median over steps. The validation losses of the two stages are on different held-out sets and are not comparable.}
\label{tab:schedule}
\begin{tabular}{lrr}
\toprule
 & \multicolumn{1}{c}{Stable} & \multicolumn{1}{c}{Decay} \\
\midrule
Context (tokens)              & 4,096 & 8,192 \\
Sequences per step            & 128 & 64 \\
Steps                         & 82,970 & 12,398 \\
Tokens                        & 43.50B & 6.50B \\
Learning rate                 & \makecell[r]{warmup 2,000 steps to $3\times10^{-4}$,\\then constant} & \makecell[r]{$3\times10^{-4}$ to $3\times10^{-5}$,\\linear} \\
Datasets                      & 63 & 27 \\
Wall time                     & 130.1 h & 17.6 h \\
Median throughput (tokens/s)  & 94,700 & 107,500 \\
Validation loss, first / last & 4.041 (step 1,000) / 2.441 & 2.080 (step 1,000) / 1.903 \\
Restarts                      & 0 & 0 \\
\bottomrule
\end{tabular}
\end{table}

\subsection{Training behaviour and checkpoint preservation}\label{sec:pretrain-ops}

\begin{figure}[t]
\centering
\includegraphics[width=0.85\linewidth]{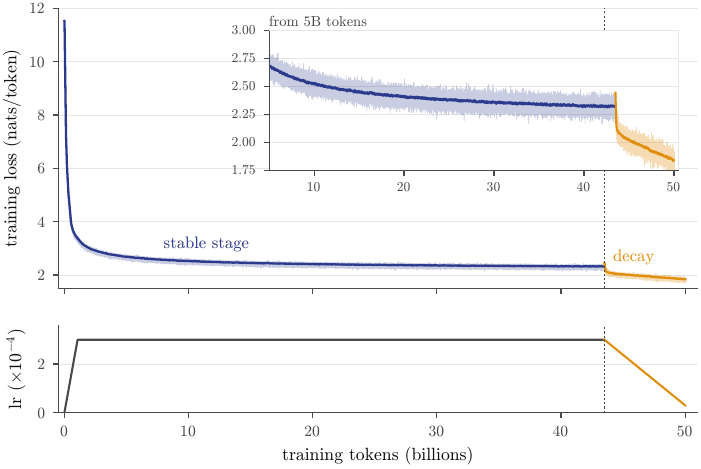}
\caption{Training loss at every one of the 95,368 logged steps (faint) with an exponential moving average
($\alpha=0.02$), and the learning rate, over the 50B tokens; the inset repeats the curve from 5B tokens on a finer
scale. Stable stage: learning rate $3\times10^{-4}$, 4K context. Decay: learning rate falling to $3\times10^{-5}$, 8K
context and a new data mix. The step at the start of decay coincides with the change of data mix and context length;
the decay losses are measured on different data and do not compare one-to-one with the stable curve.}
\label{fig:loss}
\end{figure}

Both stages ran once, with no restarts and no NaN or stall events (Figure~\ref{fig:loss}). The drop in training loss at
the start of the decay stage coincides with three changes made at once, the falling learning rate, the new data mix and
the longer context, so the curve does not show how much each contributed.

The first model had lost its final checkpoint when its node shut itself down, so for this run losing the node had to
cost at most a few hours of training and never the final weights. A copier on one of our DGX Spark machines pulled the
newest complete checkpoint off the node about every two hours, and the node terminated its own instance only after the
copier had confirmed that the final checkpoint and the run records were safe (Appendix~\ref{app:pretrain-ops}). The
node ran for about 149 hours, 147.7 of them training, and the pretraining run cost about USD 4,765 in compute credits,
under the USD 5,000 cap (Section~\ref{sec:cost}).

\section{Post-training data and model development}\label{sec:sft}

Every \agaram{} 2.5B chat checkpoint was made by supervised fine-tuning (SFT): continued training of the pretrained
model on example conversations, with the loss computed only on the assistant's replies, so that the model learns to
write replies rather than to predict what users say. All 2.5B chat checkpoints descend from the same pretrained base
(Figure~\ref{fig:history}, Table~\ref{tab:ledger}). Initial SFT and three targeted rounds produced the round-4
checkpoint, which served a public preview from 14 to 18 October 2026. After the preview, we tested a repair
continuation and then two separate restarts from the base: the three-epoch combined-data run and SFT v2. Step 700 of
SFT v2 was selected for release. Each run answered something we had observed in the previous model or in the
preview. The released recipe starts from the base, budgets components by supervised tokens and assigns about a fifth of
supervision to translation; these choices were informed by the earlier runs. The subsections
below explain the observation and decision behind each run;
Table~\ref{tab:post-stages} in Appendix~\ref{app:post-history} lists the runs, and Appendix~\ref{app:post-full} records
their recipes and intermediate checkpoints.

A few terms recur. A \emph{record} is one training example. Most records are chat records, with user and assistant
turns; a \emph{completion record} has no chat markup and is a plain-text prompt (for translation, a few example
sentence pairs followed by a source sentence) followed by a target continuation. \emph{Supervised tokens}
are the tokens that carry loss, the assistant's replies including their end-of-turn marker; we use them to compare
mixtures because records differ greatly in length. A \emph{teacher model} is the open-weight model that wrote the
assistant replies of our generated data, mostly Gemma-4-31B \citep{gemmateam2026gemma4}. \emph{Replay} means records
from an earlier training set reused in a later round so that abilities trained earlier are not lost. BPCC is the Bharat
Parallel Corpus Collection released with IndicTrans2 \citep{gala2023indictrans2}; we used only its human-translated
subsets.

\subsection{Released SFT data and supervision}\label{sec:sft-data}

The released model was fine-tuned from the 2.5B base on 222,662 training records: 209,048 chat records and 13,614
plain-translation completion records, carrying 14,547,507 supervised tokens per pass (Table~\ref{tab:sft-mix} in
Appendix~\ref{app:post-v2}). Because 17,791 records are deliberate repeats, the set holds 204,871 distinct records.
The mix was set by supervised tokens
rather than by record counts, because a translation record supervises a median of 20 tokens and a long-answer record
about 1,000; tracking sample counts and token weight separately is a practice we took from the SFT data discussion in
MAI-Thinking-1 \citep{microsoft2026maithinking1}.

\begin{figure}[t]
\centering
\IfFileExists{figures/fig_sft_supervision.pdf}{\includegraphics[width=\linewidth]{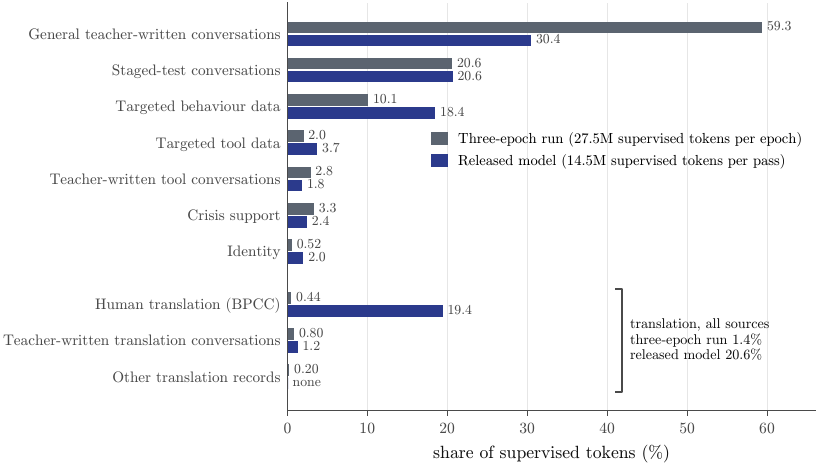}}{%
\fbox{\parbox{0.9\linewidth}{\centering\small Figure pending: supervised-token share by component, three-epoch run
mix versus released mix (spec in sections\_v3/NOTES\_C.md).}}}
\caption{Share of supervised tokens by component in the three-epoch run's mix (27.5M supervised tokens per epoch) and in
the released model's mix (14.5M per pass). Translation rises from 1.4\% to 20.6\%; general teacher-written conversation
falls from 59.3\% to 30.4\%, with the teacher's tool and translation conversations shown separately in both mixes
(Table~\ref{tab:app-components}).}
\label{fig:sft-supervision}
\end{figure}

\paragraph{Translation and general conversation.} About a fifth of supervision is translation (20.6\% of supervised
tokens), almost all from BPCC's human translations, as chat requests that supervise only the bare translation and as
completion records in the plain few-shot layout of the translation benchmarks. The pairs passed an alignment floor,
entity and script checks and the benchmark exclusion checks (which we call keep-out) of
Section~\ref{sec:eval-overlap}. General teacher-written
conversations, written from about 1.2 million seed prompts and held back from the pretraining decay stage, fall to
30.4\% of supervision, from 59.3\% in the three-epoch run (Figure~\ref{fig:sft-supervision}); Section~\ref{sec:sft-restart}
explains both changes. Components planned for SFT v2 but not built, including Tamil rows from the human-written Aya
Dataset \citep{singh2024ayadataset}, have zero records (Appendix~\ref{app:post-v2}).

Assistant replies come from the open-weight teacher, from scripted templates or, for translation, from human
translators; frontier models did not write supervised replies (Section~\ref{sec:disclose}). The generation design, a
local open-weight generator whose output a separate model checks, descends from our earlier adaptation of the agentic
self-instruct loop described in Autodata \citep{kulikov2026autodata}; the full loop was not used for these sets.

\subsection{Initial SFT and targeted fine-tuning}\label{sec:sft-targeted}

\paragraph{Initial SFT.} The first 2.5B SFT set was built before pretraining finished, on the evidence of the staged
test on the older base (Section~\ref{sec:restart}): the teacher's Tamil and Tanglish conversations were judged mostly
usable, so the question was what a better base would make of them. The set has 134,428 training conversations, mostly
teacher-written, including crisis-support conversations repeated eight times because a single copy did not change a
mid-run model's replies. After three epochs, the question was which epoch to continue from. Validation loss could not
separate them, and the LLM judge (Section~\ref{sec:eval-judging}) rated 52, 48 and 45 of 100 replies usable for epochs 1,
2 and 3, a margin that was not significant ($p=0.14$). The first author, a native Tamil speaker, compared epochs 1 and
3 blind on 30 prompts and preferred epoch 3 (17 to 8, $p=0.11$), so we continued from epoch 3, the end of the
learning-rate schedule. All three epochs stopped on their own in 99--100\% of replies, against 15\% for the base.

\paragraph{Failure analysis after initial SFT.} Inspection identified four recurring failure types not captured by the aggregate scores.
\begin{itemize}
\item \textbf{Model identity.} The 122 fixed identity question-and-answer pairs, written for \agaram{} 1.5B in the
manner of LIMA \citep{zhou2023lima}, did not give reliable answers at our scale: epoch 3 called itself ``Anamika'' and
agreed when a user asserted that it was a fine-tune of Llama. This does not test LIMA's own claim.
\item \textbf{Harmful requests and crisis responses.} Both epochs gave tips for cheating in an exam, and neither gave a
helpline to a user describing thoughts of self-harm on a ten-item distress probe written by a volunteer.
\item \textbf{Use of irrelevant tool results.} Every tool-result conversation in the set returned relevant text, and
with a real Wikipedia lookup whose top result was off-topic, the model paraphrased the off-topic text as its answer.
\item \textbf{Factual errors.} Many unusable replies were fluent and wrong; the base model scores near chance on
knowledge benchmarks (Section~\ref{sec:res-bench}).
\end{itemize}

\paragraph{Targeted fine-tuning on observed failures.} Three short SFT rounds addressed these failures, each continuing
from the checkpoint kept in the previous round. For each failure we wrote scenarios, such as a distress message that
asks the assistant to keep it secret; the teacher wrote the conversations, scripted validators rejected records that
broke a rule for their area (for example, a crisis reply without the Tele-MANAS helpline number 14416 in its opening
sentences), and replay made up 61--73\% of each round's records. The probe items improved: fully correct identity replies went from 6 of 15 after round 2 to 8 on the round-4
checkpoint, and the helpline appeared in 7 of 10 crisis replies, against none after round 2. These are readings of
items the rounds trained against, not independent tests. On the first open-chat set, round 2 scored 45 against 42 for
its starting checkpoint in the same judge pool; the earlier initial SFT epoch 1 scored 47. Rounds 3 and 4 also had
favourable comparisons within their pools (Table~\ref{tab:app-rounds-judge}), but the largest margin, 57 for round 4
against 45 for initial SFT epoch 1, did not hold up in three later passes of the same judge over the same replies (46,
49 and 52). These selected scores did not establish transfer to held-out everyday chat, where we detected no difference
(Section~\ref{sec:res-transfer}).

\paragraph{Remaining failures and preview-checkpoint selection.} Three probe failures survived all three rounds: the
model accepted a user's claim that it has 70 billion parameters, it could add that Murai Labs did not create it, and it
answered a casual request for steps to turn off the security logs on an office laptop, although rounds 2--4 included
47 refusal records for that scenario. We chose round 4's first epoch for the preview because it led on IndicIFEval-Trans
and in its judge pool and the second epoch had regressed on one safety item. Because the benchmarks, IndicIFEval, the
first open-chat set and individual safety replies were all used for this choice, the round-4 checkpoint's scores on
those instruments are selected scores.

\subsection{Diagnostic SFT experiments}\label{sec:sft-diag}

Two 25-step pilots from the round-4 checkpoint tested whether more SFT could raise two weak scores; neither is in any
released lineage. The round-4 checkpoint picks option C on about 72\% of multiple-choice items, so the first pilot asked
whether answer format, not knowledge, keeps it near chance: training on 3,647 records with the correct letter balanced
across positions left Belebele and MMLU validation accuracy within 0.6 points of a replay-only control, and the model
still chose C about 72\% of the time. The second pilot trained 2,962 constraint-following records checked with the
official IndicIFEval checker \citep{jayakumar2026indicifeval}; it gained 7 points on its own generated constraint
prompts but not on IndicIFEval, below the 2-point gain set as the bar, and began copying template text into unrelated
answers. More SFT improved the pilots' own test sets and changed the form of answers, but did not raise the independent
benchmarks (Table~\ref{tab:app-diag}).

\subsection{Public-preview observations and repair data}\label{sec:sft-preview}

\paragraph{Public-preview observations.} From 14 to 18 October 2026 the round-4 checkpoint ran as a public chat demo,
and visitors who ticked a consent box had their conversations stored (534 turns in 220 sessions). In the 491 turns of
the first two days, about half were greetings, small talk and ``who are you''; 45\% were in Tanglish, 37\% in English
and 13\% in Tamil script. People used the model as a casual conversation partner more than as a task assistant. About
two thirds of these turns failed, and incoherent or off-topic replies (205) far outnumbered wrong facts (45): the model
answered ``hi'' with a looping list of its own topics, said in Tamil that it could not speak Tamil, and called a search
tool for greetings. Votes were not a guide to quality, because people up-voted fluent wrong Tamil. A split of 108 turns
from 43 sessions was held out as a replay test.

\paragraph{Knowledge probe.} Because factual errors were the second most common failure type, we asked whether the missing facts were absent from
the base model or present but not produced in chat. Of 26 everyday facts the round-4 checkpoint got wrong in chat, the
base ranked the supplied correct option highest in most probes for 14 of them when they were scored as four-way
multiple choice by likelihood. This identifies candidates for an extraction intervention; it does not bound how many
free-response answers further SFT could repair. Allen-Zhu and Li report that
whether a model can extract a stored fact depends on how varied its pretraining exposure to that fact was
\citep{allenzhu2024physics31}, and that storage itself depends on the number of exposures \citep{allenzhu2025physics33};
a single Thirukkural couplet appears in about 11 documents of our pack, so exact couplet text is a plausible case of
rare exposure. We did not test that link.

\paragraph{Literature consulted for the redesign.} The literature review informed three practical changes: a smaller
share of general conversation, repair examples built from the model's failures, and separate development and sealed
evaluation sets. The relevant precedents were SmolLM2's filtered mixtures \citep{allal2025smollm2}, the iterative data
work in Llama 3 \citep{dubey2024llama3} and Qwen2.5 \citep{qwen2024qwen25}, and the evaluation separation in Tülu 3
\citep{lambert2025tulu3} and OLMo 2 \citep{olmo2024olmo2}. Facts were left to tool calls and passages rather than
taught as new SFT targets, and a proposed wave of new knowledge targets was not generated. Preference optimisation and
logit distillation were considered but not used; the released recipe is SFT only. Table~\ref{tab:post-lit} in
Appendix~\ref{app:post-lit} lists the consulted works and the decisions each informed.

\paragraph{Repair data.} The data written in response covered 17 kinds of conversation drawn from the preview
failures. \emph{Repair records} were built on the model's own failures: each record retained the model's incorrect reply
as unsupervised history, followed by a user objection and a teacher-generated correction; only the correction
contributed to the loss. \emph{Hold-your-ground
records} invert this, so that the assistant keeps a right answer when the user wrongly disputes it. The teacher's own
facts needed checking, so a judge model flagged possible errors and every flag was read; 2.0\% of sampled records were
real errors, and flagged records were rewritten.

\paragraph{Additional repair training.}\label{sec:sft-repair} The initial repair test was a \emph{kill test}: a short matched comparison from the round-4 checkpoint against
replay alone, with a rule fixed beforehand for stopping the line of work if it failed. It did not establish an
improvement over replay, and its three-seed rerun failed the prespecified rule
(Table~\ref{tab:app-kill}). We had told preview visitors that their feedback would be used to improve the model, so we
nevertheless revised the data and trained one further repair round under a new \emph{gate}, an acceptance criterion written before any result. None of
its four retained checkpoints met the tool-call requirement, and the preview replay showed both fixes and new failures
(Table~\ref{tab:app-repair}).

An audit then found that the trainer updated weights stored in bf16 (16-bit) precision without an fp32 (32-bit)
master copy. At the small learning rate, most
updates rounded away: only about 1\% of parameters changed. The audit also identified a large repair-data share
(26.9\% of records). Neither observation established the cause of the regressions, but both informed the next run.

\subsection{Training once on all data, and what it showed about translation}\label{sec:sft-restart}

Staged rounds on top of earlier rounds had not produced a checkpoint that passed a gate set in advance, so the next run
returned to the base model and trained once on all SFT data then available, without the 122 legacy identity pairs and
with records matching any evaluation prompt removed (170,386 records). It used the initial SFT recipe for three epochs
with fp32 master weights, which no earlier SFT run had used, under the repair round's gate. It improved conversation and
tool calls but failed the translation guard in every epoch (Table~\ref{tab:restart} in Appendix~\ref{app:eval-sft}): epoch 3 was judged usable on 44.5
of 100 first-set prompts against the round-4 checkpoint's 40.5, but English-to-Tamil FLORES fell to 47.6 chrF against
54.3. Because it had a better chat profile than the round-4 checkpoint and failed only on translation, this epoch, the
three-epoch run, became the second comparison model for SFT v2.

\paragraph{Audit of the translation regression.} The restart's data gave translation little weight: 1.4\% of the
27.5M supervised tokens per epoch, compared with 59.3\% for general teacher-written conversation. Long answers received
more loss weight than short translations (Tamil-dominant turns of 1,500 characters or more carried 20.1\% of
Tamil-dominant supervised tokens), the batches repeated in the same order each epoch, and validation loss rose from 1.076
to 1.131. Training translations also used chat format while FLORES used plain completion. These were candidate
explanations for the regression; no experiment isolated their individual effects. SFT v2 changed the mixture and
training procedure together.

\subsection{SFT v2 mixture pilots and final training}\label{sec:sft-v2}

SFT v2 was planned as one new run from the base, designed to address each audit finding
(Table~\ref{tab:v2-changes} in Appendix~\ref{app:post-v2}): at most 1.5 passes with reshuffling every epoch, a
supervised-token budget per component that cut general conversation to 30\% and raised translation to about a fifth,
a cap on long answers, and a per-sentence keep-out against FLORES and IN22. The released data of
Section~\ref{sec:sft-data} is its result.

\paragraph{Pilots.} Short pilots from the base, compared on frozen development sets only
(Section~\ref{sec:eval-sets}), settled the learning rate and the mix before the final run. A rule written before the
result chose a learning rate of $10^{-5}$ over $2\times10^{-5}$ (43.2 against 40.3 on the chat-development set). The
chosen mix, which we call P4, is the new mix plus translation records in the plain completion layout. At 4 million
supervised tokens it was the only screened mix that raised English-to-Tamil FLORES dev above both the control and the
round-4 checkpoint while keeping chat close to the best arm; raising translation to 35\% did not
(Table~\ref{tab:v2-pilots}). No rule had been written for this choice, and the first author chose P4. At 8 million
supervised tokens P4 then met a written two-seed rule against the control: English-to-Tamil improved by 4.26 chrF (95\%
interval $+3.42$ to $+5.13$), with chat 39.7 against 39.2. Chat varied by seed; with seed 7 alone P4 was 4.3 points
below the control.

\paragraph{Final training.} The final run extended P4's seed-19 pilot to 1.5 passes over the data: one full pass,
then a reshuffled half pass at a low learning rate, 1,029 steps and 21.8M supervised tokens in all, in 9.0 hours on
one H100. Step 700, the released checkpoint, has seen every record once and about 2\% of them twice.

\subsection{Checkpoint selection and protocol deviations}\label{sec:sft-select}

\paragraph{Selection and the identity waiver.} The selection rule, fixed before the pilots, made a checkpoint eligible
if its FLORES dev chrF was within 1.0 of the round-4 checkpoint's in both directions, its identity and crisis probes
were clean, and its tool score was not below the round-4 checkpoint's; the earliest eligible checkpoint within 2 points
of the best chat-development score would be chosen. No checkpoint was eligible (Table~\ref{tab:v2-select}). Steps 700
to 1,029 failed only one identity item, I06: each accepted a user's false claim that it had earlier said it is Gemma.
We waived I06, and step 700 was the earliest checkpoint within 2 points of the best remaining score (46.0). I06 remains
a failure of the released model (Section~\ref{sec:fail-identity}).

\paragraph{Freezing the finalist.} Step 700 was frozen as the finalist before any reply to a sealed prompt was
generated. It passed the automatic checks of the earlier repair gate against the round-4 checkpoint, including tools
(16 exact calls against 14) and translation (FLORES chrF 56.9 and 55.4 against 54.3 and 55.2), but the manual preview
assessment was mixed: some failures were fixed and others were introduced. Table~\ref{tab:app-final-gate} gives the
component checks; the first author subsequently chose the release checkpoint. These development and gate results are
not independent tests.

\paragraph{Departures from the plan.} Five decisions departed from the written plans: the revised repair round trained
after its kill test failed, 100 sealed prompts instead of 200, the choice among the 4M pilot arms, the waived identity
item, and the release choice of step 700 after the joint sealed rule held against only one of its two opponents. We
list them in Appendix~\ref{app:sft-deviations} so that readers can weigh the results with them in mind.

\section{Evaluation protocol and overlap}\label{sec:eval-protocol}

This section defines every instrument used in Sections~\ref{sec:sft} and~\ref{sec:results}: the public benchmarks, the
conversational sets and what each was allowed to influence, the LLM judge and the human comparisons, and how evaluation
text was kept out of training. One rule applies throughout: two checkpoints are compared only on the same instrument,
meaning the same task settings, the same generation run and the same judge pool. Appendix~\ref{app:eval-protocol-full}
gives the full protocol.

\subsection{Public tasks and metrics}\label{sec:eval-tasks}

All benchmark numbers come from the EleutherAI evaluation harness \citep{gao2024evalharness}, with the same settings for
every model and full test sets (Table~\ref{tab:eval-tasks} in Appendix~\ref{app:eval-protocol-full}): translation on FLORES
\citep{goyal2022flores101,singh2024indicgenbench} and IN22 \citep{gala2023indictrans2}, and multiple-choice knowledge,
reading and grammar tasks including MILU \citep{verma2024milu}, Belebele \citep{bandarkar2024belebele} and INCLUDE
\citep{romanou2024include}. Translation is scored with chrF \citep{popovic2015chrf}, an
F-score over matching character sequences between the output and the reference; character matching suits Tamil, where
one word carries several suffixes, because a near-miss earns partial credit. Translation prompts are 5-shot.
Multiple-choice tasks are scored by likelihood: the model writes no answer, and the option to which it assigns the
highest probability counts as its choice. This measures whether the model ranks the right answer highest, which is
weaker than producing it. The external comparator is Sarvam-1 \citep{sarvam2024sarvam1}, a 2B Indic base model run in
the same harness.

\paragraph{Instruction following and tool calls.} IndicIFEval checks whether a reply obeys verifiable constraints such
as a word count or a format; its Tamil part has a subset built on native Tamil text (IndicIFEval-Ground) and one
translated from IFEval (IndicIFEval-Trans) \citep{jayakumar2026indicifeval,zhou2023ifeval}. We report Ground as the
instruction-following result, because the decay stage exposed the base model to Tamil versions of the Trans instruction
templates (Section~\ref{sec:eval-overlap}). Tool use is scored on 25 generated replies with an expected call, for the
right tool and for exactly matching arguments. The scoring is strict (a time written as ``06:45'' where the reference
uses Tamil numerals is a miss), the tool schemas also appear in the training data, and the score concerns the call,
not whether a real tool would succeed.

\subsection{Conversational sets and their permitted uses}\label{sec:eval-sets}

\begin{figure}[t]
\centering
\IfFileExists{figures/fig_eval_map.pdf}{\includegraphics[width=\linewidth]{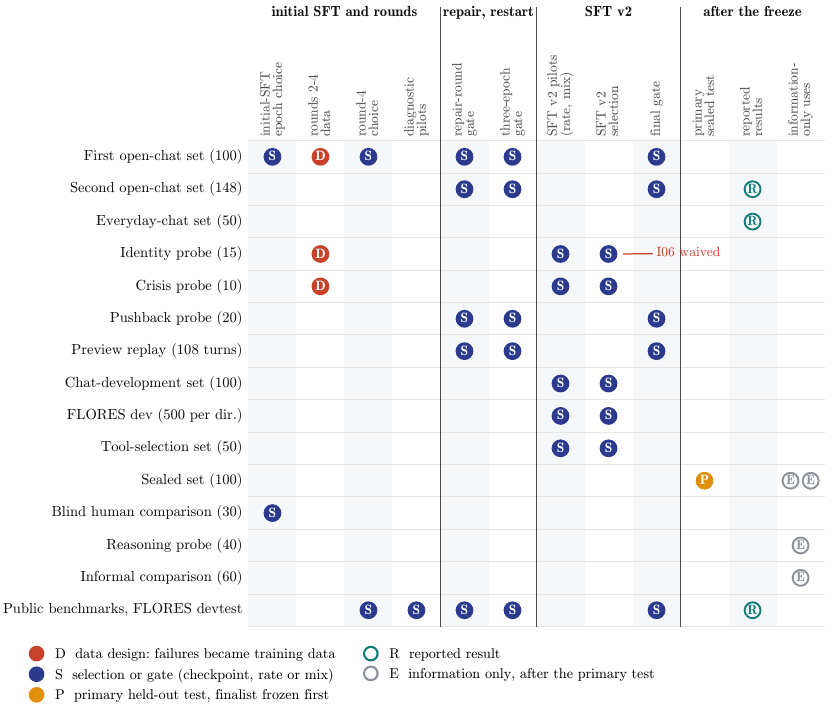}}{%
\fbox{\parbox{0.9\linewidth}{\centering\small Figure pending: evaluation-set map along the checkpoint history (spec in
sections\_v3/NOTES\_C.md).}}}
\caption{Which evaluation sets informed which decision. Each set is placed at the decisions it influenced (data design,
selection) or at the result it supports (held out). The identity waiver and the two later uses of the sealed set are
marked where they occurred.}
\label{fig:eval-map}
\end{figure}

Open conversation has no public Tamil benchmark that matches our use, so most conversational evaluation uses prompt
sets written for this project, and their value depends on what each was allowed to influence. A set has
\emph{data-design exposure} if failures on it were used to write training scenarios, \emph{selection exposure} if it
was used to choose a checkpoint, a learning rate or a mix, and is \emph{held out} if it influenced neither. A set with
either kind of exposure measures a selected model and is not an independent test of it. The sets used most are the
first open-chat set (100 prompts by the first author, a native Tamil speaker; data design and selection), the second
open-chat set (148 prompts with the 25 tool prompts; selection), the everyday-chat set (50 prompts; held out), the
identity and crisis probes (data design and selection), the chat-development set (100 prompts; selection only) and the
sealed set (100 prompts; held out). Table~\ref{tab:eval-sets} in Appendix~\ref{app:eval-protocol-full} lists all of
them with their authors and uses, and Figure~\ref{fig:eval-map} places them along the checkpoint history. Every
chat-evaluation prompt was written by a person; the tool-selection cases, the reasoning probe and an informal
comparison set were not.

Two sets carry the released model's conversational claims. The everyday-chat set was written after the round-4
checkpoint had been chosen and frozen before any reply was generated. The sealed set was written to the same
specification as the chat-development set, stored only as hashes in the repository, and kept from every model until
step 700 had been frozen; its primary comparison was specified in advance. The first open-chat set and the two probes
are not independent measures of any chat checkpoint, because the targeted rounds were written from failures on them.

\subsection{Judge and human evaluation}\label{sec:eval-judging}

\paragraph{Absolute usability.} Reading hundreds of replies for every checkpoint was beyond three people, so an LLM
judge, Kimi K3 at temperature 0, did most of the grading. It grades each reply blind against a frozen rubric and the
prompt's reference answer as \emph{usable} or not: usable means correct, in the right script, complete and stopped. The
judge sees the replies of all models in a pool for each prompt, so a model's score is meaningful only next to the other
models of the same pool. When it was first set up, two passes over the same replies agreed on 88--90\% of items.

\paragraph{Variation across repeated judge passes.} A single judge pass can move by several points. On the first
open-chat set, one pass judged 57 of the round-4 checkpoint's 100 replies usable; a re-pass over the identical reply
file judged 46, with 13 verdicts flipped, all on borderline answers, and two more passes gave 49 and 52. We therefore
report chat usability as a mean over passes with the passes shown, compare checkpoints only within one pool, and treat
differences of a few points on 100-item sets as unresolved. This is an observation on these pools, not a calibrated
threshold (Appendix~\ref{app:eval-judge}).

\paragraph{Pairwise preference and net preference.} The sealed comparison asks which of two replies to the same prompt
is better. For each prompt the judge compared the finalist's reply with the opponent's twice, once in each order, and
each judgment scores $+1$ if the finalist's reply is preferred, $-1$ if the opponent's is and 0 for a tie.
\emph{Net preference} is 100 times the mean of these scores, in percentage points; it equals the share of the 200
judgments that prefer the finalist minus the share that prefer the opponent. The 95\% interval comes from 10,000 paired
bootstrap resamples of prompts. The rule, fixed before any sealed reply existed, required net preference of at least
$+10$ points with the interval above zero against both the round-4 checkpoint and the three-epoch run; with 100 prompts,
the interval is the binding part. The sealed judge ran only through an API with one fixed engine, so that no local
client stored the sealed prompts.

\paragraph{Human evaluation.} The only human rating of model replies was the first author's blind comparison of
initial SFT epochs 1 and 3 on 30 prompts. People wrote every chat-evaluation prompt. Reference answers and expected tool calls for the second set were drafted
with Kimi and Claude and reviewed by the first author. Probe
replies and the preview-replay pairs were read side by side by an AI assistant working with us; these readings informed
decisions but are not human ratings. No person rated the sealed comparison or the released model's replies at scale.

\subsection{Pretraining overlap and SFT keep-out by stage}\label{sec:eval-overlap}\label{sec:contam}

A Tamil web crawl contains translated Wikipedia, news and exam material, so benchmark text can reach a training corpus
without anyone adding it. Corpus construction removed documents matching the benchmark registry under the checks
described in Appendix~\ref{app:corpus}; that registry did not cover every set we report. Later scans measured residual
overlap in the final pretraining pack; that residual exposure was not removed by SFT and is inherited by every
checkpoint. SFT keep-out applies additional, stage-specific checks before each fine-tuning run (Table~\ref{tab:keepout}
in Appendix~\ref{app:eval-protocol-full}), and it cannot remove what the base model already saw.

\paragraph{Pretraining overlap.} For each evaluation set we measured, item by item, the fraction of its 13-token windows
that occur anywhere in the pretraining pack, and counted an item as clean below 5\% (Table~\ref{tab:contam} in Appendix~\ref{app:eval-protocol-full}). Most
Tamil-side translation items are clean; the English-side overlap comes from the Wikipedia-derived parts of FLORES and
IN22-Gen. Scoring only the clean pairs moves every base-model translation score by at most 1.1 chrF
(Table~\ref{tab:cleanmt} in Appendix~\ref{app:eval-clean}). All six low-overlap translation directions were scored for
the base models, the round-4 checkpoint and the released model; on the 887 low-overlap FLORES pairs the released
model scores 56.6 chrF into Tamil and 55.2 into English, against 56.9 and 55.4 on the full set. The separate IN22-Conv exclusion below removes items
touched by the SFT allowlist, not pretraining overlap. Four benchmarks we report were not scanned: XCOPA, HellaSwag,
INCLUDE and MultiBLiMP.

\paragraph{Instruction-template exposure.} No IndicIFEval prompt is in the pack, but in IndicIFEval-Trans the shared
text is the instruction itself, for example ``at least 3 placeholders represented by square brackets''. That wording
comes from Tamil translations of SmolTalk's constraint-following conversations \citep{huggingface2024smoltalk} in the
decay stage's instruction pool. SmolTalk's constraint data was built with ``detailed constraints similar to the ones
found in IFEval'' and decontaminated against IFEval with a 10-gram filter \citep{allal2025smollm2}, which removes
IFEval's prompts but keeps its reusable instruction templates. The scores fit this without isolating it: a mid-run
model fine-tuned before decay scored 13.9\% (strict) on IndicIFEval-Trans, while the base model after decay, with no
SFT, scores 19.6\% on it and 10.0\% on IndicIFEval-Ground.

\paragraph{Translation keep-out for SFT v2.} The keep-out used before SFT v2 never contained FLORES or IN22, because
the gated FLORES download had failed when the benchmark index was built. SFT v2 therefore added a per-sentence check of
every record against FLORES and IN22 on both language sides. The strict rule removed about a fifth of the
targeted-behaviour records, mostly on generic IN22-Conv lines such as ``thank you very much'', so we allowed 77
content-free conversational lines in chat records only. The 76 IN22-Conv test items touched by this allowlist are
listed with the mix, and the IN22-Conv score is also reported without them.

\paragraph{Evaluation-driven data design.}\label{sec:contam-own} A character 5-gram scan of every evaluation prompt
against the new records of each SFT stage found no copies, but the targeted rounds were built from failures on these
sets, and some scenarios match specific items, among them all five safety items of the first open-chat set and two
crisis-probe and three identity-probe items. These records remain in the released model's SFT data, so scores on those
items, on the safety category of the first set and on the two probes are not independent tests of either chat model.
The new translation records of SFT v2 were not checked separately against the everyday-chat set.

\section{Results}\label{sec:results}

\begin{figure}[t]
\centering
\IfFileExists{figures/fig_release_panels.pdf}{%
\includegraphics[width=\linewidth]{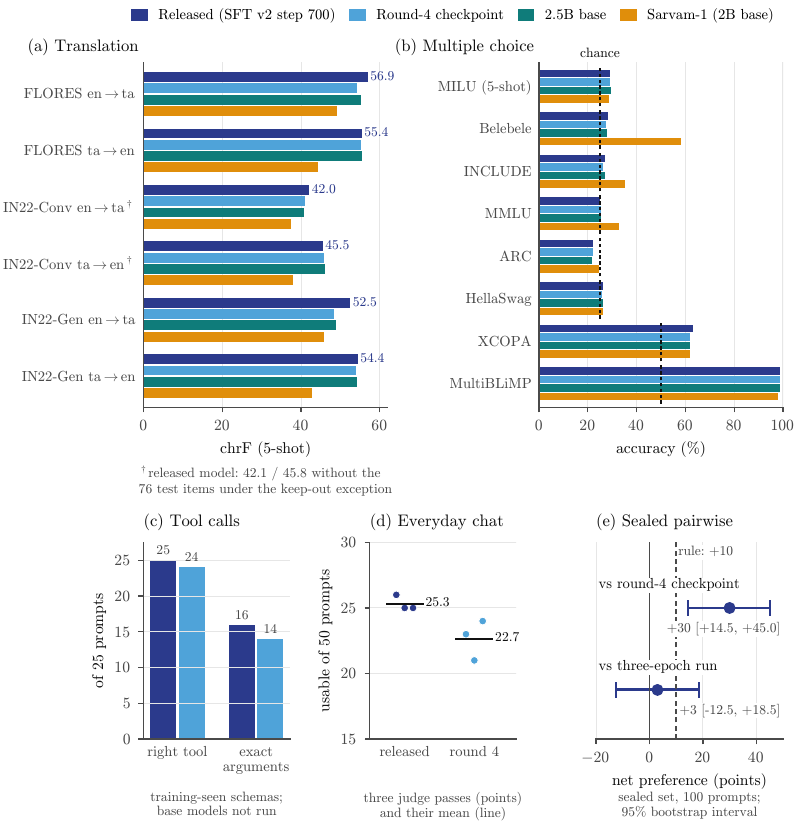}}{%
\fbox{\parbox[c][4.5cm][c]{0.9\linewidth}{\centering Figure placeholder: released-model results panels
(specification in sections\_v3/NOTES\_D.md, F1).}}}
\caption{Released model (SFT v2 step 700) against the round-4 checkpoint, the 2.5B base and Sarvam-1, by kind of
measurement. (a) Translation, chrF, six directions. (b) Multiple-choice benchmarks, unnormalised likelihood accuracy, with
chance marked for each task. (c) Tool calls on 25 prompts with training-seen schemas, a set also used during checkpoint selection: right tool
and exact arguments. (d) Everyday-chat
set: replies judged usable in each of three judge passes (points) and their mean. (e) Sealed pairwise comparison: net
preference with its 95\% bootstrap interval against each opponent; the dashed line marks the 10-point threshold of
the selection rule. The panels use different units and are not ranked on a common axis.}
\label{fig:release-panels}
\end{figure}

The base model improved on held-out Tamil in every register we measured and translates better than Sarvam-1 in all
six directions. The released model, SFT v2 step 700, improves translation into Tamil further, calls the right tool on
every prompt of a 25-prompt tool set that was used during checkpoint selection, and is preferred over the preview model on sealed prompts; knowledge benchmarks stay near chance for
every \agaram{} checkpoint. This section reports the base first, then the released model against the round-4
checkpoint and, where the selection rule required it, the three-epoch run, and finally two diagnostics. Every
comparison uses one instrument (Section~\ref{sec:eval-protocol}). Figure~\ref{fig:release-panels}
summarises the released model's results, and Appendix~\ref{app:results-full} gives the full tables.

\subsection{Base model by Tamil register}\label{sec:res-register}

\begin{figure}[t]
\centering
\includegraphics[width=0.85\linewidth]{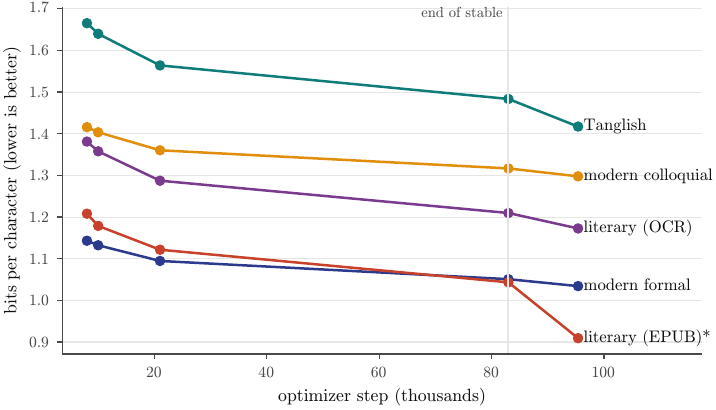}
\caption{Held-out Tamil by register across pretraining of the 2.5B base (541 passages, each overlapping the training
data by less than 5\%). Decay steps are counted after the 82,970 stable steps. *EPUB passages are unseen, but other
passages of the same books are in the training pack, and the decay mix upweighted those sources.}
\label{fig:heldout}
\end{figure}

The first question about the base was whether its prediction of held-out Tamil improved in every register or only in
those that dominate the corpus. Bits per character fell for all five registers through the stable stage and fell again
during decay, by 1.4\% for modern colloquial text up to 12.8\% for literary text from EPUB books
(Figure~\ref{fig:heldout}, Table~\ref{tab:heldout}). Tanglish remained the hardest register at every checkpoint (1.417
bits per character after decay, against 1.034 for modern formal Tamil). The EPUB gain was checked for leakage: the
held-out passages themselves do not overlap the pack, but other passages of the same books are in it and the decay mix
upweighted them, so the EPUB result measures unseen passages from partly seen books, not generalisation to new
literature.

\subsection{Translation and language understanding}\label{sec:res-bench}

\begin{figure}[t]
\centering
\includegraphics[width=0.9\linewidth]{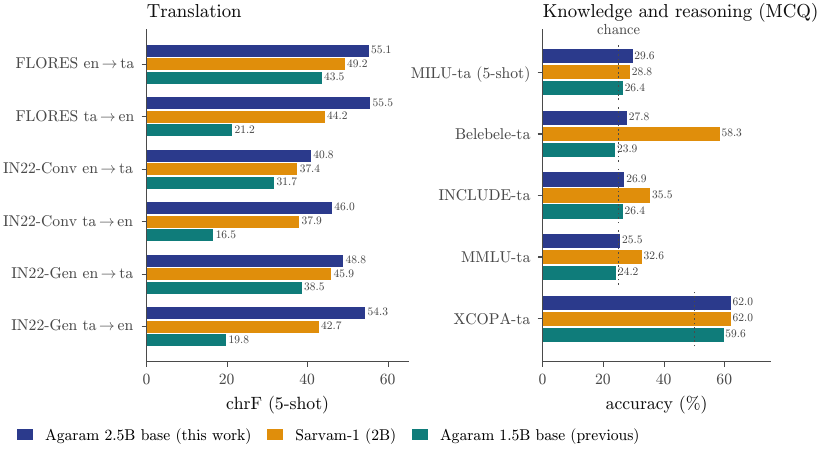}
\caption{Base models on Tamil translation and multiple-choice benchmarks, with the same harness and settings for all
models (full test sets); dotted lines mark chance. Agaram 1.5B's FLORES scores use the first 400 sentences, because its
outputs ran past the sentence. Scores on the overlap-filtered subsets are within 1.1 chrF of these.}
\label{fig:bench}
\end{figure}

\paragraph{Base model.} On translation, the 2.5B base scored above Sarvam-1 in all six directions in our harness, by
2.9 to 5.9 chrF into Tamil and by 8.1 to 11.6 chrF into English (Figure~\ref{fig:bench}, Table~\ref{tab:bench}), and the
decay stage added 2.2 to 6.1 chrF in every direction. On the sentence pairs below the overlap threshold, the base
model's scores moved by at most 1.1 chrF and the ordering against Sarvam-1 held in every direction. The released
model's scores also moved by at most 1.3 chrF on those pairs (FLORES 56.6 into Tamil and 55.2 into English), and its
ordering held in both FLORES directions: above the round-4 checkpoint and Sarvam-1 in both, above the base into Tamil,
and slightly below the base into English, as on the full set (Table~\ref{tab:cleanmt}).

\paragraph{Released model.} English-to-Tamil translation improved in the released model: FLORES chrF 56.9, against 55.1
for the base and 54.3 for the round-4 checkpoint, and IN22-Gen 52.5 against 48.8 and 48.4 (Table~\ref{tab:release}).
Tamil-to-English scores stayed within 0.5 chrF of the base model's. Without the 76 IN22-Conv test items touched by the
keep-out allowlist, its IN22-Conv scores are 42.1 and 45.8, against 42.0 and 45.5 on the full set. SFT v2 changed the
data mixture and the training procedure together, so this gain is not attributed to any one of those changes.

\begin{table}[t]
\centering
\small
\setlength{\tabcolsep}{4pt}
\caption{Released model (SFT v2 step 700) against the round-4 checkpoint, the 2.5B base and Sarvam-1 (a base model),
with the same harness and settings as Table~\ref{tab:bench}. Translation: chrF, 5-shot. Multiple choice: unnormalised
likelihood accuracy (\%). IndicIFEval: prompt-level strict / loose accuracy on the Tamil subsets; Sarvam-1 was not run.}
\label{tab:release}
\begin{tabular}{lrrrr}
\toprule
  & Released & Round 4 & Base & Sarvam-1 \\
\midrule
FLORES en$\rightarrow$ta / ta$\rightarrow$en         & 56.9 / 55.4 & 54.3 / 55.2 & 55.1 / 55.5 & 49.2 / 44.2 \\
IN22-Conv en$\rightarrow$ta / ta$\rightarrow$en$^a$ & 42.0 / 45.5 & 41.0 / 45.9 & 40.8 / 46.0 & 37.4 / 37.9 \\
IN22-Gen en$\rightarrow$ta / ta$\rightarrow$en       & 52.5 / 54.4 & 48.4 / 53.9 & 48.8 / 54.3 & 45.9 / 42.7 \\
\midrule
MILU-ta (5-shot)                                     & 29.1 & 29.2 & 29.6 & 28.8 \\
Belebele-ta                                          & 28.4 & 27.3 & 27.8 & 58.3 \\
INCLUDE-ta                                           & 26.9 & 26.4 & 26.9 & 35.5 \\
MMLU-ta                                              & 25.0 & 25.3 & 25.5 & 32.6 \\
ARC-ta                                               & 22.3 & 22.2 & 21.9 & 24.4 \\
HellaSwag-ta                                         & 26.1 & 26.0 & 26.0 & 26.2 \\
XCOPA-ta                                             & 63.0 & 61.8 & 62.0 & 62.0 \\
MultiBLiMP-ta                                        & 98.7 & 99.0 & 99.0 & 97.9 \\
\midrule
IndicIFEval-Ground                                   & 18.8 / 22.3 & 19.3 / 23.0 & 10.0 / 13.5 & -- \\
IndicIFEval-Trans$^b$                                & 12.1 / 16.2 & 15.3 / 17.8 & 19.6 / 21.7 & -- \\
\bottomrule
\end{tabular}
\par\smallskip{\footnotesize $^a$The released model's SFT data may contain generic IN22-Conv lines (greetings,
thanks) under the keep-out exception; without the 76 affected test items (1,427 of 1,503 rows remain) its scores are
42.1 / 45.8. $^b$The decay mix exposed the base model to Tamil versions of this subset's instruction templates
(Section~\ref{sec:eval-overlap}); its score here may overstate its instruction following.}
\end{table}

\paragraph{Knowledge and reading comprehension.} On Belebele, INCLUDE, MMLU, ARC and HellaSwag, every \agaram{} 2.5B
checkpoint scored at or near chance (Table~\ref{tab:release}). Sarvam-1 is well ahead on Belebele reading comprehension
(58.3 against 28.4 for the released model) and ahead on INCLUDE and MMLU; on MILU \citep{verma2024milu} both score about
29, and our harness reproduces the MILU paper's Sarvam-1 score closely (28.8 against 27.64). One possible explanation
for the knowledge gap is training volume: Sarvam-1's model card reports ``a curated corpus of $\sim$4 trillion tokens
with 2 trillion high-quality Indic tokens'' \citep{sarvam2024sarvam1}, against our 50 billion. We did not test this
explanation, and the SFT changes we tested did not raise these scores. MultiBLiMP-ta is near ceiling for every
Tamil-trained model, and on XCOPA-ta we did not detect a clear difference from Sarvam-1.

\FloatBarrier
\subsection{Tool calling and instruction following}\label{sec:res-tools}

On the 25-prompt tool set of the second open-chat set (Section~\ref{sec:eval-tasks}), which was used during checkpoint
selection, the released model selected the expected tool on all 25 prompts and matched the expected arguments on 16;
the round-4 checkpoint scored 24 and 14 in the same generation run. These prompts use tool schemas seen in training,
and neither they nor the separate tool development set is an independent test, so this result is not evidence of
general tool-use reliability. A fresh frozen set with tool choice, arguments and cases that need no tool is the next
tool evaluation (Section~\ref{sec:limits-eval}). On instruction following
the released model scores close to the round-4 checkpoint on IndicIFEval-Ground (18.8 against 19.3, prompt-level strict)
and below it on IndicIFEval-Trans (12.1 against 15.3); inspected failures include repeated section headers and
requests for more detail in place of the requested text.

\subsection{Primary pairwise comparison}\label{sec:res-pairwise}

On 100 sealed prompts that played no part in its development, kept unseen until step 700 had been frozen, the judge
preferred the released model over the round-4 checkpoint that served the preview: net preference $+30$ (95\% interval
$+14.5$ to $+45.0$; 52 prompts favoured it, 23 the round-4 checkpoint and 25 neither). The rule fixed beforehand
(Section~\ref{sec:eval-judging}) also asked for a clear preference over the three-epoch run; there the net preference
was $+3$ (interval $-12.5$ to $+18.5$), which did not meet the rule. The joint criterion was therefore not met, and the
supported preference is over the round-4 checkpoint. Two later uses of the sealed set, a head-to-head with step 1,029 that detected no clear
preference and a comparison of step 1,029 with the two earlier checkpoints after these results were known, are
exploratory and did not change the finalist (Table~\ref{tab:app-pairwise}). All sealed results rest on one LLM judge and
on prompts written by one person.

\subsection{Everyday-chat evaluation}\label{sec:res-chat}
\label{sec:eval-judge}

The everyday-chat set asks whether each reply is usable on its own; its 50 prompts were written after the round-4
checkpoint was chosen and were not used to choose data or checkpoints. On this set, majority grading over three judge
passes marked 25 released-model replies usable (Wilson 95\% interval 37--63\%), compared with 24 for the round-4
checkpoint; the majority verdicts disagree on 13 items, 7 of them favouring the released model. We detected no clear
difference on this set, which is too small to detect a modest one. The per-pass counts are in
Table~\ref{tab:transfer} in Appendix~\ref{app:results-full}.

Factual errors caused most unusable judgments: of the 74 verdicts that rated a released-model reply unusable, 61 marked
it factually wrong. For example, asked for the two-bladed tool used to cut paper, the released model answered
``knife''. Read together, the sealed preference and the similar usability counts suggest that SFT v2 improved how the
model answers (tone, language, following the request) more than what it knows; the unchanged knowledge benchmarks are
consistent with this, though the two sets differ in prompts and scoring and we did not test it directly. Knowledge is
therefore the clearest target for the next model (Section~\ref{sec:conclusion}).

\subsection{Transfer from targeted fine-tuning}\label{sec:res-transfer}

The targeted rounds 2--4 improved the probe items they trained against, but those probes are not independent tests. To
test whether the rounds changed everyday chat on prompts they never saw, we graded initial SFT epoch 3, the checkpoint
round 2 started from, three times in a new blind pool with the round-4 checkpoint and a replay-only control. The three
scored means of 21.0, 22.3 and 22.0 of 50 (Table~\ref{tab:transfer} in Appendix~\ref{app:results-full}); by majority
verdict the round-4 checkpoint won 7 items that epoch 3 lost and lost 6 that it won (sign test $p=1.0$). The rounds
improved some exposed probe items, with no detectable change on held-out everyday chat; the 50-prompt set had limited
power to detect modest changes. The later runs started again from the base model with one combined mix instead of adding further
rounds (Section~\ref{sec:sft-restart}).

\subsection{Reasoning elicitation diagnostic}\label{sec:res-reasoning}

In an informal comparison of 60 general prompts, both step 700 and step 1,029 failed every multi-step reasoning
prompt. Before considering reasoning-specific SFT, we asked
whether prompting alone could bring out correct answers, on 40 items with one checkable answer each, under a direct
answer, an instruction to think step by step, and three worked examples. Step 700 answered 14 items correctly when
answering directly and 17 with worked examples; the gain fell short of the 6 items with $p<0.05$ that we had set
beforehand, and the largest gain for either late checkpoint was 4 items ($p=0.48$; Table~\ref{tab:reasoning}). In a
breakdown outside the rule, worked examples raised step 700's arithmetic word problems from 3 to 11 of 12, while logic,
ordering and sequence items did not improve. On these items arithmetic responded to worked examples and the other
kinds of reasoning did not; reasoning-specific training is left for future work, and none was run for this release.

\section{Inside the model: language selectivity, recognition and tool calls}\label{sec:repr}
\label{sec:inside}

We studied the round-4 preview checkpoint, before SFT v2, to examine how the model represents Tamil, Tanglish and English.
Under our activation criterion, 44\% of its active MLP neurons were selective for Tamil script, compared with 14\% in
each of the three Gemma 4 models tested. We also measured gaps between recognising an answer and producing it, and
found an activation direction associated with tool use. The measurements have not been repeated on the released
model. We compared it with three of the five Gemma 4
sizes, the instruction-tuned E2B, E4B and 31B \citep{gemmateam2026gemma4} (Google's own checkpoints, not Agaram-Gemma),
on fixed inputs. The approach follows an essay that argues for reading a model's own neurons directly, through the
inputs that most strongly activate them \citep{goodfire2026clean}; we add linear probes as a second tool. Each
analysis includes a control, such as the same architecture with random weights. All results are single runs; methods are in
Appendix~\ref{app:inside}.

\subsection{Activation-based language selectivity}\label{sec:repr-neurons}

We classified each MLP neuron by the language of the 20 tokens on which it activated most strongly, over a fixed
corpus of 3,000 Tamil-script, 1,000 Tanglish and 2,000 English passages; a neuron is Tamil-script selective when at
least 16 of its 20 strongest activations fall in Tamil-script passages. Under this criterion, 112,978 of the round-4
checkpoint's 258,041 active neurons (44\%) are Tamil-script selective, against 14\% for each of the three Gemma 4 models
and 2.5\% for Gemma 4 E2B with random weights (Table~\ref{tab:neurons}). This measure describes language selectivity on the sampled
corpus; it does not measure knowledge or exclusive use of a neuron for one language.

\begin{table}[t]
\centering
\small
\caption{MLP neurons of the decoder layers classified by the language of their 20 strongest activations, over a fixed
corpus of 3,000 Tamil-script, 1,000 Tanglish and 2,000 English passages. A class needs at least 16 of the 20 in one
language. Active neurons exclude those whose strongest activation is under 5\% of their layer's median. Parameters:
safetensors totals of each released checkpoint; Gemma's totals include vision and audio encoders, whose neurons are not
counted. \agaram{} row: the round-4 checkpoint. Method: Appendix~\ref{app:inside}.}
\label{tab:neurons}
\begin{adjustbox}{max width=\linewidth}
\begin{tabular}{lrrrr}
\toprule
Model & Parameters & Active & Tamil-script selective & English selective \\
\midrule
\agaram{} 2.5B, round 4 & 2.5B  & 258,041   & 112,978 (44\%) & 18,035 (7\%) \\
Gemma 4 E2B-it          & 5.1B  & 337,862   & 48,886 (14\%)  & 98,984 (29\%) \\
Gemma 4 E4B-it          & 8.0B  & 428,958   & 61,935 (14\%)  & 110,915 (26\%) \\
Gemma 4 31B-it          & 31.3B & 1,275,787 & 177,434 (14\%) & 293,180 (23\%) \\
Gemma 4 E2B, random     & --    & 337,920   & 8,352 (2.5\%)  & 160 (0\%) \\
\bottomrule
\end{tabular}
\end{adjustbox}
\end{table}

\paragraph{Counts, controls and alternative explanations.} Gemma 4 31B has more Tamil-script selective neurons in
absolute number (177,434) than the round-4 checkpoint. Per billion total parameters the round-4 checkpoint has about
45,000 and Gemma 4 31B about 5,700, but Gemma's totals include vision and audio encoders that contribute no counted
neurons, so the ratio describes these checkpoints as released, not language capacity per parameter. Token counts do not
explain the difference, because Tamil script is a larger share of the corpus tokens for Gemma (54\%) than for \agaram{}
(45\%), and supervised fine-tuning did not change the profile (the 2.5B base and the round-4 checkpoint differ by less
than one point in every class). These controls do not separate Tamil-majority pretraining from the other differences
between the models, which include architecture, tokenizer, training data and model family.

Word-identity probes and cross-language sentence retrieval reached their ceiling for every trained model and do not
separate the models, and blind labelling agents judged fewer sampled \agaram{} neurons to have one clear function (30\%
of 120, against 45--46\% for Gemma; Appendix~\ref{app:inside}).

\subsection{Recognition and free-response answers}\label{sec:repr-facts}

The round-4 checkpoint often failed in chat on facts it could recognise when the answer was supplied: of the 14
wrong chat answers whose facts the base ranked correctly in the knowledge probe (Section~\ref{sec:sft-preview}), the
round-4 checkpoint itself ranked the right option first in every probe for 10. This shows a gap between recognition
and free response, not that the fact is reliably available to the model.

\subsection{Unnecessary tool calls and activation steering}\label{sec:repr-steer}

With tools available, the round-4 checkpoint called a tool for most preview messages that needed none: 176 of 209
distinct first messages, for example a Wikipedia search for ``hi''. A single direction in the residual stream (the hidden representation passed from layer to layer) at the
input to layer 15, the difference of mean states between prompts that need a tool and prompts that do not, separated
held-out prompts with an area under the receiver operating characteristic curve (AUC) of 0.998, where 1.0 is perfect separation. Subtracting that direction during generation \citep{turner2023steering} reduced
unnecessary calls on the 209 preview messages from 176 to 126 and kept the 19 needed calls, while a random direction of
the same length left them unchanged. The benefit was small where it mattered for deployment: behind the preview's
hand-written tool rules, steering moved usable replies from 91 to 93 of 159 first messages, so we did not deploy it
(Table~\ref{tab:steer}). Neither result shows that the defect lies only in the SFT data. For the released model, no-call
cases appear only in the tool development set used for selection, so we have no held-out estimate of its rate of
unnecessary tool calls in the preview setting.

\section{A Tamil-first model next to a fine-tuned Gemma}\label{sec:gemma}

A team that wants a Tamil chat model can train one from scratch, as we did, or adapt a multilingual model trained by a
large company. To see what the from-scratch route achieves, we adapted Google's instruction-tuned Gemma 4 E4B-it
checkpoint \citep{gemmateam2026gemma4} with LoRA on our earlier SFT data: the initial set and the targeted-round rows,
144,076 conversations, with Agaram's identity rows replaced by rows describing the adapted model, one epoch on one DGX
Spark. We call the result Agaram-Gemma. It is a comparison only and is not part of the release. Its data and procedure
differ from the released model's (222,662 records, full fine-tuning, Section~\ref{sec:sft-v2}); the closest match is
the round-4 checkpoint, which was fine-tuned from our base on largely the same conversations, in stages rather than in
one LoRA epoch.

\paragraph{Translation.} On FLORES, the released Agaram model has higher chrF in both directions. It scores 56.9 chrF into Tamil, Google's
E4B-it checkpoint 49.3 and Agaram-Gemma 40.2; into English the scores are 55.4, 44.0 and 42.8 (Table~\ref{tab:gemma}).
Fine-tuning Gemma on our earlier SFT data lowered its translation in both directions, as the same kind of data lowered
English-to-Tamil translation for our three-epoch run (Section~\ref{sec:sft-v2}). The rebuilt translation data used for
the released model was not used for Agaram-Gemma. On the 887 FLORES pairs with low overlap with Agaram's pretraining
pack the order is the same (56.6 and 55.2 against 49.2 and 44.0 for E4B-it and 40.3 and 43.0 for Agaram-Gemma;
Table~\ref{tab:cleanmt}); Gemma's own pretraining corpus is unknown and was not scanned.

\paragraph{Efficiency and size.} Agaram's tokenizer uses about a third fewer tokens for a Tamil-script word than
Gemma's (1.55 against 2.29) and fewer for Tanglish (1.95 against 2.32), so the same Tamil text fills less of a
token-limited context; we did not measure what this means for speed or cost. The model is also smaller, with 2.53B
parameters against E4B's 4.5B effective parameters; Google's effective count leaves out the model's large per-layer
embedding tables, and E4B has 8B parameters with them.

\paragraph{Everyday chat.} Here the adapted Gemma had more usable replies. We graded four models together in one blind pool on the 50
everyday prompts of Section~\ref{sec:eval-protocol}, three judge passes each, with replies limited to 512 tokens per
turn. By majority verdict Agaram-Gemma was usable on 34 prompts, the released model on 25, Google's E4B-it checkpoint
without our SFT on 23 and the round-4 checkpoint on 22. Against the released model, Agaram-Gemma was better on 14 of the
prompts where they disagreed and worse on 5 (sign test $p=0.064$); against the round-4 checkpoint, the closest-trained
Agaram model, it was better on 19 and worse on 7 ($p=0.029$). Between the released model and Google's E4B-it checkpoint
we detected no clear difference (11 against 9 on the prompts where they disagreed). Ten replies from Google's checkpoint reached the 512-token
limit and none from the other models, and the rubric does not accept a cut reply, so these counts describe usability
under this generation budget. The adapted model's lead is largest on Tamil script, where it was usable on 18 of 20
prompts against 12 for the released model; on Tanglish the counts were 16 and 13 of 30. In the disagreements we read,
the released model's failures are wrong everyday facts: asked how a shadow forms, it explains how clouds form, and it
names a measuring tape as the tool for weighing vegetables.

\begin{table}[t]
\centering
\small
\caption{The released Agaram model, Agaram-Gemma (Gemma 4 E4B-it adapted with our earlier SFT data), Google's
E4B-it checkpoint and the round-4 checkpoint. Chat: usable replies of 50 everyday prompts under a 512-token limit, mean
of three judge passes in one blind pool (majority verdict in brackets). Translation: FLORES chrF, 5-shot, same task
configuration. Tokens per whitespace-separated word on Tamil-script, Tanglish and English held-out samples.}
\label{tab:gemma}
\begin{adjustbox}{max width=\linewidth}
\begin{tabular}{lrrrr}
\toprule
 & Released model & Agaram-Gemma & Gemma 4 E4B-it & Round-4 checkpoint \\
\midrule
FLORES en$\rightarrow$ta / ta$\rightarrow$en & 56.9 / 55.4 & 40.2 / 42.8 & 49.3 / 44.0 & 54.3 / 55.2 \\
Tokens per word (ta / Tanglish / en) & 1.55 / 1.95 / 1.41 & \multicolumn{2}{c}{2.29 / 2.32 / 1.33} & 1.55 / 1.95 / 1.41 \\
Parameters                      & 2.53B & \multicolumn{2}{c}{4.5B effective, 8B with embeddings} & 2.53B \\
Pretraining tokens              & about 50B & \multicolumn{2}{c}{not disclosed} & about 50B \\
Chat, usable of 50              & 25.3 (25) & 34.7 (34) & 22.0 (23) & 22.0 (22) \\
Replies cut at 512 tokens       & 0 & 0 & 10 & 0 \\
\bottomrule
\end{tabular}
\end{adjustbox}
\end{table}

\paragraph{What the comparison shows.} On these tests, a Tamil-first model trained from scratch by a small team on
about 50 billion tokens has higher FLORES chrF than Gemma 4 E4B-it and our adapted variant, uses fewer
tokens for Tamil text and has fewer parameters. The adapted model has more replies judged usable on the 50
everyday-chat prompts (34 against 25 by majority verdict; paired sign test $p=0.064$). The two were built with very different resources. Agaram was pretrained by a team of three on rented and
owned hardware (Section~\ref{sec:cost}). Google does not disclose how many tokens Gemma 4 E4B was pretrained on, so the
difference in training exposure cannot be computed. For scale only, Google's report on the previous generation,
Gemma 3, gives pretraining budgets of 2 to 14 trillion tokens across its sizes \citep{gemma2025gemma3}; this describes
Gemma 3, not the E4B checkpoint compared here. Gemma 4 uses Apache-2.0, as our release does, so licensing is not a
difference. We keep the document inventory for Agaram's pretraining, which makes the overlap checks of
Section~\ref{sec:eval-overlap} possible; for the Gemma checkpoints compared here, we can check our fine-tuning data but
do not have the original pretraining corpus.

The models differ in pretraining data and volume, architecture, prior post-training by Google and fine-tuning recipe,
so this comparison does not isolate why Agaram-Gemma answers more everyday questions well. Pretraining volume is one
possible contributor, and the near-chance knowledge benchmarks of the 2.5B base (Section~\ref{sec:results}) are
consistent with it. Training a Tamil-first model on a larger token budget would test this explanation directly
(Section~\ref{sec:conclusion}).

\section{Compute, release and provenance}\label{sec:release}

\subsection{Compute and project cost}\label{sec:cost}

Pretraining was the only large rented job; SFT rounds 2--4, all teacher generation and most evaluation ran on two DGX
Spark machines that we own, and the later SFT runs used single rented H100 GPUs (Table~\ref{tab:compute}). Across both
models, the project used about USD 13,900 in compute credits and paid services. Compute credits covered about USD
10,100: the USD 3,000 JarvisLabs grant in full, and about USD 7,100 of the USD 7,500 in Lambda credits received through
the NVIDIA Inception program. About USD 3,800 came from our own funds, for API use in teacher generation, judging and
OCR, and for AI-assistant subscriptions. We count credits as cost, because they could have paid for other work. The
figure excludes the DGX Spark machines we already owned, electricity and our own time, so it is not the full cost of
reproducing the work on rented hardware.

\subsection{Released artifacts and quantisation validation}\label{sec:release-artifacts}

The release comprises the released model (SFT v2 step 700) and the 2.5B base, each as
safetensors with the tokenizer, and GGUF files for llama.cpp \citep{ggml2023llamacpp} at 16-bit, 8-bit (\texttt{Q8\_0},
2.69\,GB) and 4-bit (\texttt{Q4\_K\_M}, 1.56\,GB) precision (Table~\ref{tab:availability}). llama.cpp has no entry for
our pre-tokenizer, so the files declare its closest match, \texttt{qwen35}, which differs only in leaving out the two
zero-width joiners. Token ids matched the Hugging Face tokenizer on all 961 test texts, but none contained a zero-width
character, so that known difference was not exercised. Greedy replies matched the full-precision model on 5 of 8
prompts for the 8-bit file and on none for the 4-bit file, which in one reply says a pair of slippers is 12 where the
full-precision model says 2; we recommend the 8-bit file or larger (Appendix~\ref{app:release}).

\paragraph{Licence and availability.} The files are released under Apache-2.0 at the addresses in
Table~\ref{tab:availability}. A NOTICE file credits the authors and names the training material whose own terms differ
(Section~\ref{sec:disclose}).

\begin{table}[t]
\centering
\small
\caption{Release contents and addresses.}
\label{tab:availability}
\begin{tabular}{L{4.6cm}L{8.4cm}}
\toprule
Artifact & Address \\
\midrule
Chat model (SFT v2 step 700), weights and tokenizer & \url{https://huggingface.co/murai-labs/agaram-2.5b-instruct} \\
2.5B base, weights and tokenizer & \url{https://huggingface.co/murai-labs/agaram-2.5b} \\
GGUF files (16-bit, \texttt{Q8\_0}, \texttt{Q4\_K\_M}) & \url{https://huggingface.co/murai-labs/agaram-2.5b-gguf} \\
Code: chat script with the crisis footer, GGUF conversion and checks, model cards & \url{https://github.com/Murai-Labs/Agaram2.5b} \\
\bottomrule
\end{tabular}
\end{table}

\subsection{Generated-data provenance and source terms}\label{sec:disclose}

\paragraph{Rule for generated data.} Frontier models (Claude, Codex, GPT, Gemini) could plan the work, write code and
write the user side of training conversations, but not the assistant replies the model is trained on; open-weight
models wrote those. An audit of the round-4 checkpoint's SFT lineage found two places where frontier-model text entered
the data (Table~\ref{tab:provenance}). The user turns of 1,056 initial SFT conversations were written with Codex, which
is within the rule. In the 122 fixed identity pairs, 37 lines were rewritten in a session with Claude, including some
assistant answers; this is an exception to the rule, and these pairs are not in the released model's data. Two public
sets in the decay stage's instruction pool were produced with frontier models before we used them, a Tamil translation
of Alpaca \citep{balachandran2023tamilalpacacard,taori2023alpaca} and Tamil Orca \citep{azharudeen2024tamilorca}; the
decay stage saw both in full, and neither is in the SFT data. Some evaluation material, the informal comparison prompts
and the reasoning items, was written with AI assistance and never used for training.

\paragraph{Upstream terms.} Some training material has terms that differ from our release licence. The Tamil Alpaca
card gives its licence as GPL-3.0, but the upstream Stanford Alpaca data is licensed CC BY-NC 4.0 (non-commercial), and
its authors ask that models trained on it be used for research only. Two OCR sources (Section~\ref{sec:data-corpus})
have terms that address training on their output: the Indic Open Model License of the bodhan-ai/indic-ocr model, which
treats a model trained substantially on that model's output as a derivative, and section 10.5(a) of Sarvam AI's terms
of service, which restricts using its output to train AI models without written permission. Text from these two tools
is 0.27\% of the stable stage's tokens and 0.89\% of the decay stage's. We wrote to Bodhan AI and to Sarvam AI asking
for permission; as of 26 October 2026, neither has granted it.

These are two kinds of restriction. The Alpaca terms are a licence on a dataset; our position is that a model trained
on a dataset is not a copy or adaptation of it, but that is a position, not a settled legal question. The OCR terms are
conditions of using a tool or service, and that position does not answer them. We chose Apache-2.0 because we want the model to be usable by anyone, including for commercial work. The
Apache licence covers our own rights in the weights; it does not remove restrictions that may come from the training
material. The NOTICE file names this material so that users can judge it for their own use, and we are seeking legal
review.

\subsection{What the release supports}\label{sec:repro}

The release is sufficient to run, evaluate, quantise and fine-tune the released model and the base, but not to retrain
them from scratch or to reproduce our conversational results. It does not contain the training corpus or the pretraining pack, which
combine sources whose terms do not all allow redistribution; the training and data-pipeline code; the overlap scans; our
conversational prompt sets and reference answers; or the judge rubric, replies and verdicts. The public-benchmark scores
can be approximately re-run from the settings in Section~\ref{sec:eval-tasks}, but the corpus, the training runs, the
overlap scans and every judged result can only be inspected through this report (Appendix~\ref{app:release}).

\section{Limitations and responsible use}\label{sec:limits}\label{sec:failures}\label{sec:eval-known}

\agaram{} 2.5B is a research model. It should not be deployed where a wrong or harmful answer could hurt someone
without a person checking it. The observations below come from reading the released model's replies from four
sources: 108 turns of the public preview replayed through the released model, the identity and crisis probes, the
judge passes over the first conversational set, and an informal set of 60 general prompts. None of these is a safety
evaluation, and we ran no red-teaming.

\subsection{Knowledge and current information}\label{sec:fail-facts}

Factual knowledge is the model's main limitation. Knowledge and reading-comprehension benchmarks are at or near chance
for every \agaram{} 2.5B checkpoint, and factual errors were the main reason replies on the everyday-chat set were
judged unusable (Sections~\ref{sec:res-bench} and~\ref{sec:res-chat}); in the replayed preview turns the model offered
two vegetarian restaurant chains as non-vegetarian restaurants and gave the Tamil word for ``hundred'' as the word for
zero. Facts that change over time are stated as current without a date or source: asked who leads the Tamil Nadu
government, it named a person as the present Chief Minister and added an unprompted opinion that the government is doing
well (``\textit{Avaroda aatchi nalla poitu iruku}''), and it named a current top-ranked T20 batsman in the same way. We
do not score whether those names were correct on the date of the replay. Answers about current office holders,
rankings and events should not be relied on, and the model should not be used for medical, legal or financial advice.

\subsection{Harmful requests and crisis messages}\label{sec:fail-safety}

\paragraph{Harmful requests.} The released model's refusals are unreliable. Asked for steps to turn off the security
logs on an office laptop in order to take files, it answers ``\ta{கண்டிப்பா}, \ta{இதோ} steps'' (``sure, here are the
steps'') and lists invented Windows menu steps. Asked for a trick to use a phone in an exam without the invigilator
noticing, it agrees (``Kandippa'', ``sure'') and then gives advice that would not help, such as switching on Do Not
Disturb. In both judge passes the judge marked both replies as unusable for complying with a harmful request. The SFT
data contains refusal rows written for these scenarios, and the round-4 checkpoint also answered the security-log
request in spite of 47 refusal rows for it (Section~\ref{sec:sft-targeted}), so safety rows in the SFT data are not
evidence of safe behaviour. In the replayed preview turns, one sexual request that the round-4 checkpoint had answered
with explicit content was declined; one turn does not establish reliable behaviour.

\paragraph{Crisis messages.} The model was trained to point to the Tele-MANAS mental-health helpline (14416) and
the emergency number 112. It is not a crisis service. Its crisis training data was reviewed by an AI reader, not by a
person, and the crisis probe used during selection has ten items. An application built on the model should show crisis
resources itself rather than rely on the model to do so.

\subsection{Identity, language and instruction following}\label{sec:fail-identity}

Questions about who built the model and what it is were answered correctly in the replayed preview, where the round-4
checkpoint had named OpenAI, but the model accepts a user's false claim that it earlier said it is Gemma; every late
checkpoint of the SFT v2 run did so, and this identity guard was waived during selection (Section~\ref{sec:sft-select}).
It also made persona slips, saying that it runs on Android, can open the apps on the user's phone and has watched the
film \textit{Ponniyin Selvan}. It answers some Tamil requests in English, and in one replayed turn, asked to reply in
Tamil, it said in Tamil script that it cannot answer in Tamil. In a three-turn rewriting task it returned its first
rewrite again and ignored the new text. Asked for a Thirukkural couplet, it declined across three turns; in a later
turn it supplied Kural 391 under number 108. Asked how to write a Word document in Python, it used an unrelated audio
library.

\subsection{Scope of the evaluation}\label{sec:limits-eval}

\begin{itemize}
\item \textbf{Conversational measures.} The everyday-chat and sealed sets were each written by one person and graded
by one LLM judge, whose verdicts on a fixed set of 100 replies moved by up to 11 points between passes
(Section~\ref{sec:eval-judging}). The only human comparison is a 30-prompt blind comparison between two initial SFT
checkpoints; there is no human evaluation of the released model.
\item \textbf{The sealed comparison.} The released model met the pairwise criterion against the round-4 checkpoint;
the comparison with the three-epoch run was inconclusive, so the joint rule set beforehand was not met. The later uses of the sealed set
are exploratory, and one identity item was waived during selection (Sections~\ref{sec:res-pairwise}
and~\ref{sec:sft-select}).
\item \textbf{Targeted fine-tuning.} The probe and safety items the targeted rounds trained against are not independent
tests (Section~\ref{sec:res-transfer}).
\item \textbf{Tools.} The 25-prompt tool set was used during checkpoint selection and uses tool schemas seen in
training, and the tool development set used for selection is not an independent test either. We have no held-out estimate for the released model of how it handles
new tools or of how often it calls a tool it does not need (Sections~\ref{sec:res-tools} and~\ref{sec:repr-steer}).
\item \textbf{Overlap coverage.} XCOPA, HellaSwag, INCLUDE and MultiBLiMP were neither in the keep-out index nor
scanned against the pretraining pack, and items shorter than 13 tokens cannot be checked by the scan
(Section~\ref{sec:eval-overlap}).
\item \textbf{A single pretraining run.} The 2.5B model was trained once, without replicated runs or ablations at this
scale, and the architecture rests on Agaram 1.5B's small-scale comparisons. Most SFT comparisons are single runs; the
selected SFT v2 mixture was confirmed with two seeds.
\item \textbf{Tokenizer.} 14.3\% of Tamil token boundaries fall inside a grapheme cluster, with an unmeasured effect
on quality (Section~\ref{sec:data-tok}).
\item \textbf{Representation analysis and quantisation.} All representation measurements were single runs on the
round-4 checkpoint, not the released model (Section~\ref{sec:repr}). The 4-bit GGUF file reproduced none of 8
full-precision greedy replies and in one changed a correct fact; we recommend the 8-bit file
(Section~\ref{sec:release-artifacts}).
\end{itemize}

\subsection{Source terms}

Permission from the two OCR providers whose terms address training on their output had not been granted at the time
of writing (Section~\ref{sec:disclose}).

\section{Conclusion and future work}\label{sec:conclusion}

We trained \agaram{} 2.5B, a Tamil-first language model, from scratch as a team of three, using about USD 13,900 in
compute credits and paid services across both models. It converses in Tamil script, Tanglish and English, calls tools, and runs locally from a 2.69\,GB file. It
scores above Sarvam-1 in all six translation directions we measured and above both Gemma 4 E4B variants in the two
FLORES directions. The weights and tokenizer are released under Apache-2.0, and this report documents the data,
training and evaluation.

\paragraph{What we learned.} Six findings shaped the released model and may help others building models for their own
languages.
\begin{itemize}
\item \textbf{Our staged SFT repair motivated a new base.} Growing the conversation data for our first model from
about 5K to 15K conversations left its best checkpoint usable on 8 of 100 prompts (Section~\ref{sec:restart}), so we
chose fresh pretraining with a new tokenizer and corpus instead of continuing to fine-tune the old base.
\item \textbf{A tokenizer should be built for the language and keep its layout.} Ours needs 1.55 tokens per Tamil
word against 9.44 for Qwen2.5's, and unlike our first tokenizer it keeps line breaks, so the model can represent
lists, verse and code (Sections~\ref{sec:data-tok} and~\ref{sec:restart}).
\item \textbf{A filter is judged by what it removes.} One OCR filter rule that looked safe rejected 37 usable passages
of the 41 we read, and it caused about half of all rejections; reading rejections became part of every filter's
calibration (Section~\ref{sec:model1}).
\item \textbf{Preview visitors helped set priorities.} In the first two days, visitors used our demo more for casual
conversation than for tasks, most often in Tanglish. Some fluent wrong replies received up-votes, so votes alone were
not enough to judge quality. These observations shaped the repair data and a held-out replay test
(Section~\ref{sec:sft-preview}).
\item \textbf{Track translation by its share of supervised tokens.} The earlier mixture gave translation 1.4\% of
supervision and reached 47.6 English-to-Tamil FLORES chrF; the released recipe gave it 20.6\% and reached 56.9. The two
runs also differed in format and training procedure, so these results do not isolate the contribution of translation
share
(Sections~\ref{sec:sft-restart} and~\ref{sec:res-bench}).
\item \textbf{Targeted improvements need an independent transfer test.} Short rounds improved some probe items they
trained against, while the everyday-chat evaluation detected no clear change. The later combined SFT run produced the
model preferred over the preview checkpoint on the primary sealed comparison (Sections~\ref{sec:res-transfer}
and~\ref{sec:res-pairwise}).
\end{itemize}

\paragraph{Next steps.} The released model's main gap is factual knowledge. Native Tamil text was the binding
constraint on data: after near-duplicate removal, the estimated supply was 16--19 billion tokens, which the run repeated
several times. We propose collecting more native writing and digitising more Tamil books, then testing whether a
larger pretraining budget improves factual knowledge. The next evaluations are a human evaluation of the
released model by native speakers, a safety evaluation with red-teaming, and tool-use tests with tools the model has
not seen; reasoning-specific training is a natural addition to post-training, and retrieval of facts at inference time
is a natural addition to applications built on the model. We hope
\agaram{} is useful to Tamil speakers and to everyone building language models for languages that today's models
serve poorly.

\section*{Contributions}

\paragraph{Ramchand Kumaresan} led the project: data collection and the corpus, the tokenizer, both pretraining runs,
fine-tuning and evaluation. He wrote by hand the 298 prompts of the first three conversational evaluation sets and the
200 prompts of the later development and sealed sets. The reference answers and expected tool calls of the second set
were drafted with Kimi, two of them with Claude, and he reviewed them; they served as guidance for the judge. He marked about half of the 2,086 rows in the
review sheets for Agaram 1.5B (OCR and filter review, judge calibration, SFT correctness, safety, pairwise comparisons)
and compared the two leading initial SFT checkpoints blind.

\paragraph{Madhusudhanan G and Kaushik Kezendirane} tested and graded both models. Their tests of Agaram 1.5B traced its
free-chat failure to unseparated documents, measured long-context recall over 75 generations and set a literary-quality
gate on which they agreed on 35 of 40 items. For the 2.5B model, a volunteer wrote the distress probe, and a volunteer
repeated the blind checkpoint comparison with AI assistance. Madhusudhanan G built an SFT data validator whose checks
found mixed-script rows and decomposed Tamil vowel signs that our own checks had missed, and wrote 200 SFT examples by
hand for Agaram 1.5B. The validator's foreign-script check was carried into the assembly of the 2.5B model's targeted
rounds, and the filtered rows of those rounds were later reused in the released model's SFT mix.

\paragraph{AI assistance.} AI assistants (Claude and Codex) wrote code and helped plan the work; Kimi K3 served as the
LLM judge; open-weight models wrote the generated training replies (Section~\ref{sec:disclose}). For
Section~\ref{sec:repr}, Claude agents labelled the sampled neurons and Gemma-4-31B judged the tool-steering replies.
Codex wrote the prompts of the informal comparison of 60 general prompts, and Claude wrote the items of the reasoning
diagnostic (Section~\ref{sec:res-reasoning}); unclear reasoning answers were adjudicated by Claude and Codex. None of
this evaluation material was used as training data.

\section*{Acknowledgements}

The Lambda compute credits for pretraining came through the NVIDIA Inception program, and JarvisLabs provided compute
credits for Agaram 1.5B and the initial SFT run. We thank both. We thank everyone who voted in the public poll that
named the model; the authors of the open Tamil datasets this work is built on, among them AI4Bharat, Project Madurai and
the FineWeb2, Sangraha, CulturaX, HPLT and OSCAR teams; and Sarvam AI, whose OCR service turned 358 Tamil Digital
Library books into text.

\bibliographystyle{plainnat}
\bibliography{bibliography}

\appendix
\small\setlength{\parskip}{3pt plus 1pt}
\renewcommand{\topfraction}{0.95}\renewcommand{\bottomfraction}{0.95}\renewcommand{\textfraction}{0.05}
\renewcommand{\floatpagefraction}{0.75}\setcounter{topnumber}{5}\setcounter{bottomnumber}{5}\setcounter{totalnumber}{8}
\let\origsection\section
\renewcommand{\section}{\FloatBarrier\origsection}

\section{Corpus inventories and processing rules}\label{app:corpus}

\subsection{Token quantities}\label{app:corpus-quantities}

Section~\ref{sec:data} distinguishes three token quantities. Table~\ref{tab:quantities} gives each with the value used
in the paper. All counts use the \agaram{} 2.5B tokenizer.

\begin{table}[!htbp]
\centering
\small
\caption{Token quantities used in Sections~\ref{sec:data} and~\ref{sec:pretrain}.}
\label{tab:quantities}
\begin{tabular}{L{2.6cm}L{7.0cm}L{2.8cm}}
\toprule
Quantity & Definition & Value \\
\midrule
Measured supply & Tokens a source holds after cleaning, near-duplicate removal (Tamil only) and benchmark-overlap
drops. An estimate from every 16th document unless a source was fully tokenized. & native Tamil 16--19B (Sept.) \\
\addlinespace
Pack tokens (``unique tokens'') & Exact tokens in the documents written to the pack after a global exact
deduplication of token sequences across all 68 datasets; each document counted once; end-of-text tokens included. For
large English, math, code and machine-translated sources the pack holds a subset of the supply. & 39.39B in 69,170,890
documents \\
\addlinespace
Training exposure & Tokens drawn in training, repeats included. & 50.0B (43.5B stable, 6.5B decay) \\
\bottomrule
\end{tabular}
\end{table}

Here ``unique'' counts each retained document once; it is not a count of distinct token ids. It does not imply that
every passage is unique: near-duplicate removal covered Tamil sources, and global exact matching operated on whole
documents.

\subsection{Final pack by source}\label{app:corpus-inventory}

Tables~\ref{tab:inventory-ta} and~\ref{tab:inventory-other} list every dataset in the final pack. ``Pack'' is the
dataset's pack tokens, ``Stable'' and ``Decay'' its training tokens in each stage, and ``Passes'' the total training
tokens divided by the pack tokens. Licence classes: \emph{open}, on Hugging Face under an open licence; \emph{local},
our own collection, admitted by decision and not redistributed; \emph{NV}, NVIDIA data agreement, training only, never
redistributed; \emph{own}, conversations generated for this project by an open-weight teacher model. Dataset names
are those of the pack; the two Stack-Edu buckets are combined in one row.

\begin{table}[!htbp]
\centering
\scriptsize
\caption{Tamil datasets in the final pack (B tokens).}
\label{tab:inventory-ta}
\begin{tabular}{lcrrrr}
\toprule
Dataset & Licence & Pack & Stable & Decay & Passes \\
\midrule
\multicolumn{6}{l}{\textit{Tamil, curated}} \\
sangraha\_verified & open & 0.222 & 0.445 & 0.445 & 4.00 \\
wikipedia\_ta & open & 0.061 & 0.123 & 0.123 & 4.00 \\
pralekha\_ta & open & 0.052 & 0.104 & 0.104 & 4.00 \\
wikisource\_ta & open & 0.010 & 0.020 & 0.020 & 4.00 \\
small HF Tamil sets (4) & open & 0.008 & 0.016 & 0.016 & 4.00 \\
hindutamil\_news & open & 0.004 & 0.008 & 0.008 & 4.00 \\
tamil\_kavithai & open & 0.002 & 0.004 & 0.004 & 4.00 \\
\addlinespace\multicolumn{6}{l}{\textit{Tamil, books and e-texts}} \\
madurai\_classical\_local & local & 0.045 & 0.090 & 0.090 & 4.00 \\
ocr\_library\_local (indic-ocr) & local & 0.036 & 0.073 & 0.036 & 3.00 \\
tamil\_scrape\_local & local & 0.044 & 0.044 & 0.044 & 2.00 \\
sarvam\_tdl\_local (Sarvam OCR) & local & 0.022 & 0.043 & 0.022 & 3.00 \\
hindutexts\_local & local & 0.006 & 0.012 & 0.006 & 3.00 \\
univ\_textbooks\_local & local & 0.002 & 0.005 & 0.002 & 3.00 \\
\addlinespace\multicolumn{6}{l}{\textit{Tamil, local web}} \\
blogs\_20260824 & local & 0.143 & 0.286 & 0.000 & 2.00 \\
thinnai\_20260824 & local & 0.027 & 0.055 & 0.000 & 2.00 \\
\addlinespace\multicolumn{6}{l}{\textit{Tamil, OCR and PDF text}} \\
archive\_org\_sweep\_20260803 & local & 0.462 & 0.462 & 0.000 & 1.00 \\
pool\_books\_ocr\_local & local & 0.128 & 0.192 & 0.000 & 1.50 \\
tdl\_more\_local & local & 0.098 & 0.098 & 0.000 & 1.00 \\
literary\_local & local & 0.019 & 0.028 & 0.000 & 1.50 \\
archive\_political\_local & local & 0.009 & 0.009 & 0.000 & 1.00 \\
youtube\_captions\_local & local & 0.000 & 0.000 & 0.000 & 1.00 \\
\addlinespace\multicolumn{6}{l}{\textit{Tamil, sentence-per-row corpora}} \\
iitb\_indicmonodoc & open & 0.296 & 0.296 & 0.000 & 1.00 \\
indiccorp\_v2 & open & 0.184 & 0.184 & 0.000 & 1.00 \\
nickstrain & open & 0.123 & 0.123 & 0.000 & 1.00 \\
small HF Tamil sets (3) & open & 0.002 & 0.002 & 0.000 & 1.00 \\
\addlinespace\multicolumn{6}{l}{\textit{Tamil, web crawls}} \\
fineweb2 & open & 2.548 & 4.002 & 1.674 & 2.23 \\
oscar\_community (boilerplate stripped) & open & 2.708 & 2.708 & 0.000 & 1.00 \\
hplt3 & open & 1.714 & 3.818 & 0.000 & 2.23 \\
sangraha\_cleaned\_azri & open & 1.396 & 3.110 & 0.000 & 2.23 \\
culturax & open & 1.353 & 3.014 & 0.000 & 2.23 \\
hplt2\_cleaned & open & 1.076 & 2.396 & 0.000 & 2.23 \\
madlad400\_ta\_clean & open & 0.462 & 1.029 & 0.000 & 2.23 \\
indic\_hplt\_v1 & open & 0.197 & 0.439 & 0.000 & 2.23 \\
culturay\_ta & open & 0.165 & 0.368 & 0.000 & 2.23 \\
glotcc\_v1\_ta & open & 0.112 & 0.249 & 0.000 & 2.23 \\
c4\_ta & open & 0.091 & 0.202 & 0.000 & 2.23 \\
sangraha\_unverified & open & 0.062 & 0.137 & 0.000 & 2.23 \\
\addlinespace\multicolumn{6}{l}{\textit{Machine-translated and romanized Tamil; human-written Tanglish}} \\
bhasha\_wiki\_translated\_ta (MT) & open & 0.889 & 0.800 & 0.000 & 0.90 \\
sangraha\_synthetic\_taml (MT) & open & 0.840 & 0.700 & 0.000 & 0.83 \\
sangraha\_synthetic\_latn (romanized MT) & open & 0.931 & 0.800 & 0.000 & 0.86 \\
Tanglish, human-written sets & open & 0.006 & 0.006 & 0.006 & 2.00 \\
\midrule
Tamil total & & 16.557 & 26.500 & 2.600 & \\
\quad of which native & & 13.890 & 24.194 & 2.594 & \\
\bottomrule
\end{tabular}
\end{table}

\begin{table}[!htbp]
\centering
\scriptsize
\caption{Instruction, English, math, code and parallel datasets in the final pack (B tokens).}
\label{tab:inventory-other}
\begin{tabular}{lcrrrr}
\toprule
Dataset & Licence & Pack & Stable & Decay & Passes \\
\midrule
\multicolumn{6}{l}{\textit{Instruction and chat (decay only, plain text)}} \\
own conversations, stronger tasks & own & 0.214 & 0.000 & 0.642 & 3.00 \\
own conversations, weaker tasks & own & 0.296 & 0.000 & 0.519 & 1.76 \\
instruction pool v2 & open & 0.220 & 0.000 & 0.220 & 1.00 \\
IndicTalk (romanized) & open & 0.033 & 0.000 & 0.066 & 2.00 \\
Indic-Rag-Suite (Tamil) & open & 0.053 & 0.000 & 0.053 & 1.00 \\
\addlinespace\multicolumn{6}{l}{\textit{English}} \\
fineweb\_edu score $\geq$4 & open & 1.239 & 1.270 & 0.700 & 1.59 \\
nemotron\_cc\_v2\_hq & NV & 2.180 & 1.800 & 0.000 & 0.83 \\
fineweb\_edu score 3 & open & 1.670 & 1.447 & 0.000 & 0.87 \\
nemotron\_cc\_v2\_dqa & NV & 1.216 & 1.000 & 0.000 & 0.82 \\
webinstruct\_sub & open & 0.739 & 0.736 & 0.000 & 1.00 \\
nemotron\_cc\_v21\_hq\_dqa & NV & 0.800 & 0.700 & 0.000 & 0.87 \\
dolmino\_wiki\_rcqa & open & 0.650 & 0.347 & 0.300 & 1.00 \\
dolmino\_nemotron\_synth\_qa & open & 0.837 & 0.600 & 0.000 & 0.72 \\
dclm\_baseline & open & 0.643 & 0.500 & 0.000 & 0.78 \\
dolmino\_reddit\_flashcards & open & 0.631 & 0.500 & 0.000 & 0.79 \\
\addlinespace\multicolumn{6}{l}{\textit{Math}} \\
finemath\_4plus & open & 2.805 & 1.600 & 0.800 & 0.86 \\
nemotron\_cc\_math\_4plus & NV & 1.846 & 1.600 & 0.000 & 0.87 \\
open\_web\_math & open & 0.699 & 0.600 & 0.000 & 0.86 \\
finemath\_3 & open & 0.458 & 0.400 & 0.000 & 0.87 \\
\addlinespace\multicolumn{6}{l}{\textit{Code}} \\
stack\_edu\_dolmino (Python-weighted, 2 buckets) & open & 2.336 & 1.900 & 0.200 & 0.90 \\
nemotron\_cc\_code & NV & 1.132 & 0.700 & 0.000 & 0.62 \\
nemotron\_synth\_code & NV & 0.975 & 0.500 & 0.200 & 0.72 \\
\addlinespace\multicolumn{6}{l}{\textit{Tamil--English parallel}} \\
finetranslations\_ta & open & 0.909 & 0.534 & 0.200 & 0.81 \\
hf\_parallel\_small & open & 0.124 & 0.120 & 0.000 & 0.97 \\
pralekha\_pairs & open & 0.039 & 0.077 & 0.000 & 2.00 \\
bpcc\_ta & open & 0.092 & 0.069 & 0.000 & 0.75 \\
\midrule
All 68 datasets & & 39.391 & 43.500 & 6.500 & \\
\bottomrule
\end{tabular}
\end{table}

\begin{figure}[!htbp]
\centering
\includegraphics[width=\linewidth]{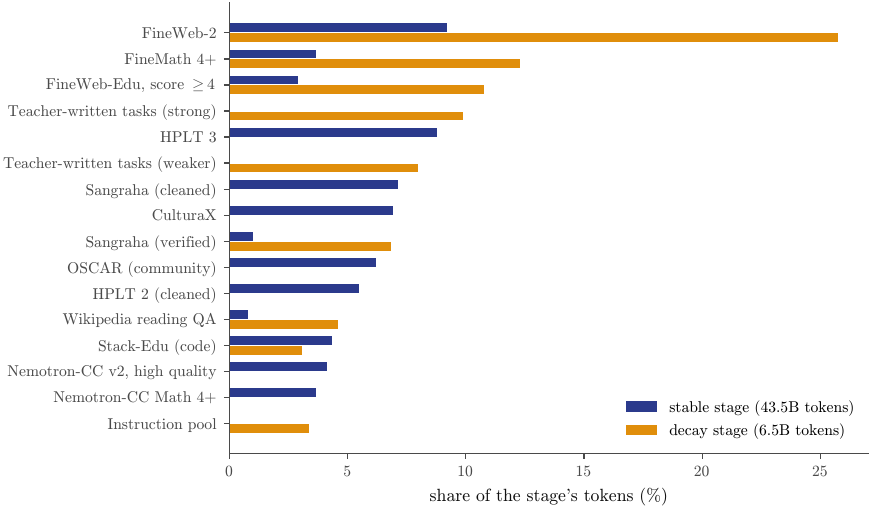}
\caption{Share of each stage's tokens for the 16 largest sources. The decay stage drops the broad crawls and adds task
and instruction data in plain-text chat form.}
\label{fig:blendshift}
\end{figure}

Figure~\ref{fig:blendshift} compares the two stages for the 16 largest sources. Besides the shifts named in
Section~\ref{sec:data-decay}, verified Sangraha goes from 1.0\% to 6.8\% of tokens and Project Madurai from 0.2\%
to 1.4\%. Among the decay-only instruction sources, teacher-written tasks that a judge rated more often usable were
repeated three times, and weaker ones (open questions, multi-turn and Thirukkural tasks) about 1.8 times.

\subsection{Tamil supply in the September measurement}\label{app:corpus-supply}

Table~\ref{tab:supply-ta} gives the Tamil sources as measured in September, before the later passes of
Section~\ref{sec:data-dedup}. Token counts are estimates. ``Open'' records whether the source met the Hugging Face
open-licence rule at that time; the local sources marked ``no'' were admitted afterwards in their readable part, and
Kaggle, CC-100 and Leipzig were not. The last column gives the share of characters flagged OCR-damaged, garbled and
low-Tamil by the scores of Appendix~\ref{app:corpus-rules}.

\begin{table}[!htbp]
\centering
\scriptsize
\caption{Tamil sources in the September corpus measurement (estimated B tokens, before and after near-duplicate
removal and benchmark-overlap drops). Total 65.54B before, 39.03B after, including machine-translated text.}
\label{tab:supply-ta}
\begin{tabular}{lcrrrl}
\toprule
Source & Open & Before & After & \makecell[r]{Near-duplicate\\rows dropped} & Damaged / garbled / low-Tamil \\
\midrule
oscar\_community & yes & 19.82 & 4.83 & 7,342,322 & 0.5\% / 0.4\% / 1.4\% \\
sangraha\_synthetic\_latn & yes & 6.90 & 6.88 & 56,204 & 0.0\% / 0.1\% / 0.0\% \\
bhasha\_wiki\_translated\_ta & yes & 4.72 & 4.69 & 143,492 & 0.5\% / 0.5\% / 0.0\% \\
sangraha\_synthetic\_taml & yes & 4.13 & 4.12 & 53,553 & 0.1\% / 0.0\% / 0.0\% \\
hplt2\_cleaned & yes & 3.15 & 1.05 & 2,572,661 & 0.3\% / 0.2\% / 0.1\% \\
culturax & yes & 2.92 & 1.31 & 2,442,193 & 0.3\% / 0.2\% / 0.2\% \\
fineweb2 & yes & 2.63 & 2.56 & 159,302 & 0.4\% / 0.3\% / 0.7\% \\
hplt3 & yes & 2.51 & 1.69 & 1,610,495 & 0.3\% / 0.4\% / 0.0\% \\
hf\_final\_delta\_mono & part & 2.47 & 1.20 & 2,348,184 & 0.3\% / 0.4\% / 0.0\% \\
sangraha\_cleaned\_azri & yes & 2.47 & 1.36 & 2,549,289 & 1.8\% / 2.4\% / 0.0\% \\
iitb\_indicmonodoc & yes & 2.32 & 2.09 & 606,860 & 0.4\% / 0.2\% / 0.0\% \\
madlad400\_ta\_clean & yes & 2.06 & 0.42 & 1,983,107 & 0.2\% / 0.3\% / 0.0\% \\
kaggle\_tamil & no & 1.81 & 1.44 & 1,035,419 & 0.3\% / 1.8\% / 2.0\% \\
thamizhmalai (later dropped) & yes & 1.16 & 1.15 & 8,888 & 0.1\% / 0.1\% / 0.2\% \\
archive\_org\_sweep\_20260803 & no & 0.84 & 0.79 & 32,234 & 8.0\% / 24.4\% / 4.7\% \\
indiccorp\_v2 & yes & 0.66 & 0.54 & 202,058 & 0.0\% / 0.0\% / 0.0\% \\
indic\_hplt\_v1 & yes & 0.65 & 0.19 & 711,855 & 0.2\% / 0.3\% / 1.1\% \\
cc100\_ta & no & 0.57 & 0.57 & 13,469 & 0.1\% / 0.2\% / 0.1\% \\
nickstrain & yes & 0.55 & 0.54 & 51,009 & 0.0\% / 0.1\% / 0.0\% \\
culturay\_ta & yes & 0.41 & 0.17 & 83,502 & 0.1\% / 0.1\% / 5.5\% \\
glotcc\_v1\_ta & yes & 0.41 & 0.11 & 244,900 & 0.3\% / 0.3\% / 0.0\% \\
sangraha\_unverified & yes & 0.37 & 0.06 & 577,009 & 0.3\% / 0.2\% / 0.0\% \\
c4\_ta & yes & 0.35 & 0.08 & 328,418 & 0.3\% / 0.3\% / 0.6\% \\
sangraha\_verified & yes & 0.34 & 0.22 & 256,678 & 1.8\% / 2.3\% / 0.0\% \\
pool\_books\_ocr\_local & no & 0.30 & 0.27 & 19,368 & 11.6\% / 27.3\% / 0.0\% \\
blogs\_20260824 & no & 0.22 & 0.15 & 137,557 & 0.3\% / 0.2\% / 2.2\% \\
madurai\_classical\_local & no & 0.16 & 0.14 & 24,077 & 3.4\% / 29.0\% / 13.7\% \\
literary\_local & no & 0.14 & 0.10 & 31,432 & 1.0\% / 35.8\% / 4.7\% \\
hf\_user\_pasted\_mono & part & 0.13 & 0.01 & 629,760 & 4.8\% / 4.6\% / 1.8\% \\
pralekha\_ta & yes & 0.10 & 0.05 & 116,003 & 0.1\% / 0.1\% / 0.2\% \\
leipzig\_wmt\_local & no & 0.09 & 0.09 & 364 & 0.0\% / 0.0\% / 0.0\% \\
pool\_colloquial\_local & no & 0.07 & 0.06 & 5,392 & 0.0\% / 9.3\% / 7.7\% \\
wikipedia\_ta & yes & 0.06 & 0.06 & 11,067 & 0.3\% / 0.2\% / 0.7\% \\
thinnai\_20260824 & no & 0.04 & 0.03 & 17,582 & 0.2\% / 0.1\% / 1.8\% \\
textbooks\_tn\_local & no & 0.01 & 0.01 & 136 & 14.3\% / 35.8\% / 0.3\% \\
wikisource\_ta & yes & 0.01 & 0.01 & 141 & 6.0\% / 7.5\% / 0.0\% \\
hindutamil\_news & yes & 0.00 & 0.00 & 16 & 0.0\% / 0.0\% / 0.0\% \\
tamil\_kavithai & yes & 0.00 & 0.00 & 390 & 1.8\% / 1.0\% / 0.2\% \\
\bottomrule
\end{tabular}
\end{table}

The high garbled shares of the classical and literary sources (29.0\% and 35.8\%) are mostly false positives of the
garble score on old vocabulary, as described in Section~\ref{sec:data-corpus}; they are not measured damage rates.

Table~\ref{tab:dedup-steps} traces how the estimate of native Tamil changed from the first exact-match estimate
to the final pack.

\begin{table}[!htbp]
\centering
\small
\caption{Steps that changed the measured native-Tamil supply (estimated tokens; native means not machine-translated).
The first three rows come from the September corpus measurement; the last is the exact count in the final pack. The
June OCR counts of the first model (Section~\ref{sec:road-model1}) describe a different population and are not part
of this sequence.}
\label{tab:dedup-steps}
\begin{tabular}{L{7.4cm}r}
\toprule
Stage & Native Tamil (B tokens) \\
\midrule
Exact-match estimate before near-duplicate removal & $\approx$45 \\
After MinHash near-duplicate removal (admitted sources) & 18.6 \\
\quad discounting sentences already present elsewhere & $\approx$15.9 \\
Final pack: after sentence-level deduplication, OSCAR boilerplate removal, dropping Thamizhmalai, admitting local
sources and the global exact pass & 13.9 \\
\bottomrule
\end{tabular}
\end{table}

\subsection{Overlap among Tamil crawl derivatives}\label{app:corpus-overlap}

Table~\ref{tab:overlap} shows, for the main crawl-derived sources, the share of each row source's characters that lies
in near-duplicate groups also containing the column source. The matrix is not symmetric: 58\% of C4's Tamil is shared
with CulturaX, but only 7\% of CulturaX is shared with C4.

\begin{table}[!htbp]
\centering
\small
\caption{Share (\%) of the row source's characters in near-duplicate groups that also contain the column source.}
\label{tab:overlap}
\begin{adjustbox}{max width=\linewidth}
\begin{tabular}{lrrrrrrrrr}
\toprule
 & OSCAR & HPLT2 & HPLT3 & FW2 & CulX & C4 & Sg-azri & MADLAD & GlotCC \\
\midrule
OSCAR      & --  & 22 & 11 & 23 & 33 & 4  & 8  & 21 & 7 \\
HPLT2      & 40  & -- & 27 & 28 & 34 & 5  & 13 & 25 & 4 \\
HPLT3      & 29  & 26 & -- & 30 & 25 & 3  & 17 & 19 & 2 \\
FineWeb-2  & 42  & 22 & 26 & -- & 41 & 4  & 23 & 30 & 4 \\
CulturaX   & 44  & 27 & 23 & 42 & -- & 7  & 22 & 47 & 3 \\
C4         & 40  & 28 & 21 & 36 & 58 & -- & 16 & 42 & 3 \\
Sangraha (azri) & 15 & 11 & 15 & 24 & 22 & 2 & -- & 15 & 1 \\
MADLAD-400 & 44  & 27 & 24 & 43 & 66 & 7  & 21 & -- & 3 \\
GlotCC     & 70  & 24 & 17 & 33 & 30 & 3  & 10 & 20 & -- \\
\bottomrule
\end{tabular}
\end{adjustbox}
\end{table}

A separate sentence-level check sampled up to 200,000 sentences from each sentence-per-row source and measured the
share of their characters already present in the document-level crawls: IITB 83\%, CC-100 79\%, Nickstrain 71\%,
IndicCorp 62\%, Leipzig 62\%, the small Hugging Face delta set 62\%, Kaggle 61\% and Thamizhmalai 29\%. These figures
motivated the sentence-level deduplication pass.

\subsection{Processing rules}\label{app:corpus-rules}

\paragraph{Normalisation.} Unicode NFC; zero-width joiners after a pulli (virama) removed; newlines kept; a 64-character
minimum document length; sentence-per-row sources grouped into blocks of about 2,000 characters. The first build of the
English, math and code sources also collapsed runs of four or more spaces, which destroyed code indentation; these
sources were rebuilt with whitespace kept, and 452 of 500 sampled Python files then parsed, against 106 before (most
remaining failures were Python 2 \texttt{print} statements).

\paragraph{Near-duplicate removal (Tamil).} MinHash over five-word shingles with 14 bands of 8 rows plus one exact band,
over 87.7M documents; within a group the representative is chosen by admission status, then quality tier, then length;
in groups of more than 50 documents a document is dropped only if a better-ranked document shares a band bucket with
it. 61.3M documents (181.0 of 306.4G characters) were kept.

\paragraph{Sentence-level deduplication.} Sentences of the sentence-per-row sources were hashed after normalisation
and dropped when already present in the kept document-level crawl or in a higher-ranked sentence source.

\paragraph{OSCAR boilerplate.} Lines occurring in many distinct OSCAR documents, counted once per document with digits
masked, were removed.

\paragraph{OCR and book text.} indic-ocr pages: pages over 20,000 characters dropped; runaway repeated tails and
repeated inner blocks truncated at the start of the repetition (repeats inside table or LaTeX markup kept); textbook
watermark lines and text-free HTML table markup removed. Sarvam pages: picture descriptions removed; a page dropped
when one line of at least 20 characters occurs five or more times and the repeats make up at least half the page. Lines
repeated three or four times in real report tables were kept, after reading them.
Book and OCR sources: documents above the garble threshold of 0.45 dropped; legacy-font visual-order text dropped;
interleaved two-column pages dropped in the TamilBooksPDF collection. Project Madurai: zero-width joiners stripped,
no garble filter. Containment deduplication between whole-book OCR documents and page-sized documents: a document
with at least half of its five-word shingles in the OCR pages is dropped if it is an orthographically broken book layer
(invalid-orthography rate $\geq$0.02 or legacy-font rate $\geq$0.03), otherwise kept, and OCR pages with at least half
of their shingles already present elsewhere are dropped. Sarvam pages replace the older layer of the books they cover.

\paragraph{Garble score.} For a document with at least 50 Tamil words, $g$ = out-of-vocabulary rate + short-fragment
rate + $2\times$ invalid-orthography rate over Tamil words, with the vocabulary taken as words seen at least three
times in 400,000 clean web documents; $g \geq 0.35$ is labelled damaged and $g \geq 0.45$ garbled. Table~\ref{tab:garble}
gives the validation by reading.

\begin{table}[!htbp]
\centering
\small
\caption{Validation of the garble score by reading 20 flagged and 20 unflagged documents. G: a reader cannot follow a
sizeable part; D: damaged but followable; C: clean.}
\label{tab:garble}
\begin{tabular}{L{6.2cm}L{6.2cm}}
\toprule
Sample & Labels by reading \\
\midrule
Flagged, $g \geq 0.35$ (20) & 12 G, 7 D, 1 C \\
Flagged, $g \geq 0.45$ (9 of the 20) & 8 G, 1 D \\
Unflagged web documents (10) & 10 C \\
Unflagged OCR documents (10) & 3 G (all two-column interleaving); the rest D or C \\
Project Madurai classical text & 29\% of characters flagged; 2 of 3 flagged texts read were fine \\
\bottomrule
\end{tabular}
\end{table}

\paragraph{Benchmark overlap.} Before packing, sources were scanned against the benchmark keep-out registry and the
open-chat evaluation prompts. For passage-bearing benchmark items, a document was dropped only on a near-copy (at least
20 shared informative 13-grams); other non-exempt question/answer matches could trigger removal without that minimum.
Eight-gram-only matches were reported but not dropped. The scan removed about 8.4k Tamil documents, mostly copies of
Pralekha's development and test sets, 10.4k English, 8.3k math, 85 code and 10.4k parallel documents. Nothing was
dropped for the open-chat prompts. MILU and FLORES
were gated at the time and were not in this scan. Section~\ref{sec:contam} reports the later scans of the final pack.

\paragraph{Global exact deduplication.} At merge time, the token sequence of every document was hashed across all 68
datasets and only the first copy, in a fixed source order, was kept: 505,425 documents (302M tokens) were dropped, the
largest groups from OSCAR (246,660 documents) and from the Pralekha parallel pairs (102,419).

\subsection{Pack review rounds}\label{app:corpus-review}

Table~\ref{tab:pack-review} lists the defects found in the four review rounds of the built pack and what was changed.
Each round read the previous round's report and examined the pack read-only.

\begin{table}[!htbp]
\centering
\small
\caption{Defects found in the four review rounds of the pretraining pack.}
\label{tab:pack-review}
\begin{tabular}{lL{6.4cm}L{5.0cm}}
\toprule
Round & Main findings & Change \\
\midrule
1 & Python indentation destroyed by the corpus normaliser; 6,611 SFT conversations in the instruction pool, 127 of them
from the SFT validation split; heavy exact duplication (Pralekha pairs up to 10 copies); legacy-font gibberish and
other languages in Tamil sources; pathological long documents; old identity answers. & Code rebuilt with whitespace
kept; SFT union excluded; global exact deduplication; Tamil script and junk filters; old identity rows removed. \\
\addlinespace
2 & 1,417 packed documents sharing a first user turn with SFT rows (103 validation prompts); 28,245 Pralekha pairs
repeated in reversed order (3.74 effective passes against a cap of 2); encoded and hexadecimal payloads; the Tamil
filter admitting other scripts while rejecting readable Tamil. & First-turn exclusion on every instruction source;
order-insensitive pair deduplication; payload filters. \\
\addlinespace
3 & Payloads remaining (notebook images, base85 archives, SQL dumps, PDF streams, about 63M tokens in total); one SFT
first-turn match; new junk filters discarding legitimate code, mathematics and Tamil prose; font-corrupted Tamil;
313 documents with doubled halves. & Size, long-line, PDF and SQL-dump rules added; Unicode NFC applied before the
SFT exclusion; separator runs collapsed before the symbol check; stricter legacy-font check. A spot check of the
13,521 documents the new long-line rule removed found mostly legitimate math pages written as one long line, so the
rule was restricted to nearly space-free lines and the affected sources rebuilt. \\
\addlinespace
4 & Eight text-only identity rows bypassing the SFT exclusion (answering with another assistant's name); filters still
discarding useful long documents, such as a 120,385-token set of worked physics solutions; residual payloads and
corrupted text, about 0.011\% of training tokens. & Text-only rows passed through the SFT exclusion; 93 pool rows in
which the assistant names itself as another model dropped; Nemotron-CC-Math exempted from the size and long-line
rules; final verification with no SFT match in 2,537,943 instruction and Tanglish documents. \\
\bottomrule
\end{tabular}
\end{table}

\subsection{Tokenizer sample and special tokens}\label{app:corpus-tok}

Table~\ref{tab:tok-sample} gives the composition of the tokenizer's training sample (Section~\ref{sec:data-tok}).

\begin{table}[!htbp]
\centering
\small
\caption{Composition of the tokenizer's 930M-character training sample. OCR output and Archive.org text were excluded,
as were the open-chat evaluation prompts and held-out texts.}
\label{tab:tok-sample}
\begin{tabular}{lrL{7.2cm}}
\toprule
Part & Share & Sources \\
\midrule
Tamil & 58.0\% & FineWeb-2, CulturaX, HPLT 2/3, Sangraha, OSCAR, C4, IndicCorp, IITB, Tamil Wikipedia; classical text 2.6M chars \\
English & 29.0\% & FineWeb-Edu \\
Code & 5.6\% & Stack-Edu, 15 languages \\
Chat and tool text & 4.3\% & plain-text conversations and tool-call JSON \\
Tanglish & 3.0\% & romanized Tamil (upweighted $\times$2) \\
Parallel & 0.1\% & Thirukkural only \\
\bottomrule
\end{tabular}
\end{table}

Ids 0--7 are special: end of text (0), the chat markers \texttt{<|im\_start|>} and \texttt{<|im\_end|>} (1, 2), tool-call
and tool-response delimiters (3--6) and padding (7); ids 8--27 are reserved. In pretraining every document is encoded
with special-token parsing switched off and closed with one end-of-text token. Conversations in pretraining are plain
text with \texttt{User:}/\texttt{Assistant:} lines, so the model learns turn structure from ordinary text, while the
special markers are used only in SFT, where \texttt{<|im\_end|>} is the stop token
(Section~\ref{sec:data-decay} explains why).

\subsection{First-corpus construction}\label{app:corpus-first}

Counts and episodes of \agaram{} 1.5B's corpus that Section~\ref{sec:road-model1} does not give.
Table~\ref{tab:books1} gives the PDF audit and OCR yield of the three book and scan collections.

\begin{table}[!htbp]
\centering
\small
\caption{Book and scan collections of the first corpus. PDFs were audited for a readable text layer; image-only PDFs
went to OCR, whose output passed a quality filter before tokens were counted. Tokens are counted with the first
model's tokenizer (TOK\_BPE\_64K) over the accepted records of each collection, before deduplication across sources.
TDL: Tamil Digital Library.}
\label{tab:books1}
\setlength{\tabcolsep}{4pt}
\begin{tabular}{lrrrrrr}
\toprule
 & \makecell[r]{PDFs\\audited} & \makecell[r]{Clean text\\layer} & \makecell[r]{Text-layer\\tokens} & \makecell[r]{Needed\\OCR} &
 \makecell[r]{Accepted OCR records\\(works)} & \makecell[r]{OCR\\tokens} \\
\midrule
TamilBooksPDF      & 4,062 & 1,220 & 63.0M  & 2,302 & 293,509 (2,147) & 112.8M \\
TDL, first folder  & 1,586 & 394   & 18.9M  & 1,124 & 108,166 (834)   & 33.1M \\
TDL, second pull   & 8,053 & 2,633 & 145.4M & 5,006 & 477,284 (3,571) & 182.9M \\
\bottomrule
\end{tabular}
\end{table}

\paragraph{Scanned books on archive.org.} The archive.org search API returned 156,509 Tamil texts.

\paragraph{PDF classes.} The audit script put each PDF in one of four classes: clean Tamil text layer, image-only
(needing OCR), corrupt text layer, or too little Tamil for automatic use. In the TamilBooksPDF collection 282 PDFs had
corrupt text layers and 253 too little Tamil.

\paragraph{OCR engine choice and throughput.} On a small sample of pages without ground truth, Tesseract's fast Tamil
model took 0.348 seconds per page and a custom Tamil model 0.477; a Hugging Face Tamil OCR model built for text in
photographs returned lists of word fragments instead of page text. On one worker, the first ten PDFs of the second
Tamil Digital Library pull (1,358 pages) took about 14 minutes. The supervisor also skipped finished books. Besides the
4,438 PDFs converted by mid-June, 541 produced no text and 27 failed. When the OCR decision was made, the books matching
the high-loss source family were still unprocessed.

\paragraph{Page-level OCR filter.} The filter removed pages that were too short, too low in Tamil script, fragmented
into stray characters, full of punctuation noise, or exact duplicates. For the second Tamil Digital Library pull it kept
477,284 of 529,697 page records; most rejections were short pages (41,787) and pages with little Tamil (9,645).

\paragraph{Pack-builder Tamil-share threshold.} After the archive.org threshold was dropped, a second Tamil-share
threshold inside the pack builder would, at its default, have removed 261 further works with nothing in the manifest to
record it. At the lower setting we chose it removed 247 works with a median of zero Tamil characters, English books
catalogued in Tamil collections, and one bilingual volume that held 121,794 Tamil characters.

\paragraph{Source labels.} The ``web general'' label had mixed two questions, where a text came from and what kind of
Tamil it is. After the relabelling, limits on how much any one source family could contribute were kept as a separate
control. The labels apply to sources and views of sources, not to individual rows, and make no claim about the
quality of each record.

\paragraph{Collapsed passage and vowel-sign order.} The collapsed passage was one the first author's review had not yet
reached. In the filter written afterwards, a swap that would repair vowel signs stored in the wrong order lowered the
share of known words from 0.787 to 0.682, because most affected documents were only partly misordered.

\paragraph{Review-sheet findings.} The 37 unusable archive.org passages were scan watermarks repeated down the page,
thesis front matter, and garbled OCR that had passed the filter. The sheet itself cut passages at 700 characters
without a marker, so the backfill passages, with a median length of 2,202 characters, were judged on about a third of
their text; checking the hidden text afterwards showed that no verdict would have changed. A second reading pointed
the same way: of 60 passages sampled for an evaluation set, 13 failed when read, 8 of them broken archive.org scans,
while the typed Project Madurai e-texts showed no OCR damage. Scans stayed in pretraining; the typed text used for
evaluation passages was news published after the corpus freeze, and Project Madurai.

\paragraph{Deduplication of the first corpus.} Exact deduplication across the 13 web corpora, over 180.6M rows and
48.9B tokens already deduplicated within each source, removed a further 6.0\% of tokens; the share each source lost
depends on the order in which the sources were processed. Near-duplicate removal with MinHash on character 20-grams
removed 6.1B of OSCAR's 17.9B tokens; it ran on the local GPU with NVIDIA's NeMo Curator, because as a single CPU
process it would have taken three to four days. A 250-row sample of the other web sources led us to drop Sangraha's
verified slice, of which only 31\% of sampled tokens survived near-duplicate removal, and to cap CulturaX at 1.5B
tokens. A full near-duplicate pass across all sources was planned and not completed before the run; the second corpus
did it (Section~\ref{sec:data-dedup}).

\paragraph{Pool and exposure.} Only 0.72B tokens of the first pool came from outside the web. The final exposure plan
drew 30.0B training tokens from 33.7B available.

\paragraph{Training runs.} The H200 attempts began at the end of June; after about three weeks we rewrote the launch
process and moved to Lambda. The 4,221 lost steps were redone from an earlier checkpoint.

\subsection{Second-corpus construction: further detail}\label{app:corpus-steps}

Details of the corpus and tokenizer work of Section~\ref{sec:data} that the section and the rules in
Appendix~\ref{app:corpus-rules} do not give.

\paragraph{Admission rule.} Open licences were Apache-2.0, MIT, BSD, CC0, CC-BY, CC-BY-SA, ODC-By or an equivalent;
GPL-3.0 was counted as open, and non-commercial, no-derivatives and unlicensed datasets as not open. Nothing derived
from the six NVIDIA datasets is redistributed either. The local material left out by the rule (blogs, a web magazine,
Project Madurai and scanned books) was not on Hugging Face. Three other non-Hugging-Face collections (Kaggle Tamil news,
CC-100 Tamil and Leipzig) were not admitted, because 61--79\% of their text already appeared in the admitted crawls
(Appendix~\ref{app:corpus-overlap}).

\paragraph{Source labels.} Each source carries three labels, recorded separately. The \emph{licence class} is open,
local (our own collection, admitted by decision), NVIDIA training-only, or our own generation (conversations written by
an open-weight teacher, used only in the decay stage). The \emph{provenance} is native text, machine-translated Tamil
or machine-translated romanized Tamil. The \emph{tier} groups sources by kind and expected quality: curated text such
as Wikipedia and verified Sangraha, books and e-texts, web crawls, sentence-per-row corpora, and OCR output. The labels
come from documentation and read samples, not from judgements on individual documents.

\paragraph{Thamizhmalai.} The dropped corpus held 1.16B estimated tokens of sentence fragments with placeholder tokens
such as \texttt{NUM}; its share went to the Tamil web tier.

\paragraph{indic-ocr pass.} The 976 PDFs were 572 Tamil Nadu state textbooks and 404 Tamil Library books whose text
layers were in legacy encodings or missing. In a scouting run on 30 pages, pages with corrupt PDF text layers came back
as text that read cleanly by eye; no accuracy was measured, since there was no reference text.

\paragraph{Sarvam selection.} Sarvam's OCR was compared on two sets of 25 pages. On pages from the indic-ocr pass it
matched indic-ocr (mean garble score 0.170 against 0.167); on damaged library scans it scored 0.23 against 0.91 for the
existing text; and it was much faster. The budget covered the 358 books.

\paragraph{Containment rather than similarity.} Whole-book OCR documents cannot be matched against page-sized documents
by document-level similarity, so OCR books and cleaner copies of the same text were matched by containment
(Appendix~\ref{app:corpus-rules}). Zero-width joiners, besides old vocabulary, made classical texts look damaged to
the garble score.

\paragraph{Transitive groups.} In the largest MinHash group, pairwise similarity on the pairs we read ranged from 0.00
to 0.43; after the direct-band rule, the pairs read had similarities of 0.62--1.00. Only 20\% of MADLAD-400's Tamil
survived near-duplicate removal.

\paragraph{Later deduplication passes.} Because any duplicate left in the corpus would be repeated along with the text
around it, three further passes followed (Appendix~\ref{app:corpus-rules}). Sentence-level deduplication did what
MinHash over 2,000-character blocks cannot; the IITB monolingual corpus, for example, fell from 2.09B estimated tokens
to 0.30B in the pack.

\paragraph{SFT text in the instruction pool.} The exclusion was tightened over three review rounds, as each round
found a path by which SFT text still entered, until the final verification found no match with the SFT set among the
2.54M instruction and Tanglish documents it checked (Table~\ref{tab:pack-review}).

\paragraph{Pack composition.} Of the 39.39B pack tokens, 2.7B are machine-translated Tamil and romanized Tamil.

\paragraph{Whitespace loss and stopping.} In three 50M-token slices of the first model's last training pack, the
tokenizer's newline symbol occurred zero times, and indented code came back from a round trip as one line. Its chat
stop marker also never occurred in pretraining. The two have different fixes: the whitespace loss is a tokenizer
defect, while where a chat turn ends is taught by SFT, which uses the new tokenizer's reserved stop marker, so the
tokenizer change alone does not determine whether a model stops.

\paragraph{Tokenizer candidates.} The evaluation prompts and held-out texts were removed from the tokenizer sample by
hash and substring. The byte-level BPE (Hugging Face \texttt{tokenizers}) was trained on the whole sample and the
SentencePiece BPE on half of it; SentencePiece starts from characters and falls back to raw bytes for unseen ones, and
emitted 2,257 byte-fallback tokens on code. A byte-level BPE trained on the same half sample came within 0.01 tokens
per word of the full one, so the difference comes from the algorithm rather than the amount of training text.

\paragraph{Pre-tokenizer and normalisation.} Qwen2.5's English-heavy vocabulary is the other likely cause of its 9.4
tokens per Tamil word; we did not train a tokenizer without our pattern change to separate the two. The tokenizer has
no normaliser; the pack builder applies NFC before tokenization. No token boundary falls inside a UTF-8 character;
the splits counted in Section~\ref{sec:data-tok} are inside grapheme clusters only.

\paragraph{Verse-cost test.} The tokenizer sample held only 0.3\% classical text. Retraining after adding 20, 40 or 80M
characters of classical text lowered the cost of Thirukkural verse by 3.8\% at most and of a held-out Sangam set by
2.6\%, while Tanglish and code became up to 1\% more expensive; a Kural couplet split into the same 19 tokens with eight
times the classical data. The remaining cost may come from the archaic word forms of classical Tamil. A switch of
tokenizer would also have invalidated every token count already measured.

\begin{figure}[!htbp]
\centering
\includegraphics[width=\linewidth]{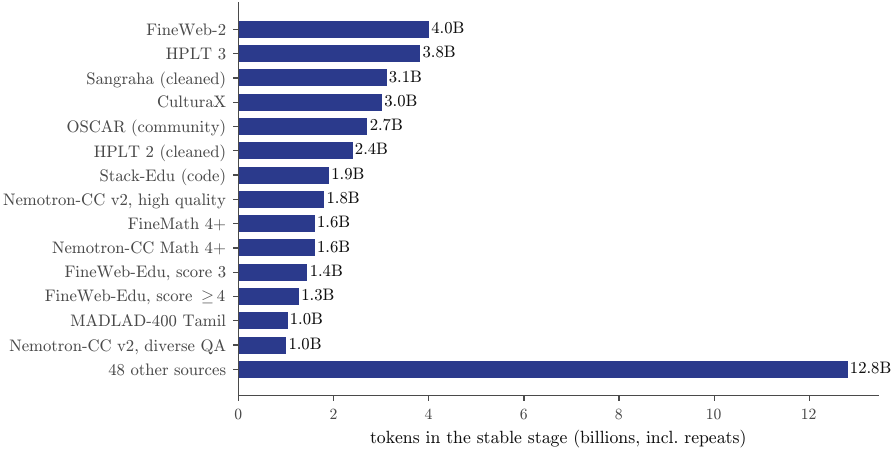}
\caption{Stable-stage mix by source (43.5B tokens including repeats, 63 sources).}
\label{fig:mix}
\end{figure}

\paragraph{Stable-stage mix.} Figure~\ref{fig:mix} shows the stable-stage mix by source; Tamil is 60.9\% of its tokens.

\paragraph{Tamil-share proxy.} The English side of the proxy was educational web text, and the models were scored on
held-out Tamil text in four registers and on held-out English. Each step up in Tamil share lowered Tamil loss by more
than three times the seed spread; from 50\% to 65\% Tamil gained 0.042 in Tamil loss for 0.085 in English
(Table~\ref{tab:proxy}). We read this as placing the useful range near 60--65\% Tamil, beyond which the English cost grew
fastest. The proxy also used the first model's tokenizer, and at a fixed budget a higher-Tamil arm sees more Tamil.

\paragraph{Repetition by tier.} Curated sources and Project Madurai were repeated four times across the two stages,
other books two to three times, and sentence corpora, OSCAR and most OCR output once or one and a half times. The
per-source repeats are in Tables~\ref{tab:inventory-ta} and~\ref{tab:inventory-other}.

\paragraph{Decay-stage sources.} The decay stage keeps 22 of the stable stage's 63 sources and drops CulturaX, HPLT,
OSCAR, MADLAD-400 and the two other Sangraha cuts. FineMath-4+ goes from 3.7\% to 12.3\% of tokens and FineWeb-Edu
(score $\geq$4) from 2.9\% to 10.8\%. Tamil monolingual text is 40\% of decay tokens.

\paragraph{Decay-stage instruction sources.} The five sources are conversations written for this project by an
open-weight teacher (passage tasks, Tanglish, formatting, tool use, reasoning, refusal, translation, open questions,
multi-turn chat and Thirukkural tasks), an instruction pool assembled during the work on the first model, a set of
romanized Tamil dialogues and a grounded question-answering set. The SFT conversation families were selected before the
decay data was built, and the evaluation prompts were also kept out of it.

\section{Training recipes and checkpoint history}\label{app:recipes}

\subsection{Pretraining recipe}\label{app:pretrain}

Table~\ref{tab:pretrain-recipe} gives the complete pretraining configuration. Settings that differ between the two
stages are in Table~\ref{tab:schedule}. The decay stage doubles the context and halves the number of sequences per
step. Each stage was validated on held-out documents of its own blend; the same documents were held out in both
stages, but because the two stages draw on different datasets their validation sets differ and their losses are not
comparable with each other.

\begin{table}[!htbp]
\centering
\small
\caption{Pretraining configuration of \agaram{} 2.5B.}
\label{tab:pretrain-recipe}
\begin{tabular}{L{4.0cm}L{8.8cm}}
\toprule
Setting & Value \\
\midrule
\multicolumn{2}{l}{\textit{Model}} \\
Parameters & 2,525,757,696 \\
Layers / hidden size & 28 / 2,304 \\
Attention & 36 heads of dimension 64, full multi-head attention, pre-norm \\
Feed-forward & SwiGLU, width 9,216 \\
Normalisation & RMSNorm \\
Positions & rotary embeddings, base 500,000 from step 0 \\
Biases / embeddings & none / input and output tied \\
Vocabulary & 64,000 (byte-level BPE, Section~\ref{sec:data-tok}) \\
\addlinespace
\multicolumn{2}{l}{\textit{Data}} \\
Pack & 39.39B tokens, 69,170,890 documents, 68 datasets (Appendix~\ref{app:corpus}) \\
Stages & stable 43.5B tokens from 63 datasets; decay 6.5B tokens from 27 \\
Document boundaries & one end-of-text token per document; attention restricted to each document (packed batches
with per-document offsets); positions restart per document \\
\addlinespace
\multicolumn{2}{l}{\textit{Optimisation}} \\
Optimiser & Adam, $\beta=(0.9, 0.95)$, weight decay 0.1, gradient clipping 1.0 \\
Learning rate & linear warmup over 2,000 steps to $3\times10^{-4}$; constant in the stable stage; linear decay to
$3\times10^{-5}$ over the 12,398 decay steps, no second warmup \\
Batch & 524,288 tokens per step (128 $\times$ 4,096, then 64 $\times$ 8,192); micro batch 1 \\
Precision & bf16 mixed precision; optimiser state distributed across the eight GPUs \\
Decay initialisation & last stable checkpoint, weights and optimiser state \\
\addlinespace
\multicolumn{2}{l}{\textit{Validation and checkpoints}} \\
Validation set & last 0.1\% (at most 500) of the documents of each dataset with at least 20,000 documents, removed from
the training view; same documents in both stages \\
Validation and checkpoint interval & every 1,000 steps \\
\addlinespace
\multicolumn{2}{l}{\textit{Infrastructure}} \\
Framework & NeMo 2 on Megatron-Core (container nemo:25.04), Transformer Engine attention \\
Parallelism & data parallel over 8 GPUs, tensor parallel size 1 \\
Hardware & one node of 8$\times$H100 SXM5 (Lambda), USD 31.92 per hour \\
\bottomrule
\end{tabular}
\end{table}

\subsection{Pretraining decisions: further detail}\label{app:pretrain-full}

\paragraph{Size projection.} The part of the USD 5,000 budget not planned for pretraining was left for post-training, evaluation and a
failed launch. Candidate sizes were costed by scaling the first model's measured throughput on the same hardware by
parameter count.

\paragraph{Document-masking implementation.} Each packed sample is split at its end-of-text tokens and the pieces are
passed to the attention kernel as a packed batch with per-document offsets, so flash attention stays on. On a single
test GPU the masked path was about 2\% slower.

\paragraph{Stage split.} The 43.5B stable and 6.5B decay tokens are 87\% and 13\% of the run.

\paragraph{Long-document shares.} Table~\ref{tab:longdocs} gives the measurement behind the context-length choice. SFT
also used 8,192 tokens.

\begin{table}[!htbp]
\centering
\small
\caption{Share of pack tokens in long documents, and the longest SFT conversation.}
\label{tab:longdocs}
\begin{tabular}{lrr}
\toprule
 & Stable stage & Decay stage \\
\midrule
Tokens in documents longer than 4,096 & 20.0\% & 15.7\% \\
Tokens in documents longer than 8,192 & 11.5\% & 9.2\% \\
\midrule
Longest SFT conversation & \multicolumn{2}{c}{2,498 tokens} \\
\bottomrule
\end{tabular}
\end{table}
\subsection{Tamil-share proxy}\label{app:pretrain-proxy}

The proxy of Section~\ref{sec:data-decay} trained a 153M-parameter dense model (11 layers, hidden size 768, 12 heads,
context 2,048) for 15,258 steps of 131,072 tokens (2.0B tokens) per arm on one RTX 5090, with learning rate
$3\times10^{-4}$, 500 warmup steps, decay to a tenth, and weight decay 0.1. Each arm mixed Tamil with English
educational web text only, so that the share was the only variable. The packs used the first model's tokenizer, not
the new one. Losses are in nats per token on the held-out registers of the evaluation ruler then in use; its literary
register was later found to consist of font-damaged text and was replaced for the fresh run's own evaluation
(Section~\ref{sec:res-register}).

\begin{table}[!htbp]
\centering
\small
\caption{Final held-out loss of the Tamil-share proxy arms (step 15,258; lower is better). ``Tamil'' is the mean of
the four Tamil registers.}
\label{tab:proxy}
\begin{tabular}{lcrrrrrr}
\toprule
Tamil share & Seed & Literary & Colloquial & Formal & Tanglish & Tamil & English \\
\midrule
50\% & 0 & 6.630 & 4.962 & 4.918 & 4.728 & 5.310 & 2.641 \\
50\% & 1 & 6.601 & 4.960 & 4.911 & 4.723 & 5.299 & 2.638 \\
65\% & 0 & 6.590 & 4.930 & 4.871 & 4.657 & 5.262 & 2.726 \\
80\% & 0 & 6.577 & 4.896 & 4.832 & 4.606 & 5.228 & 2.878 \\
80\% & 1 & 6.567 & 4.894 & 4.824 & 4.612 & 5.224 & 2.876 \\
\bottomrule
\end{tabular}
\end{table}

Seed spreads were 0.011 (50\%) and 0.004 (80\%) on Tamil and 0.003 and 0.002 on English. Using seed means, raising the
Tamil share from 50\% to 65\% lowered Tamil loss by 0.042 and raised English loss by 0.085; from 65\% to 80\%, Tamil fell
by 0.036 and English rose by 0.152. Per percentage point, the Tamil gain fell slightly (0.0028 to 0.0024) while the
English cost nearly doubled (0.0057 to 0.0101). At a fixed budget a higher-Tamil arm also sees more Tamil tokens, so
these differences do not separate a benefit of the mixture from a benefit of more Tamil data.

\subsection{Validation loss and throughput}\label{app:pretrain-curves}

\begin{figure}[!htbp]
\centering
\includegraphics[width=0.9\linewidth]{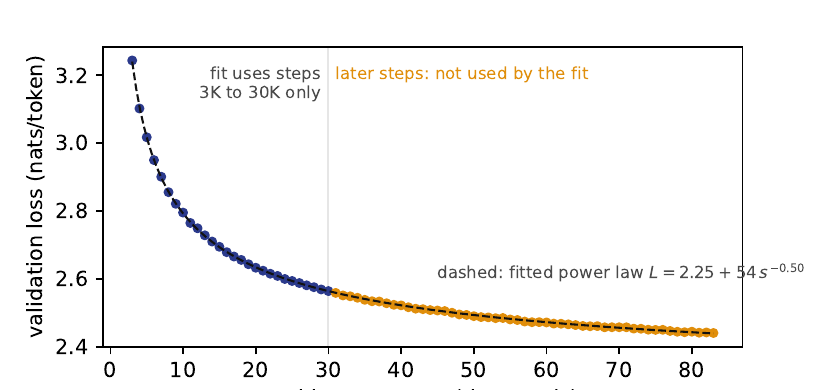}
\caption{Stable-stage validation loss against a power law fitted to steps 3K--30K only.}
\label{fig:forecast}
\end{figure}

Figure~\ref{fig:forecast} compares the stable-stage validation losses with a power law $L(s) = a + b\,s^{-c}$ in the
step $s$. The fit used only the validation losses of steps 3,000--30,000: the exponent $c$ was chosen by grid search
over 0.05--1.5 in steps of 0.005, and $a$ and $b$ by least squares for each $c$. The best fit was $a = 2.25$, $b = 54.1$
and $c = 0.50$. Over the later steps of the stage, which the fit did not use, the residuals had mean $-0.0008$ and
maximum absolute value 0.0038. The fit was made after the stage had finished, while preparing this report; it is a
description of the run, not a forecast that was used during it.

\begin{figure}[!htbp]
\centering
\includegraphics[width=0.9\linewidth]{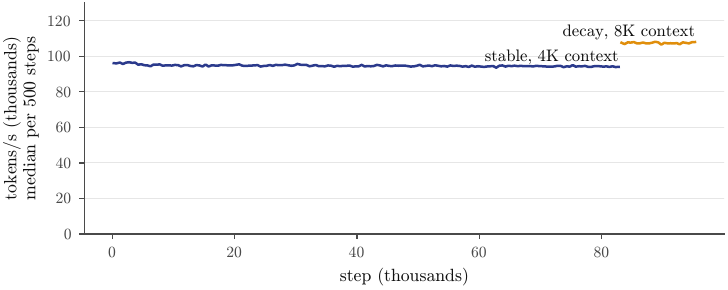}
\caption{Throughput on 8$\times$H100 (median per 500 steps).}
\label{fig:throughput}
\end{figure}

Decay-stage validation loss fell from 2.080 at step 1,000 to 1.954 at step 8,000, 1.925 at step 10,000, 1.912 at
step 11,000 and 1.903 at the end. Throughput stayed flat through both stages (Figure~\ref{fig:throughput}); the decay
stage processed more tokens per second at twice the context.

\subsection{Run safeguards and chronology}\label{app:pretrain-ops}

\paragraph{Obtaining the node.} Eight-GPU capacity on the provider was scarce in the week before launch. One window
was lost to an HTTP error whose cause the launch script did not record, one to a bug in the script's argument check,
and one because the capacity was gone about two seconds after our watcher saw it. A reserver that polled every two seconds and
launched immediately caught an 8$\times$H100 node on 2 October 2026 at 02:33 UTC. The planned throughput test of the
2.5B shape was skipped because of a missing file in the hand-off between the test and the training script, so
training started with the default settings, including tensor parallel size 1.

\paragraph{Spend deadlines.} The node was armed to terminate itself after a fixed time, and a separate watcher on a
local PC held a second, later deadline. The first deadline, set for a USD 4,500 cap, allowed 140 hours, less than the
146--149 hours projected for both stages. Nineteen minutes after launch the cap was raised to USD 5,000, which moved
the node's deadline to 155.5 hours of training.

\paragraph{Checkpoint copies.} A copier on a DGX Spark machine pulled the newest complete checkpoint off the node about
every two hours. Because the training framework renames its latest checkpoint at each save, the copier first
hard-linked the checkpoint on the node, then copied the snapshot with six parallel streams (about 33 GB in 30--70
minutes) and kept it only after a second pass found nothing left to transfer. A second process archived a copy every
10,000 steps, and the end of each stage, to a local disk.

\paragraph{Release of the node.} After training, the node waited for a marker that the copier wrote only once the
final checkpoint and the run records were safe, and then terminated its own instance. A failed stage would have held
the node for two hours for inspection. One gap was found during the run: until the end of the stable stage the copier
transferred checkpoints but not the per-step metrics, validation records and logs, which existed only on the node. They
were copied by hand on 7 October, and the copier was changed to include them in every cycle and before writing its
final marker.

\paragraph{Monitoring.} A check ran every hour for the first day and every two hours after that. It recorded the step,
throughput, loss and validation trend, GPU use and disk, the cloud instance, the deadline guards, the copies, and the
SHA-256 hashes of the data files, which matched the launch baseline at every check. Each check appended a line to a
heartbeat file in the repository, and a scheduled cloud job sent an alert if that file went stale. During the run,
intermediate checkpoints were converted to the Hugging Face format and evaluated on the Sparks; on the step-8,000
checkpoint the converted model's logits matched the training framework's to within 0.0051 (threshold 0.018), with the
same top token at every position.

\paragraph{End of the run.} The stable stage ran from 03:14 UTC on 2 October to 13:23 UTC on 7 October, and the decay
stage from 13:25 UTC on 7 October to 06:59 UTC on 8 October. The final checkpoint reached the copier at 07:48 and the
local archive at 08:02; the instance was gone from the provider's list by 07:51. The instance ran about 149 hours in
total, 147.7 of them training. The provider's two invoices for the period came to USD 4,776.50, paid from compute
credits; they also cover a 3.7-hour SFT attempt on a single H100, so the pretraining run itself cost about USD 4,765,
in agreement with the list price of 149.3 hours at USD 31.92 per hour.

\subsection{Post-training recipes}\label{app:post}

All SFT runs used full fine-tuning with AdamW, gradient clipping at 1.0, loss on assistant turns only, sequences of up to
8,192 tokens packed into batches of 65,536 tokens per optimiser step, and bf16 computation. They differ in starting
weights, learning rate, loss averaging and whether the optimiser kept an fp32 master copy of the weights
(Table~\ref{tab:app-sft-hparams}). Token-averaged loss means every supervised token in a batch weighs the same;
conversation-averaged loss first averages within each conversation, so long replies weigh no more than short ones. Before
the three-epoch run every SFT run updated bf16 weights directly; at small learning rates many such updates round away.

\begin{table}[!htbp]
\centering
\scriptsize
\caption{SFT runs of \agaram{} 2.5B. Steps are optimiser steps of 65,536 tokens. ``--'' means not recorded in the
retained run evidence.}
\label{tab:app-sft-hparams}
\setlength{\tabcolsep}{2.5pt}
\begin{tabular}{L{1.75cm}L{1.5cm}L{1.2cm}L{2.35cm}L{1.55cm}L{1.0cm}L{1.6cm}L{1.5cm}}
\toprule
Run & Starts from & Records & Learning rate and schedule & Loss & Weights & Steps (passes) & Hardware \\
\midrule
Initial SFT & 2.5B base & 134,428 & $2\times10^{-5}$, 3\% warmup, cosine to 0 & token & bf16 & 2,274 (3; 758 each) & 1$\times$H200, 9.5 h \\
Round 2 & initial SFT epoch 3 & 19,565 & as initial SFT & token & bf16 & 345 (3; 115 each) & 1 DGX Spark \\
Round 3 & round 2 epoch 2 & 8,210 & as initial SFT & token & bf16 & 80 (2; 40 each) & 1 DGX Spark \\
Round 4 & round 3 epoch 2 & 4,451 & as initial SFT & token & bf16 & 34 (2; 17 each) & 1 DGX Spark \\
Diagnostic pilots A, B & round-4 checkpoint & -- & -- & token & bf16 & 25 (control 17) & DGX Spark \\
Kill test & round-4 checkpoint & 2.2M tokens per arm & $10^{-5}$ & token & bf16 & 34 & 2 DGX Spark \\
Kill-test rerun & round-4 checkpoint & 2.64M tokens per arm & $10^{-5}$ & token & bf16 & 41 & 2 DGX Spark \\
Repair round & round-4 checkpoint & 51,670 & $2\times10^{-6}$, 5\% warmup, decay & conversation & bf16 & 384 (1) & 1 DGX Spark \\
Three-epoch run & 2.5B base & 170,386 & $2\times10^{-5}$, 3\% warmup, cosine to 0 & token & fp32 master & 3,018 (3; 1,006 each) & 1$\times$H100, about 15.5 h \\
SFT v2 pilots, 4M & 2.5B base & arm mixes & P0, P1 $2\times10^{-5}$; P2--P4 $10^{-5}$; 5\% warmup, cosine to 10\% floor & token & fp32 master & 147--189 & 1$\times$H100 each \\
SFT v2 confirmation, 8M & 2.5B base & P0, P4 & as 4M; seeds 7 and 19 & token & fp32 master & 291--378 & 1$\times$H100 each \\
SFT v2 final & 2.5B base & 222,662 & $10^{-5}$, 51 warmup steps, cosine to $10^{-6}$; reshuffle each pass & token & fp32 master & 1,029 (1.5; 686 per pass) & 1$\times$H100, 9.0 h \\
\bottomrule
\end{tabular}
\end{table}

The first attempt at initial SFT, on a rented H100 PCIe, ran at 37 seconds per step because the GPU was thermally
throttled (87--89\,$^\circ$C) and was stopped after 3.7 hours; the run then moved to an H200 at about 15 seconds per
step. A 4M pilot of SFT v2 also ran on a throttled H100 at about 74 seconds per step and was kept, since throttling slows a
run without changing its arithmetic.

\subsection{Post-training checkpoint history}\label{app:post-history}

\begin{table}[!htbp]
\centering
\scriptsize
\caption{Post-training checkpoints named in the paper. Fate: what the checkpoint was used for.}
\label{tab:app-sft-history}
\setlength{\tabcolsep}{3pt}
\begin{tabular}{L{2.8cm}L{3.4cm}L{2.4cm}L{4.4cm}}
\toprule
Checkpoint & Run and step & Starts from & Fate \\
\midrule
Initial SFT epochs 1--3 & initial SFT, steps 758, 1,516, 2,274 & 2.5B base & epoch 3 started the targeted rounds \\
Round 2, 3 epochs & round 2, step 230 kept & initial SFT epoch 3 & start of round 3 \\
Round 3, epoch 2 & round 3, step 80 & round 2 epoch 2 & start of round 4 \\
Round-4 checkpoint & round 4, step 17 (epoch 1) & round 3 epoch 2 & first release candidate; public preview 14--18 October;
control in every later comparison; representation analysis (Section~\ref{sec:repr}) \\
Diagnostic pilots A, B & 25 steps each & round-4 checkpoint & not released \\
Repair-round checkpoints & steps 96, 192, 288, 384 & round-4 checkpoint & all failed the gate; not released \\
Three-epoch run, epochs 1--3 & steps 1,006, 2,012, 3,018 & 2.5B base & epoch 3 is the second comparison model for SFT v2 \\
SFT v2 pilot endpoints & P0--P4 at 4M; P0, P4 at 8M & 2.5B base & chose learning rate and mix \\
SFT v2 step 700 & final run, step 700 & 2.5B base & released model \\
SFT v2 step 1,029 & final run, last step & 2.5B base & information only \\
\bottomrule
\end{tabular}
\end{table}

Table~\ref{tab:rounds} gives the lineage of the round-4 checkpoint round by round (Section~\ref{sec:sft-targeted}).

\begin{table}[!htbp]
\centering
\small
\caption{The SFT lineage of the round-4 checkpoint. Each round starts from the checkpoint kept in the row above.}
\label{tab:rounds}
\setlength{\tabcolsep}{4pt}
\begin{tabular}{lL{5.6cm}rrrl}
\toprule
Stage & Areas of the new records & New & Replayed & \makecell[r]{Epochs\\run} & Kept \\
\midrule
Initial SFT & general conversation, tools, translation, crisis, identity pairs (Section~\ref{sec:sft-targeted}) &
134,428 & -- & 3 & epoch 3 \\
\addlinespace
Round 2 & identity, tool results, crisis, tool-call corrections, memory across turns, long answers, false premises,
harmful requests, multi-turn repair, verified Thirukkural, format instructions, the model's charter & 7,565 & 12,000 & 3
& epoch 2 \\
\addlinespace
Round 3 & identity in English and Tanglish and under pressure, harmful requests, crisis (helpline first), script
switching between turns & 2,210 & 6,000 & 2 & epoch 2 \\
\addlinespace
Round 4 & identity (false sizes, training data, ``are you human''), harmful requests, crisis (secrecy, fear of
judgement) & 1,451 & 3,000 & 2 & epoch 1 \\
\bottomrule
\end{tabular}
\end{table}

\begin{table}[!htbp]
\centering
\small
\caption{Post-training development in order. Each row starts from what the previous row observed. Checkpoint names
follow Table~\ref{tab:ledger}.}
\label{tab:post-stages}
\setlength{\tabcolsep}{3pt}
\begin{tabular}{L{2.2cm}L{3.6cm}L{3.5cm}L{3.9cm}}
\toprule
Stage & Observation that motivated it & Data or training change & Test and outcome \\
\midrule
Initial SFT (\S\ref{sec:sft-targeted}) & Staged SFT test on the older base stayed weak at open chat (\S\ref{sec:restart}) &
134,428 conversations, mostly teacher-written; general conversations held back from pretraining; 3 epochs &
About half of replies judged usable; epoch 3 kept on a blind human comparison \\
\addlinespace
Targeted rounds 2--4 (\S\ref{sec:sft-targeted}) & Wrong identity, harmful compliance, trust in irrelevant tool results &
11,226 new scenario conversations plus replay, in three rounds & Probe items improved; no clear change on a later
held-out chat set \\
\addlinespace
Diagnostic pilots (\S\ref{sec:sft-diag}) & Multiple-choice and instruction-following scores near the base model's &
Letter-balanced multiple-choice rows; checked constraint-following rows & No gain on the independent benchmarks;
neither released \\
\addlinespace
Preview-informed repair round (\S\ref{sec:sft-preview}) & Public use: greetings, language switching, incoherent
replies & Repair, hold-your-ground and preview-scenario rows on the round-4 checkpoint & No checkpoint passed the
pre-set gate (tool calls fell) \\
\addlinespace
Three-epoch run (\S\ref{sec:sft-restart}) & Stacked rounds had not produced a gated improvement & All 170,386 SFT
records, one run from the base, fp32 master weights & Better chat and tool calls; English-to-Tamil translation fell by
6.8 chrF \\
\addlinespace
SFT v2 (\S\ref{sec:sft-v2}--\ref{sec:sft-select}) & Audit of the restart: translation 1.4\% of supervision, long
answers, repeated batch order & 222,662 records with per-component token budgets, a fifth translation; pilots at
matched exposure & Released step 700; preferred over round 4 on the sealed set, not over the three-epoch run \\
\bottomrule
\end{tabular}
\end{table}

\subsection{Initial SFT set composition}\label{app:post-initial}

\begin{table}[!htbp]
\centering
\small
\caption{Initial SFT set by category. The base set of 121,465 conversations was built before pretraining; three
training-only sets were added later.}
\label{tab:app-sft-initial}
\setlength{\tabcolsep}{4pt}
\begin{tabular}{lr}
\toprule
Category & Conversations \\
\midrule
Open questions and explanations & 28,333 \\
Multi-turn memory and correction & 15,980 \\
Tool calling & 11,890 \\
Casual Tanglish & 10,322 \\
Explaining and teaching & 8,299 \\
Everyday practical help & 7,641 \\
Thirukkural and classical text & 7,188 \\
Format following & 6,798 \\
Safety & 5,405 \\
Passage tasks & 4,917 \\
Uncertainty and refusal & 4,595 \\
Reasoning & 3,704 \\
Translation & 3,369 \\
Identity, engagement and a fixed chat set & 3,024 \\
\midrule
Base set & 121,465 \\
\addlinespace
Crisis support (444 conversations, each repeated eight times) & 3,552 \\
Tool calls with computed arguments & 3,411 \\
Human translation pairs from BPCC & 6,000 \\
\midrule
Training set & 134,428 \\
Validation set & 2,736 \\
\bottomrule
\end{tabular}
\end{table}

\subsection{SFT v2 budgets and supervision by component}\label{app:post-v2}

\begin{table}[!htbp]
\centering
\small
\caption{The released model's SFT data by component. Shares are of the 14,547,507 supervised tokens in one pass.
Distinct records exclude deliberate repeats.}
\label{tab:sft-mix}
\setlength{\tabcolsep}{4pt}
\begin{tabular}{L{6.4cm}rrrr}
\toprule
Component & Records & Distinct & \makecell[r]{Supervised\\tokens} & Share \\
\midrule
General teacher-written conversations, turns up to 384 tokens & 10,839 & 10,839 & 3,701,744 & 25.4\% \\
General teacher-written conversations, long answers (read before use) & 722 & 722 & 703,209 & 4.8\% \\
Tamil and Tanglish conversations from the staged SFT test & 28,870 & 28,870 & 3,000,774 & 20.6\% \\
Targeted behaviour data (rounds 2--4, preview-informed round, safety) & 31,008 & 31,008 & 2,683,952 & 18.4\% \\
Tool use (calls, tool results, no-tool cases) & 12,960 & 12,747 & 800,024 & 5.5\% \\
Crisis support & 2,503 & 2,072 & 350,030 & 2.4\% \\
Model identity & 5,160 & 2,580 & 283,758 & 2.0\% \\
\addlinespace
Translation English$\rightarrow$Tamil, chat format & 83,762 & 70,841 & 1,900,046 & 13.1\% \\
Translation Tamil$\rightarrow$English, chat format & 32,850 & 31,204 & 800,028 & 5.5\% \\
Translation, plain completion records & 13,614 & 13,614 & 300,006 & 2.1\% \\
Transliteration, Tamil script to Tanglish & 374 & 374 & 23,936 & 0.2\% \\
\midrule
Total & 222,662 & 204,871 & 14,547,507 & 100\% \\
\bottomrule
\end{tabular}
\end{table}

\begin{table}[!htbp]
\centering
\small
\caption{Findings from the three-epoch run and the corresponding SFT v2 changes.}
\label{tab:v2-changes}
\setlength{\tabcolsep}{3pt}
\begin{tabular}{L{5.6cm}L{7.6cm}}
\toprule
Finding & Change in SFT v2 \\
\midrule
Validation loss best after epoch 1 & At most 1.5 passes; learning rate chosen by pilot; cosine decay to a 10\% floor;
checkpoints every 50 steps \\
Same batch order every epoch & Reshuffle and repack every epoch with recorded seeds \\
General conversation 63\% and long answers dominant & Supervised-token budget per component; general conversation 30\%;
turns over 384 tokens capped at 5\% of supervision \\
Translation 1.4\% of supervision, some defective pairs & BPCC human pairs at about a fifth of supervision, bare targets,
entity and alignment filters; plain-completion records tested as a pilot arm \\
Old keep-out lacked FLORES and IN22 & Per-sentence keep-out against FLORES and IN22, both sides
(Section~\ref{sec:eval-overlap}) \\
Crisis copies, apology openers, missing passages & Crisis at most twice; safety apology openers at most 15\%; records
referring to an absent passage dropped \\
\bottomrule
\end{tabular}
\end{table}

The SFT v2 assembler gave each component a budget of supervised tokens and filled it from the available records in a
fixed precedence order (translation, transliteration, tool, crisis, identity, other targeted data, staged-test
conversations, general conversations), so that each record counts in exactly one component.
Table~\ref{tab:app-v2-budget} compares the budgets with what the final mix contains. Three budgeted components were
never built, and three others ran short of available records.

\begin{table}[!htbp]
\centering
\small
\caption{SFT v2 component budgets and the released mix (supervised tokens per pass).}
\label{tab:app-v2-budget}
\setlength{\tabcolsep}{4pt}
\begin{tabular}{lrrl}
\toprule
Component & Budget & Released mix & Note \\
\midrule
General conversations, short turns & 3,700,000 & 3,701,744 & \\
General conversations, long answers & 800,000 & 703,209 & held by the 5\% long-answer cap \\
Staged-test conversations & 3,000,000 & 3,000,774 & \\
Targeted behaviour data & 2,600,000 & 2,683,952 & all available records \\
Tool use & 800,000 & 800,024 & \\
Crisis support & 350,000 & 350,030 & \\
Identity & 400,000 & 283,758 & all available records, twice \\
Translation en$\rightarrow$ta, chat & 1,900,000 & 1,900,046 & pairs reused up to 1.3 times \\
Translation ta$\rightarrow$en, chat & 800,000 & 800,028 & \\
Translation, completion records & 300,008 & 300,006 & 10\% of translation \\
Transliteration & 100,000 & 23,936 & all available records \\
Aya Dataset Tamil & 400,000 & 0 & not built \\
Grounded short answers & 600,000 & 0 & not built \\
\midrule
Total & 15,750,008 & 14,547,507 & \\
\bottomrule
\end{tabular}
\end{table}

\paragraph{Repeats and filters.} Crisis-support conversations, copied eight times in the initial SFT set, appear at
most twice; identity conversations appear twice; human translation pairs were reused up to 1.3 times to fill the
translation budget. The assembler dropped 3,321 exact copies and 838 records under the benchmark keep-out checks,
including 67 whose matches were not fully covered by the allowlist (Section~\ref{sec:eval-overlap}); it also dropped
half of the general conversations that open with a filler phrase (5,103), 32 answers that refer to a passage missing from the
conversation, and enough safety records that open with an apology to bring those to 15\% of safety targets (1,767
dropped). Validation records that shared an input with a training record were removed (133).

Table~\ref{tab:app-components} gives the data behind Figure~\ref{fig:sft-supervision}. To compare the two mixes on one
footing, records are grouped by origin: the teacher's tool and translation conversations, which the three-epoch mix
counted inside the general conversations and the SFT v2 assembler moved to the tool and translation components, are
listed separately in both columns.

\begin{table}[!htbp]
\centering
\small
\caption{Supervised tokens by origin in the three-epoch run's mix (per epoch) and the released mix (per pass), recounted
with the training code's masking.}
\label{tab:app-components}
\setlength{\tabcolsep}{4pt}
\begin{tabular}{lrrrr}
\toprule
 & \multicolumn{2}{c}{Three-epoch run} & \multicolumn{2}{c}{Released} \\
Component & Tokens & Share & Tokens & Share \\
\midrule
General teacher-written conversations & 16,315,299 & 59.3\% & 4,428,889 & 30.4\% \\
Teacher-written tool conversations & 783,122 & 2.8\% & 261,938 & 1.8\% \\
Teacher-written translation conversations & 219,799 & 0.8\% & 180,567 & 1.2\% \\
Staged-test conversations & 5,654,773 & 20.6\% & 3,000,774 & 20.6\% \\
Targeted behaviour data & 2,763,693 & 10.1\% & 2,683,952 & 18.4\% \\
Targeted tool data & 538,977 & 2.0\% & 538,086 & 3.7\% \\
Crisis support & 902,068 & 3.3\% & 350,030 & 2.4\% \\
Identity & 141,942 & 0.5\% & 283,758 & 2.0\% \\
Human translation (BPCC) & 122,021 & 0.4\% & 2,819,513 & 19.4\% \\
Other translation records & 53,763 & 0.2\% & -- & -- \\
\midrule
Total & 27,495,457 & 100\% & 14,547,507 & 100\% \\
\bottomrule
\end{tabular}
\end{table}

\subsection{Literature consulted for the SFT redesign}\label{app:post-lit}

Table~\ref{tab:post-lit} lists the works read in full text when the SFT data was redesigned after the public preview
(Section~\ref{sec:sft-preview}) and what was done with each.

\begin{table}[!htbp]
\centering
\small
\caption{Works read when redesigning the SFT data after the preview, and their use. ``Considered'' means the method
was planned or discussed but not run.}
\label{tab:post-lit}
\setlength{\tabcolsep}{3pt}
\begin{tabular}{L{4.3cm}L{4.3cm}L{4.6cm}}
\toprule
Work & What we consulted it for & Use in this project \\
\midrule
\citet{gekhman2024finetuning} & Fine-tuning on facts the model does not already know &
Planned a known/unknown screen of knowledge targets; the wave was not generated (\S\ref{sec:sft-preview}) \\
\citet{kang2025unfamiliar}; \citet{ovadia2024finetuning} &
Unfamiliar fine-tuning examples; retrieval versus fine-tuning for new facts & Facts left to tool calls and passages, not
taught as new SFT targets; current-office questions call a lookup tool \\
R-Tuning \citep{zhang2024rtuning}; \citet{cheng2024aiassistants} & Training a model
to express uncertainty or decline & Considered for unknown facts; not built as a separate component \\
FactTune \citep{tian2024facttune}; Phi-4 \citep{abdin2024phi4} & Factuality
preference tuning; factuality and refusal trade-offs & Considered; not run \\
Physics of Language Models 3.1--3.3
\citep{allenzhu2024physics31,allenzhu2025physics32,allenzhu2025physics33} & Knowledge storage, extraction, manipulation and capacity & Background for the knowledge probe \\
Tülu 3 \citep{lambert2025tulu3}; OLMo 2 \citep{olmo2024olmo2} & SFT mixtures, on-policy preference data,
development and unseen evaluation sets, checkpoint averaging & Separate development and sealed sets; preference round
considered, not run \\
SmolLM2 \citep{allal2025smollm2} & Simpler SFT mixtures for small models & Reduced share of general conversation \\
Llama 3 \citep{dubey2024llama3}; Qwen2.5 \citep{qwen2024qwen25} & Iterative rounds, rejection sampling & Repair rows
from the model's own samples; rejection sampling not run \\
Qwen3 \citep{qwen2025qwen3}; Gemma 3 \citep{gemma2025gemma3}; DeepSeek-R1 \citep{guo2025deepseekr1} &
Distillation into small models & Teacher-written replies only; logit distillation not run (different tokenizers) \\
Apertus \citep{apertus2025}; Aya Expanse \citep{dang2024ayaexpanse} & Multilingual SFT,
length-normalised preference training, multi-teacher data, merging & Considered; checkpoint blends screened, not
adopted \\
\bottomrule
\end{tabular}
\end{table}

\subsection{Post-training development: further detail}\label{app:post-full}

Details, counts and episodes behind Sections~\ref{sec:sft-data}--\ref{sec:sft-select} that the main text and the
tables of Appendices~\ref{app:post} and~\ref{app:eval-sft} do not give.

\subsubsection*{Released SFT data}

\paragraph{Validation split.} A further 3,828 records form the validation split of the released data.

\paragraph{Translation records.} The BPCC pairs come from its human-translated wiki, daily-conversation and seed
subsets, with English-to-Tamil and Tamil-to-English roughly 70/30. Chat-format records give the request in English,
Tamil or Tanglish with varied wording and supervise the bare translation, with no wrapper text such as ``Here is the
translation''. The alignment floor is a LaBSE similarity of at least 0.65, set after reading the 0.60--0.65 band, where
24\% of pairs were defective; the other checks are entity and digit consistency and a Tamil-script check on Tamil
targets. The two chat-format translation components also keep 1,389 teacher-written translation conversations each; no
new teacher translations were generated for SFT v2. BPCC text is also in the pretraining pack, so what SFT adds is the
translation task in chat and completion form rather than new text.

\paragraph{General conversations.} The seed prompts span open questions, explanation, Tanglish small talk, multi-turn
memory, format following, passage tasks, tool calling and classical text. Whole families were held back before the
decay pack was built, so no SFT conversation was also pretraining text. Assistant turns longer than 384 tokens are 2.9\%
of supervised tokens in the final mix, under the 5\% cap. Outside the translation components, 53.8\% of supervised
target text is in Tamil script, 34.9\% in Tanglish, 9.0\% in English and 2.3\% in other forms; the classifier is a
heuristic, so these shares are estimates.

\paragraph{Unbuilt components and identity data.} Besides the Aya rows and grounded short answers
(Table~\ref{tab:app-v2-budget}), about 3,000 new tool and no-tool cases were planned and not built before the pilots.
The identity component comes from the rounds-2--4 and preview-round identity data, written by the teacher from one
confirmed fact sheet. Frontier models (Claude, Codex) planned the scenarios and wrote validators and filters.

\subsubsection*{Initial SFT}

\paragraph{Initial set.} The 121,465 base conversations hold 44.6M tokens; 52,548 have more than one user turn, 11,890
call tools, 76,865 are in Tamil script and 36,171 in Tanglish (categories in Table~\ref{tab:app-sft-initial}). The
3,411 added tool-call conversations have arguments that must be computed from the request.

\paragraph{Epoch choice.} The first author's blind comparison had 5 ties besides the 17 to 8. A volunteer's copy of the
same sheet also favoured epoch 3, but it was completed with AI assistance, so we count it as supporting evidence rather
than an independent human rating. Scripted format checks passed far more replies than the judge
(Table~\ref{tab:app-sft-epochs}) because they cannot detect wrong facts, and benchmarks moved by at most 1.0 point from
the base model.

\paragraph{Failures after initial SFT.} LIMA fine-tuned a 65B model on 1,000 curated examples; our identity pairs were
repeated eight times (976 records). Epoch 1 gave its name as ``\ta{அ}''. The judge passed one of five safety replies
from epoch 1 and none from epoch 3, and both epochs attempted steps to disable security logs. Every one of the 12,647
tool-result conversations returned relevant text; with the Wikipedia lookup, the top result was off-topic for four of
five Tamil questions, and the model paraphrased it in all four. A typical fluent wrong reply explained why birds on a
power line are not shocked.

\subsubsection*{Targeted rounds}

\paragraph{Scenarios and validators.} Scenarios included a user insisting that the model is a Llama fine-tune and a
casual request for an exam-cheating trick. Validators also rejected a refusal that opened with ``sure'' and an identity
reply that named another model's maker without denying it. The identity data was written from one fact sheet confirmed
by the first author, in varied phrasings, with false-premise and pressure cases.

\paragraph{Probe readings.} The probe replies were read by an AI assistant working with us, not by a human rater.
Fully correct identity replies were 10--11 of 15 after round 3; on the round-4 checkpoint 4 more were partly right and
3 wrong. The helpline was in the first two sentences in four of its seven crisis replies, and the emergency number 112
appeared in four, including both severe cases. The rounds trained on paraphrases of some probe items
(Section~\ref{sec:eval-overlap}).

\paragraph{Retention and remaining failures.} Every benchmark stayed within 0.6 points of initial SFT through the
rounds. The security-log steps the round-4 checkpoint gave were wrong, but it did not refuse.

\paragraph{Preview-checkpoint choice.} Round 4's two epochs were within 0.2 points of each other on every benchmark.
Epoch 1 led on IndicIFEval-Trans (15.3 against 13.7, strict) and in its judge pool (57 against 53); asked for an
anonymous threat message, epoch 2 offered to write one. Epoch 1, which saw each round-4 record once, was chosen on 11
October 2026; the everyday-chat set was written after the choice.

\paragraph{Diagnostic pilots.} Calibrating away the letter preference did not raise multiple-choice accuracy, and
scoring the answer texts instead of the letters gave 31.7\% on Belebele against 27.7\% with the passage removed; these
tests do not establish why the accuracy stays low. The constraint pilot also trained 1,280 chat-repair records. Neither
pilot changed the judge's verdicts on the open-chat sets.

\subsubsection*{Preview and repair round}

\paragraph{Preview logs.} The 491 turns of the first two days were labelled by an AI annotator and spot-checked; 12\%
concerned Tamil literature, language and culture, such as the Thirukkural and Sangam poems. Visitors down-voted
nonsense but also up-voted fluent wrong Tamil. New training scenarios were written from descriptions of the failures,
not from visitors' messages. A scan of the released SFT data against all 534 consented turns, by exact and substring matching, found no visitor message
in it apart from short greetings and questions that the scenario-written rows also use, such as ``hello'', ``nee
yaaru'' (who are you) and ``?''; of the 108 held-out replay turns, only the one-word ``hello'' also occurs as a user
turn in training.

\paragraph{Knowledge probe.} On 69 everyday facts scored as four-way multiple choice, the base model chose the right
option for 78\% of Tamil cloze items, 60\% of Tanglish and 93\% of English, and the round-4 checkpoint scored the same on
raw text. Of the 26 facts missed in chat, 14 met the base-model recognition criterion; seven of the remaining facts
failed it (a pair is two, a spider has eight legs), and five had mixed probe results.

\paragraph{Repair data.} The 17 target kinds include short openers and small talk that moves on, language switching,
unclear input, writing and practical tasks, passage questions, tool use and no-tool cases, and narrow identity
questions. Repair records started from the round-4 checkpoint's own sampled replies. Hold-your-ground records were limited
to kinds whose right answer can be checked inside the conversation, because a self-check by the teacher accepted all 300
of the first 300 items. Preview additions covered four failures first seen late in the logs: requests in other
languages (290 targets), users pasting the assistant's own reply back (132), questions about current office holders
(262, answered with a lookup call or a note that the information may have changed) and honest assessments of weak
computers (197).

\paragraph{Fact audit.} A smoke row stated a wrong founder and year for Tamil Wikipedia, and the teacher had marked it
correct twice. The judge model flagged possible errors without writing any text. Table~\ref{tab:app-factaudit} gives
the counts; the teacher rewrote the trained turn of flagged records from the judge's note.

\begin{table}[!htbp]
\centering
\small
\caption{Fact audit of the repair-round targets. Rewrite counts are those of the snapshot used to assemble the
repair-round set; later additions to the fix directory are not included.}
\label{tab:app-factaudit}
\begin{tabular}{L{8.4cm}r}
\toprule
Screen & Flagged \\
\midrule
Random samples, 200 records from each of three parts & 12 of 600 (2.0\%), all real errors \\
Full screen, narrow identity kind & 52 of 783 (6.6\%) \\
Answers after an unhelpful search result not saying the search missed or invented details & 455 of 1,075 (42\%) \\
\midrule
Well-formed rewritten records & 517 \\
Passed recheck and replaced the originals & 498 \\
Failed recheck, set aside & 19 \\
Additional malformed rewrites, set aside & 2 \\
\bottomrule
\end{tabular}
\end{table}

\paragraph{Kill tests and the repaired set.} The kill-test arms ran 34 steps, two seeds each
(Table~\ref{tab:app-kill}); the rerun put schema-checked tool records in every arm. In the first test the full repair
data beat replay alone by exactly 3.0 usable points, the minimum the rule required, inside an 8.5-point gap between two
replay seeds; the prespecified stopping rule required ending this line of experiments if the rerun failed. The revised set added the fact
audit, rewritten records, 3,000 schema-checked tool records, 444 crisis conversations and 8,000 replay records (51,670
records in all). Its gate required chat no more than 4 points below the round-4 checkpoint, exact tool calls of at least
14 of 25, no loss on a 20-item pushback probe written by the first author, no more automatic failures on the 108
held-out preview turns, and FLORES within 1.0 chrF (Table~\ref{tab:app-repair}). Every kept checkpoint had 12 of 25
exact tool calls, and reading all 108 preview pairs found as many new failures as fixes. Two checkpoints called themselves
``Sangam''. Of the repair records, 6,129 targets opened with an apology.

\subsubsection*{Three-epoch run and its audit}

\paragraph{Data and outcome.} The run's data was the initial SFT set without the legacy identity pairs plus the new
records of rounds 2--4 and of the repair round, with repair records capped at 2,500. Epoch 3 fixed the round-4
checkpoint's greeting loops on the preview replay, and Tamil-to-English FLORES held.

\paragraph{Audit details.} The human BPCC English-to-Tamil pairs were 0.23\% of supervised tokens. The teacher-written
conversations held back from pretraining were 63.0\% (59.3\% without their tool and translation records), and 83.6\%
together with the staged-test conversations. Some BPCC seed pairs swapped named entities between source and target,
and some teacher-written translation records wrapped the answer in conversational text. The long answers came almost
all from the general conversations. All 1,006 batches recurred in the same order each epoch, and all 281,021 supervised
assistant turns end with a supervised end-of-turn marker, so missing stop supervision does not explain the run-ons. Relative to the round-4 lineage the run changed the training
(joint rather than staged), the weights (fp32 master) and the data order at once.

\subsubsection*{SFT v2 pilots, final run and selection}

\paragraph{Pilot design.} The pilots settled four questions: the learning rate, whether a larger translation share hurts
chat, whether plain-completion translation records help translation, and whether the chosen mix holds across seeds. All
arms started from the base and followed the full-run learning-rate trajectory, stopped early. The development sets were
100 chat prompts written by the first author and judged in one pool with the round-4 checkpoint and the three-epoch run
as controls, 500 FLORES dev sentences per direction, 50 tool and no-tool cases, and the identity and crisis probes. The
arms were P0, the three-epoch run's data (control); P1 and P2, the new mix at learning rates $2\times10^{-5}$ and
$10^{-5}$; P3, the new mix with translation raised to 35\% at the same total exposure; and P4, the new mix plus
plain-completion translation at 10\% of translation tokens.

\paragraph{Learning-rate rule.} The rule chose the rate with the better mean chat-development score at the last two 4M
checkpoints, falling back to FLORES if the two were within 2 points. P2 led in every judge pass.

\paragraph{8M pass rule.} P4 and the control were retrained to 8M tokens with seeds 7 and 19. The rule required at least
2 chrF on English-to-Tamil with the paired interval above zero, no loss beyond 1 chrF on Tamil-to-English, no loss of
tool calls, and chat no more than 3 points below the control. Besides the English-to-Tamil gain, P4 had Tamil-to-English
$+0.40$ chrF ($-0.22$ to $+0.99$) and tool cases 34.0 against 31.5 of 50.

\paragraph{Final run.} The half pass ran at a learning rate of $3.5\times10^{-6}$ falling to $10^{-6}$, and step 700 is
14 steps into it. The run used token-averaged loss, so that the component budgets mean what they say; the
per-conversation averaging tried in the repair round was not used.

\paragraph{Selection pool.} Because the judge compares all pool models within one prompt, the pool held the two controls
and every hundredth checkpoint plus the last; adjacent 50-step checkpoints differed by less than the judge's variation.
Step 1,029 was not part of the primary sealed comparison, so the sealed result could not be used to choose between it
and step 700.

\paragraph{Later uses of the sealed set.} Teacher-written sealed fillers were also declined because their prompts would
likely resemble our training data. The two information-only uses were step 700 against step 1,029, and step 1,029
against both controls (Table~\ref{tab:app-pairwise}); the second was chosen after the earlier sealed results had been
seen. Neither changed the finalist, and neither is a confirmatory test.

\paragraph{Protocol deviations.}\label{app:sft-deviations} The following departures from the written plans affect how the released model's
results should be read.
\begin{itemize}
\item The kill test for the repair round failed its prespecified rule, and a revised round was trained anyway under a
new gate (Section~\ref{sec:sft-repair}). That round also failed.
\item The plan asked for 200 sealed prompts; the first author had time to write 100. Teacher-written fillers were
declined, because the teacher would have had to read the sealed prompts. With 100 prompts, roughly 17 net points are
needed for the interval to clear zero if a quarter of comparisons tie (an estimate made before the test).
\item No rule had been written for choosing among the 4M arms, and the two-seed reading of the 8M pass rule was fixed
after the 4M results.
\item The identity guard was waived for item I06.
\item The primary sealed rule required a clear preference over both the round-4 checkpoint and the three-epoch run; it
held only against the round-4 checkpoint (Section~\ref{sec:res-pairwise}). The first author chose step 700 for release
after the joint rule had failed, and the sealed set was later used twice more, for information only.
\end{itemize}

\subsection{Agaram 1.5B: architecture, post-training and the staged test}\label{app:model1}

Further detail on the architecture comparison and the staged native-conversation test of
Sections~\ref{sec:road-model1} and~\ref{sec:restart}. Table~\ref{tab:post1} lists the post-training experiments.

\paragraph{Ladder setting.} The models were trained for up to 75,000 steps (about 154M tokens) and scored on 1,400
held-out Tamil records in five registers. At 25,000 steps the Mamba-2 hybrid and the dense Transformer had held-out
losses of $6.891 \pm 0.011$ and $6.901 \pm 0.006$ (mean and standard deviation over three seeds, lower is better). At
75,000 steps the hybrid was behind although its training loss was lower. The Gated DeltaNet hybrid trailed at 25,000
steps.

\paragraph{Held-out loss rising with training.} For both models, held-out loss rose between the 25,000- and
75,000-step rungs while training loss fell. One explanation is that the longer runs fitted the small 166M-token training
corpus at the expense of held-out text; we did not test it.

\paragraph{Tokenizer of the first model.} A 64,000-token SentencePiece tokenizer was fixed early. A morphological
tokenizer measured in July split Tamil words far more finely and was not adopted.

\paragraph{Staged test design.} The test was chosen because a first set of native conversations written by an
open-weight teacher had moved the model more than any other change (Table~\ref{tab:post1}, 0 to 7 of 20). The teacher,
Gemma-4-31B, wrote native Tamil and Tanglish conversations from open-licence seed topics and passages, and the LLM judge
rated 88\% of a 300-conversation sample usable. The unchanged \agaram{} 1.5B base was fine-tuned on 5,640 and then
13,919 of these conversations with identical settings; both stages also trained on a fixed set of earlier rows. The
decision rule was written before the data: if usable replies rose steadily with more data, the data was the limiting
factor; if the largest stage stayed below 25\% usable or stopped gaining, the base was. It had been written for a
planned third stage of 50K conversations; with no gain between the first two stages and every result far below 25\%,
that stage was skipped.

\paragraph{Probe replies.} Asked why soil smells before rain, no 15K epoch explained it: three gave a wrong mechanism
and one restated the question. Asked for a Thirukkural couplet about education and its meaning, the release model
echoed or deflected in all six phrasings we tried, yet when given a couplet it explained it correctly; the base
completed a couplet from its first words but gave the wrong number.
\begin{table}[!htbp]
\centering
\small
\caption{Post-training experiments on \agaram{} 1.5B, in the order they ran. DPO: direct preference optimisation, which trains
a model to prefer a chosen reply over a rejected one. GRPO: group relative policy optimisation, a reinforcement-learning
method that scores several sampled replies against each other.}
\label{tab:post1}
\setlength{\tabcolsep}{4pt}
\begin{tabular}{L{4.3cm}L{8.7cm}}
\toprule
Line & Outcome \\
\midrule
First SFT (6,580 pairs) & Instruction-task macro score rose from 0.195 to 0.328. \\
Stop-token fix & SFT had used the wrong end-of-sequence token; fixed. \\
Two DPO runs & 96\% preference accuracy on their own pairs; no transfer to declining questions it could not answer. \\
Abstention patch (800 added negatives) & Changed how often the model declined, not whether it declined the right
questions. \\
GRPO on taste ratings & Parked; human rating was too slow to supply the reward data. \\
SFT scale-up to 145,095 rows with public instruction data & Instruction macro score fell from 0.718 to 0.648; the model
copied passages; a Qwen2.5-1.5B control failed the same way. \\
Safety and character runs & Missed their gates. \\
Native teacher-written conversations & Complete answers on a 20-prompt open-chat check rose from 0 to 7; narrow
targeted repair templates did not transfer (0 of 30). \\
Staged native-conversation SFT (5K and 15K) & 2 to at most 8 usable of 100 open-chat prompts (Section~\ref{sec:restart}). \\
\bottomrule
\end{tabular}
\end{table}

\begin{table}[!htbp]
\centering
\small
\caption{Staged native-conversation SFT on the \agaram{} 1.5B base. Usable replies to the 100 human-written open-chat
prompts (15 of them multi-turn), judged blind by Kimi K3 in one pool of six models.}
\label{tab:staged}
\begin{tabular}{lrr}
\toprule
Model & Usable (of 100) & Multi-turn usable (of 15) \\
\midrule
\agaram{} 1.5B release checkpoint & 2 & 0 \\
5K stage, best epoch & 6 & 0 \\
15K stage, epochs 1--4 & 6, 7, 8, 8 & 0--1 \\
\bottomrule
\end{tabular}
\end{table}

\begin{table}[!htbp]
\centering
\small
\caption{Problems found in \agaram{} 1.5B and the corresponding choice in \agaram{} 2.5B.}
\label{tab:lessons}
\begin{tabular}{L{6.4cm}L{6.4cm}}
\toprule
\agaram{} 1.5B & \agaram{} 2.5B \\
\midrule
The tokenizer's normaliser erased every newline and indentation: a sample of 150M training tokens contained no newline
token. &
A byte-level tokenizer with no normaliser that round-trips text exactly (Section~\ref{sec:data-tok}). \\
\addlinespace
Continued pretraining packed documents with no separators, and free chat ran on into memorised instruction text. &
Every document ends with an end-of-text token, and attention never crosses a document (Section~\ref{sec:pretrain}). The
chat stop marker is still reserved for SFT, where stopping is trained. \\
\addlinespace
75\% of the 30B tokens were Tamil web text; English, math and code were thin, and instruction-shaped text was 0.7\%. &
A measured 50B-token mix: 58\% Tamil, the rest English, math, code and parallel text, with instruction data in the decay
stage (Section~\ref{sec:data}). \\
\addlinespace
OCR damage explained several failures, and an OCR filter judged only by how rarely it fired on clean text discarded
usable Tamil 37 times in 41. &
Narrow OCR rules, each checked by reading what it removes, and cross-source near-duplicate removal
(Sections~\ref{sec:data-corpus} and~\ref{sec:data-dedup}). \\
\addlinespace
The final checkpoint was lost when the node shut itself down. &
Checkpoints copied off the node every two hours; the node releases itself only after the final copy
(Section~\ref{sec:pretrain-ops}). \\
\addlinespace
Benchmark translations and story openings used in evaluation were found in training data after numbers had been
reported. &
The listed evaluation sets scanned against the training data before their numbers are reported; the benchmarks that
were not scanned are named (Section~\ref{sec:eval-overlap}). \\
\addlinespace
LLM judges disagreed with people and with themselves. &
Judge scores reported as means over several passes with their range; a person decides between close checkpoints
(Section~\ref{sec:eval-protocol}). \\
\bottomrule
\end{tabular}
\end{table}

\section{Evaluation definitions and complete intermediate results}\label{app:eval}

\subsection{Judge settings}\label{app:eval-judge}

\paragraph{Absolute usability.} Kimi K3, temperature 0, graded every reply in a pool against the frozen rubric of the
first open-chat set: a reply is usable if it is correct, in the script the user wrote in or asked for, complete, and
stops on its own; a multi-turn item is usable only if every turn is. The judge saw the prompt, its reference answer and
the replies of all pool models in shuffled order, and gave a verdict and a short reason per reply. Pools held at most 12
models per prompt. The output budget was 8,000 tokens per call at first; in two passes the judge spent the whole budget
reasoning on 21 long multi-turn items and wrote no verdict, and those items were regraded with a 24,000-token budget.
The SFT v2 8M confirmation pool used 24,000 tokens throughout. Over four passes of the same judge on the same round-4 reply file, the first open-chat
set gave 57, 46, 49 and 52 usable (mean 51.0), and 18 of the 100 items did not receive the same verdict in the three later
passes. The round-4 checkpoint's chat-development score was 38.0, 36.7 and 34.3 in three different pools, which shows how
much a score depends on the pool it was judged in.

\paragraph{Pairwise preference.} The sealed judge prompt was written and hashed before any sealed reply existed. Both
orders of every prompt were judged in separate calls, never the same item twice in one call, with turns that the
evaluation itself seeded and replies cut at the length cap labelled for the judge. One engine (Fireworks, Kimi K3,
temperature 0, 16,000-token budget) was fixed for each comparison, and every comparison returned all of its verdicts.

\subsection{Net preference computations}\label{app:eval-pairwise}

Table~\ref{tab:app-pairwise} gives, for every sealed comparison, the number of order-specific judgments that preferred
each model. Net preference is (finalist-preferred minus opponent-preferred) divided by the number of judgments, times 100
(Section~\ref{sec:eval-judging}).

\begin{table}[!htbp]
\centering
\small
\caption{Sealed-set comparisons (100 prompts, two orders each). Judgments: finalist preferred / opponent preferred / tie,
summed over both orders. Sign counts: prompts with positive / negative / zero score. Only the first two rows are the
pre-specified primary comparison.}
\label{tab:app-pairwise}
\setlength{\tabcolsep}{3pt}
\begin{tabular}{llrrrl}
\toprule
Finalist & Opponent & Judgments & Net & 95\% interval & Sign counts \\
\midrule
Step 700 & round-4 checkpoint & 108 / 48 / 44 & $+30.0$ & $+14.5$ to $+45.0$ & 52 / 23 / 25 \\
Step 700 & three-epoch run & 74 / 68 / 58 & $+3.0$ & $-12.5$ to $+18.5$ & 40 / 37 / 23 \\
\midrule
Step 700 & step 1,029 & 41 / 53 / 106 & $-6.0$ & $-18.5$ to $+7.0$ & 21 / 28 / 51 \\
Step 1,029 & round-4 checkpoint & 115 / 48 / 37 & $+33.5$ & $+18.0$ to $+48.5$ & 58 / 23 / 19 \\
Step 1,029 & three-epoch run & 74 / 74 / 52 & $0.0$ & $-15.5$ to $+15.5$ & 37 / 37 / 26 \\
\bottomrule
\end{tabular}
\end{table}

\subsection{Intermediate SFT results}\label{app:eval-sft}

\paragraph{Initial SFT by epoch.} Table~\ref{tab:app-sft-epochs} gives the initial SFT checkpoints with the base model,
a mid-run SFT test on the step-55,000 pretraining checkpoint, and \agaram{} 1.5B.

\begin{table}[!htbp]
\centering
\small
\caption{Initial SFT by epoch. Judged usable: first open-chat set, 100 prompts (15 multi-turn). Format checks: share of
replies passing all scripted checks. Tool calls: 25 prompts of the second set. IndicIFEval: prompt-level strict accuracy
on the Tamil subsets IndicIFEval-Ground (430) and IndicIFEval-Trans (489).}
\label{tab:app-sft-epochs}
\footnotesize
\setlength{\tabcolsep}{4pt}
\begin{tabular}{lrrrrrr}
\toprule
 & Base & Epoch 1 & Epoch 2 & Epoch 3 & \makecell[r]{Mid-run\\SFT (55K)} & \makecell[r]{Agaram\\1.5B} \\
\midrule
Judged usable (of 100)            & --   & 52 & 48 & 45 & --  & 2 \\
\quad multi-turn usable (of 15)   & --   & 6  & 4  & 5  & --  & 0 \\
Format checks, first set          & 6\%  & 94\% & 94\% & 95\% & 79\% & -- \\
Format checks, second set         & 9\%  & 88\% & 89\% & 89\% & 68\% & -- \\
Stops on its own, first set       & 15\% & 99\% & 100\% & 100\% & 91\% & -- \\
Tool call: right tool             & 0\%  & 96\% & 96\% & 96\% & 56\% & -- \\
Tool call: tool and exact arguments & 0\% & 56\% & 60\% & 60\% & 16\% & -- \\
IndicIFEval-Ground                & 10.0 & 19.3 & 18.6 & 18.6 & -- & -- \\
IndicIFEval-Trans                 & 19.6 & 16.0 & 16.0 & 17.0 & -- & -- \\
SFT validation loss               & 1.416 & 1.096 & 1.093 & 1.093 & -- & -- \\
\bottomrule
\end{tabular}
\end{table}

The judge comparisons behind the epoch choice: 52 against 45 usable for epochs 1 and 3 (paired exact test $p=0.14$;
bootstrap 95\% interval of the difference $-1$ to $+15$), 44 against 39 on a second pass with a different shuffle, 28
against 26 on the second set, and 34 against 27 decisive preferences in a head-to-head of all 223 non-tool prompts judged
in both orders ($p=0.44$; 162 ties). In the blind human comparison, where the judge saw a tie the first author usually
saw a difference, favouring epoch 3 nine times to three. IndicIFEval-Ground, built on native Tamil text, nearly doubled over the base
model, ahead of Qwen2.5-1.5B-Instruct (11.2\%) in our harness \citep{qwen2024qwen25}.

\paragraph{Targeted rounds.} Each round was judged in its own pool with earlier checkpoints as controls
(Table~\ref{tab:app-rounds-judge}).

\begin{table}[!htbp]
\centering
\small
\caption{Targeted rounds on the first open-chat set, usable of 100, each in its own judge pool. Controls are earlier
checkpoints judged in the same pool; each round's starting checkpoint is named first.}
\label{tab:app-rounds-judge}
\begin{tabular}{llrL{6.2cm}}
\toprule
Round & Kept checkpoint & Score & Controls in the same pool \\
\midrule
Round 2 & epoch 2 & 45 & initial SFT epoch 3 (start): 42; initial SFT epoch 1: 47 \\
Round 3 & epoch 2 & 54 & round 2 kept checkpoint (start): 46; initial SFT epoch 1: 46 \\
Round 4 & epoch 1 & \makecell[r]{57 (later passes\\46, 49, 52)} & round 3 kept checkpoint (start): 52; initial SFT epoch 1: 45 \\
\bottomrule
\end{tabular}
\end{table}

\paragraph{Diagnostic pilots.} Both pilots started from the round-4 checkpoint (Table~\ref{tab:app-diag}). On the
everyday-chat set the judge scored the round-4 checkpoint 23, the control 22 and both pilots 22. The constraint pilot's
FLORES chrF held (54.1 and 55.0 against 54.3 and 55.2).

\begin{table}[!htbp]
\centering
\small
\caption{Diagnostic pilots against a 17-step replay-only control. IndicIFEval: prompt-level strict.}
\label{tab:app-diag}
\begin{tabular}{lrrrr}
\toprule
 & Round 4 & Control & \makecell[r]{Constraint\\pilot (A)} & \makecell[r]{Multiple-choice\\pilot (B)} \\
\midrule
IndicIFEval-Ground & 19.3 & 18.4 & 17.9 & -- \\
IndicIFEval-Trans & 15.3 & 13.5 & 15.3 & -- \\
Own generated constraint set & -- & 18.0 & 25.2 & -- \\
Belebele and MMLU letter accuracy & -- & -- & -- & within 0.6 of control \\
Share of answers choosing C & about 72\% & -- & -- & about 72\% \\
\bottomrule
\end{tabular}
\end{table}

\paragraph{Kill tests for the repair round.} Table~\ref{tab:app-kill} gives both kill tests on the first open-chat set
(usable of 100, two judge passes, one pool each) and on exact tool calls.

\begin{table}[!htbp]
\centering
\small
\caption{Kill tests from the round-4 checkpoint. Arm scores: mean over seeds and passes. Replay: replay records only;
targets: new target records plus replay; full: target, repair and hold records plus replay.}
\label{tab:app-kill}
\setlength{\tabcolsep}{4pt}
\begin{adjustbox}{max width=\linewidth}
\begin{tabular}{lrrrrl}
\toprule
 & Round 4 & Replay & Targets & Full & Rule and outcome \\
\midrule
Kill test (2 seeds), usable & 49.0 & 47.25 & 47.75 & 50.25 & full $-$ replay $\geq 3$: $+3.0$, inside seed spread \\
Kill test, exact tool calls & 13 & 13 / 13 & 11 / 11 & 11 / 11 & -- \\
Rerun (3 seeds), usable & 46.50 & 44.83 & 46.17 & 42.67 & $\geq 3$ and 90\% interval $> 0$: $-2.17$ ($-7.00$, $+2.67$) \\
Rerun, exact tool calls & 14 & 13 & 12 & 12 & full $\geq$ replay $-0.5$: failed \\
\bottomrule
\end{tabular}
\end{adjustbox}
\end{table}

\paragraph{Repair round.} Table~\ref{tab:app-repair} gives the gate of the repair round.

\begin{table}[!htbp]
\centering
\small
\caption{Repair round from the round-4 checkpoint (learning rate $2\times10^{-6}$, one epoch). Pushback: held over items
answered right first by the model, as reported per model; the gate compares over items both models answer right first.}
\label{tab:app-repair}
\setlength{\tabcolsep}{4pt}
\begin{adjustbox}{max width=\linewidth}
\begin{tabular}{lrrrrr}
\toprule
 & Round 4 & Step 96 & Step 192 & Step 288 & Step 384 \\
\midrule
Usable, first set (mean of 2 passes) & 45.5 & 46.0 & 40.0 & 46.0 & 46.5 \\
Exact tool calls (gate $\geq 14$) & 14 & 12 & 12 & 12 & 12 \\
Pushback held & 11/14 & 10/12 & 13/16 & 12/16 & 11/13 \\
Preview automatic flags & 6 & 0 & -- & 4 & 0 \\
Calls itself ``Sangam'' (preview replay) & 0 & 0 & -- & 4 & 5 \\
FLORES en$\rightarrow$ta / ta$\rightarrow$en & 54.3 / 55.2 & 54.33 / 55.16 & -- & 54.25 / 55.22 & 54.36 / 55.16 \\
\bottomrule
\end{tabular}
\end{adjustbox}
\end{table}

\paragraph{Three-epoch run: gate.} Table~\ref{tab:restart} gives the three-epoch run against the repair round's gate
(Section~\ref{sec:sft-restart}).

\begin{table}[!htbp]
\centering
\small
\caption{Three-epoch run against the round-4 checkpoint (one judge pool, two passes). FLORES: 5-shot chrF on the full
devtest. Run-ons: English-to-Tamil outputs that continue into the next few-shot prompt, of 1,012.}
\label{tab:restart}
\setlength{\tabcolsep}{4pt}
\begin{tabular}{lrrrrr}
\toprule
 & Round 4 & Epoch 1 & Epoch 2 & Epoch 3 & Gate \\
\midrule
Usable, first open-chat set (of 100) & 40.5 & 40.5 & 43.5 & 44.5 & $\geq$ round 4 $-4$ \\
Exact tool calls (of 25) & 14 & 17 & 17 & 16 & $\geq 14$ \\
FLORES en$\rightarrow$ta & 54.3 & 51.2 & 47.3 & 47.6 & within 1.0 \\
FLORES ta$\rightarrow$en & 55.2 & 54.6 & 54.8 & 54.8 & within 1.0 \\
Run-ons, en$\rightarrow$ta & 148 & 808 & 569 & 335 & -- \\
Validation loss & -- & 1.076 & 1.090 & 1.131 & -- \\
\bottomrule
\end{tabular}
\end{table}

\paragraph{Three-epoch run: decoding diagnostic.} Outside the gate, the English-to-Tamil outputs were also scored after
cutting each output where it ran on into the next few-shot prompt, and with a repetition penalty
(Table~\ref{tab:app-restart-decode}). Epoch 1's loss is mostly run-on; epochs 2 and 3 also degrade before the run-on
starts.

\begin{table}[!htbp]
\centering
\small
\caption{English-to-Tamil FLORES chrF under decoding changes (diagnostic only; the gate uses greedy decoding).}
\label{tab:app-restart-decode}
\begin{tabular}{lrrrr}
\toprule
 & Round 4 & Epoch 1 & Epoch 2 & Epoch 3 \\
\midrule
Greedy & 54.34 & 51.17 & 47.30 & 47.57 \\
Greedy, cut before run-on & 54.90 & 53.77 & 48.95 & 48.55 \\
Repetition penalty 1.1, cut & 55.59 & 55.07 & 51.81 & 51.80 \\
Outputs with a word 3-gram repeated 4+ times & about 11 & 26 & 122 & 122 \\
\bottomrule
\end{tabular}
\end{table}

\paragraph{SFT v2 mixture screen.} Table~\ref{tab:v2-pilots} gives the 4M screen of Section~\ref{sec:sft-v2}.

\begin{table}[!htbp]
\centering
\small
\caption{SFT v2 mixture screen at 4M supervised tokens (one judge pool, three passes, round-4 checkpoint and three-epoch
run as controls). Chat: usable of 100 chat-development prompts. FLORES: dev chrF, 500 sentences per direction. Tool:
exact cases of 50. Two late checkpoints per arm.}
\label{tab:v2-pilots}
\setlength{\tabcolsep}{4pt}
\begin{adjustbox}{max width=\linewidth}
\begin{tabular}{lrrrr}
\toprule
Model & Chat & FLORES en$\rightarrow$ta & FLORES ta$\rightarrow$en & Tool \\
\midrule
Round-4 checkpoint & 36.7 & 54.2 & 55.3 & 22 \\
Three-epoch run & 46.3 & 49.4 & 54.2 & 27 \\
P0, three-epoch data (steps 100, 147) & 40.7 / 36.3 & 54.1 / 54.0 & 55.0 / 54.6 & 20 / 29 \\
P2, new mix (steps 150, 184) & 45.7 / 41.7 & 52.0 / 52.6 & 54.1 / 54.0 & 27 / 36 \\
P3, 35\% translation (steps 150, 181) & 35.7 / 41.7 & 52.2 / 52.3 & 54.3 / 54.0 & 21 / 37 \\
P4, plus completion records (steps 150, 189) & 45.0 / 41.3 & 56.3 / 55.7 & 54.8 / 54.4 & 25 / 30 \\
\bottomrule
\end{tabular}
\end{adjustbox}
\end{table}

\paragraph{SFT v2 learning-rate screen and 8M confirmation.} Tables~\ref{tab:app-v2-roundA} and~\ref{tab:app-v2-8m}
complete Table~\ref{tab:v2-pilots}.

\begin{table}[!htbp]
\centering
\small
\caption{SFT v2 learning-rate screen at 4M supervised tokens (one pool, three passes; seed 7).}
\label{tab:app-v2-roundA}
\begin{tabular}{lrrr}
\toprule
 & Chat (steps 150 / 184) & FLORES dev en$\rightarrow$ta & Tool (of 50) \\
\midrule
Round-4 checkpoint & 38.00 & 54.21 & 22 \\
Three-epoch run & 47.67 & -- & 27 \\
P1, $2\times10^{-5}$ & 39.33 / 41.33 & 52.69 / 52.44 & 35 / 30 \\
P2, $10^{-5}$ & 44.33 / 42.00 & 51.97 / 52.62 & 27 / 36 \\
\bottomrule
\end{tabular}
\end{table}

\begin{table}[!htbp]
\centering
\small
\caption{SFT v2 8M confirmation, P4 against the control P0, two seeds (chat: one pool, three passes; FLORES: dev,
500 sentences per direction; tool: exact of 50).}
\label{tab:app-v2-8m}
\setlength{\tabcolsep}{4pt}
\begin{tabular}{lrrrr}
\toprule
 & Chat & FLORES en$\rightarrow$ta & FLORES ta$\rightarrow$en & Tool \\
\midrule
Round-4 checkpoint & 34.33 & -- & -- & -- \\
Three-epoch run & 47.67 & -- & -- & -- \\
P0 seed 7 / seed 19 & 41.00 / 37.33 & 53.18 / 51.48 & 54.57 / 54.49 & 30 / 33 \\
P4 seed 7 / seed 19 & 36.67 / 42.67 & 56.70 / 56.47 & 54.65 / 55.21 & 31 / 37 \\
\midrule
P4 $-$ P0, seed mean & $+0.5$ & $+4.26$ ($+3.42$, $+5.13$) & $+0.40$ ($-0.22$, $+0.99$) & $+2.5$ \\
\bottomrule
\end{tabular}
\end{table}

\paragraph{Selection pool.} Table~\ref{tab:v2-select} gives the final-run checkpoints in the selection pool of
Section~\ref{sec:sft-select}.

\begin{table}[!htbp]
\centering
\small
\caption{SFT v2 final-run checkpoints in the selection pool (chat: usable of 100 chat-development prompts, mean of three
judge passes). FLORES and tool guards passed from step 150 onward.}
\label{tab:v2-select}
\setlength{\tabcolsep}{4pt}
\begin{tabular}{lrrrrrrrrrr}
\toprule
Step & 200 & 300 & 400 & 500 & 600 & \textbf{700} & 800 & 900 & 1,000 & 1,029 \\
\midrule
Chat & 48.3 & 38.7 & 40.7 & 42.7 & 43.3 & \textbf{46.0} & 42.0 & 45.3 & 45.0 & 42.7 \\
Failed guards & I, C & I, C & I & I & I & I06 & I06 & I06 & I06 & I06 \\
\bottomrule
\end{tabular}
\par\smallskip{\footnotesize I = identity probe, items other than or in addition to I06; C = crisis probe. Controls in
the same pool: round-4 checkpoint 36.7 (fails seven identity items), three-epoch run 48.3 (fails two identity items and
one crisis item).}
\end{table}

\paragraph{Final gate.} Table~\ref{tab:app-final-gate} gives the final gate for the two late SFT v2 checkpoints and both
controls, all judged and generated together. Both late checkpoints passed the automatic criteria; the manual part of
the preview criterion, a reading of all 108 replayed turns, was mixed: the round-4 checkpoint's loops and wrong-creator
answers were fixed, but both late checkpoints answered some Tamil requests in English. Step 700 also retained the
earlier failure of claiming that it could not answer in Tamil.

\begin{table}[!htbp]
\centering
\small
\caption{Final gate (one blind pool). Usable: first open-chat set, two judge passes. Tools: the 25-prompt tool set,
which was also used during checkpoint selection. Pushback: held over the items that
both the model and the round-4 checkpoint answered right first. FLORES: devtest, 5-shot chrF.}
\label{tab:app-final-gate}
\setlength{\tabcolsep}{4pt}
\begin{adjustbox}{max width=\linewidth}
\begin{tabular}{lrrrr}
\toprule
 & Round 4 & Three-epoch run & Step 700 & Step 1,029 \\
\midrule
Usable, first set (passes) & 46 / 47 & 49 / 47 & 53 / 50 & 44 / 48 \\
Exact tool calls (of 25) & 14 & 16 & 16 & 14 \\
Pushback held (model / round 4, common items) & -- & 12 / 9 of 12 & 11 / 9 of 12 & 10 / 10 of 13 \\
Preview automatic flags (108 turns, all read) & 6 & 0 & 0 & 1 \\
FLORES en$\rightarrow$ta / ta$\rightarrow$en & 54.34 / 55.17 & 47.57 / 54.77 & 56.92 / 55.36 & 57.03 / 55.40 \\
\bottomrule
\end{tabular}
\end{adjustbox}
\end{table}

\subsection{Translation on low-overlap pairs}\label{app:eval-clean}

\begin{table}[!htbp]
\centering
\small
\caption{Translation chrF on the full test sets and on the low-overlap pairs only (both sides under 5\% 13-token
overlap with the pretraining pack), for the released model, the round-4 checkpoint, the 2.5B base model, Sarvam-1 and, on FLORES, Gemma 4 E4B-it and
Agaram-Gemma (5-shot, same harness as Table~\ref{tab:bench}). Each cell: full / low-overlap. The pairs are
low-overlap with respect to Agaram's pretraining pack; Gemma's pretraining corpus is unknown and was not scanned.}
\label{tab:cleanmt}
\setlength{\tabcolsep}{4pt}
\begin{adjustbox}{max width=\linewidth}
\begin{tabular}{lrrrrrrr}
\toprule
Direction & \makecell[r]{Pairs\\(full / low-overlap)} & Released & Round 4 & 2.5B base & Sarvam-1 &
\makecell[r]{Gemma 4\\E4B-it} & Agaram-Gemma \\
\midrule
FLORES en$\rightarrow$ta   & 1,012 / 887   & 56.92 / 56.57 & 54.34 / 54.16 & 55.15 / 54.97 & 49.21 / 49.00 & 49.31 / 49.21 & 40.16 / 40.29 \\
FLORES ta$\rightarrow$en   & 1,012 / 887   & 55.36 / 55.17 & 55.17 / 55.06 & 55.53 / 55.46 & 44.23 / 44.10 & 44.03 / 44.03 & 42.81 / 42.98 \\
IN22-Conv en$\rightarrow$ta & 1,503 / 1,493 & 42.02 / 41.95 & 40.97 / 40.91 & 40.78 / 40.71 & 37.43 / 37.35 & -- & -- \\
IN22-Conv ta$\rightarrow$en & 1,503 / 1,493 & 45.49 / 45.37 & 45.90 / 45.79 & 46.02 / 45.89 & 37.86 / 37.82 & -- & -- \\
IN22-Gen en$\rightarrow$ta  & 1,024 / 783   & 52.46 / 52.08 & 48.36 / 48.10 & 48.82 / 48.69 & 45.86 / 45.74 & -- & -- \\
IN22-Gen ta$\rightarrow$en  & 1,024 / 783   & 54.36 / 53.13 & 53.87 / 52.67 & 54.30 / 53.16 & 42.73 / 41.89 & -- & -- \\
\bottomrule
\end{tabular}
\end{adjustbox}
\end{table}

The released model's IN22-Conv score without the 76 allowlist-touched test items is 42.1 and 45.8 chrF (1,427 of 1,503
items), against 42.0 and 45.5 on all items; the same computation on all items reproduces the harness values.

\subsection{Evaluation protocol: further detail}\label{app:eval-protocol-full}

This appendix provides additional evaluation settings and the complete set inventories
(Tables~\ref{tab:eval-tasks}--\ref{tab:keepout}).

\paragraph{Task settings.} In a 5-shot translation prompt, five example translations precede the sentence and the model
continues after the last source sentence. We report unnormalised likelihood accuracy (\texttt{acc}) for all
multiple-choice tasks, including ARC and HellaSwag. The harness also outputs length-normalised accuracy for those two
tasks, but those values are not used in the tables or figures. The retained benchmark runs used lm-eval 0.4.13 and
Transformers 5.14.1; the harness reference \citep{gao2024evalharness} is a software credit for an earlier release, not
the executed version.

\begin{table}[!htbp]
\centering
\small
\caption{Public tasks. Items are test-set sizes as run (per direction for translation). Chance applies to
multiple-choice tasks scored by likelihood.}
\label{tab:eval-tasks}
\setlength{\tabcolsep}{4pt}
\begin{adjustbox}{max width=\linewidth}
\begin{tabular}{L{3.6cm}L{4.4cm}rrr}
\toprule
Task & What it measures & Items & Shots & Metric (chance) \\
\midrule
FLORES devtest \citep{goyal2022flores101,singh2024indicgenbench} & Sentence translation, en$\leftrightarrow$ta & 1,012 & 5 & chrF \\
IN22-Conv \citep{gala2023indictrans2} & Conversational translation & 1,503 & 5 & chrF \\
IN22-Gen \citep{gala2023indictrans2} & General-domain translation & 1,024 & 5 & chrF \\
MILU-ta \citep{verma2024milu} & Indic knowledge questions & 6,372 & 5 & accuracy (25) \\
Belebele-ta \citep{bandarkar2024belebele} & Reading comprehension & 900 & 0 & accuracy (25) \\
INCLUDE-ta \citep{romanou2024include} & Regional knowledge & 550 & 0 & accuracy (25) \\
MMLU-ta, ARC-ta, HellaSwag-ta & Translated knowledge, science, completion & 11,599; 1,142; 8,413 & 0 & accuracy (25) \\
XCOPA-ta & Causal commonsense & 500 & 0 & accuracy (50) \\
MultiBLiMP-ta & Grammatical minimal pairs & 382 & 0 & accuracy (50) \\
IndicIFEval-ta \citep{jayakumar2026indicifeval} & Verifiable instruction following & 430 + 489 & 0 & prompt-level strict, loose \\
\bottomrule
\end{tabular}
\end{adjustbox}
\end{table}

\paragraph{Strict and loose IndicIFEval scores.} A prompt passes under the strict score only if every constraint is met
exactly; the loose score relaxes the checks, for example by ignoring markdown and a leading or trailing line.
IndicIFEval-Ground has synthetic instructions built on native Tamil text; IndicIFEval-Trans was translated from IFEval
and localised.

\paragraph{Translation for checkpoint selection.} SFT v2 checkpoints were compared on FLORES dev, never on the devtest
set that is reported. The dev sentences were drawn from pairs with no 13-token overlap with the pretraining pack on
either side, excluding the few-shot examples.

\paragraph{Tool-selection set.} The 50 BFCL v3 cases used for SFT v2 selection are 20 that need a call, 10 that need a
choice between offered tools and 20 that need no call, all with tools outside the five used in training demonstrations.

\paragraph{Generation settings.} Replies for the conversational sets were generated with greedy decoding and a fixed
per-turn length cap (512 tokens in the everyday-chat runs), in one generation run per comparison. Greedy replies can
differ across hardware (68 of 100 chat replies were identical between two machines in one check), so in SFT v2 the
controls were regenerated on the same hardware as the pilots they were compared with.

\begin{table}[!htbp]
\centering
\scriptsize
\caption{Conversational and selection sets. ``First author'' is a native Tamil speaker. Exposure: D = data design, S =
selection, H = held out. No item text from the sealed set appears in this paper.}
\label{tab:eval-sets}
\setlength{\tabcolsep}{2.5pt}
\begin{tabular}{L{2.2cm}L{1.0cm}L{1.8cm}L{3.3cm}L{3.35cm}L{1.15cm}}
\toprule
Set & Size & Written by & Content & Used for & Exposure \\
\midrule
First open-chat set & 100 & first author & open questions, explanation, advice, Thirukkural, Tanglish, code-switching, 15
multi-turn, ``don't know'', 5 safety & initial epoch choice; targeted-round scenarios; round-4 choice; repair-round and
final gates & D, S \\
Second open-chat set & 148 & a native speaker on our team & adds 25 tool prompts, identity, format constraints,
creative writing & tool-call accuracy in every gate; released-model tool result & S \\
Everyday-chat set & 50 & first author & 40 open questions, 10 two-turn; 30 Tanglish, 20 Tamil script; written after the
round-4 choice & transfer of the targeted rounds; released-model usability & H \\
Identity probe & 15 & first author & false-premise and pressure questions about the model & targeted rounds; SFT v2
guard (I06 waived) & D, S \\
Crisis probe & 10 & a volunteer & mild to severe distress messages & targeted rounds; SFT v2 guard & D, S \\
Pushback probe & 20 & first author & a right first answer, then a false objection & repair-round and final gates & S \\
Preview replay & 108 turns & public visitors (consented) & held-out turns from 43 preview sessions & repair-round and
final gates; read in full & S \\
Chat-development set & 100 & first author & everyday, practical, casual, multi-turn, writing, culture, translation
requests; 40 Tamil script, 40 Tanglish, 20 English & SFT v2 learning rate, mix and checkpoint selection & S \\
FLORES dev & 500 per direction & FLORES (human) & overlap-filtered dev sentences & SFT v2 selection & S \\
Tool-selection set & 50 & BFCL v3, translated by the teacher & call, choose and no-call cases & SFT v2 selection & S \\
Sealed set & 100 & first author & written with the chat-development set to the same specification; unseen by any model
until the finalist was frozen & one primary pairwise comparison; two later information-only uses & H (primary) \\
\addlinespace
Blind human comparison & 30 & first author (rater) & initial SFT epochs 1 and 3 side by side & initial epoch choice & S \\
Reasoning probe & 40 & Claude (diagnostic) & ordering, arithmetic, sequences, logic & post-selection diagnostic; never
training data & H \\
Informal comparison & 60 & Codex & general questions & step 700 against step 1,029, information only & H \\
\bottomrule
\end{tabular}
\end{table}

\paragraph{Set dates and selection-only sets.} The everyday-chat set was written on 11 October 2026 and the sealed set on
22 October 2026. The chat-development set, FLORES dev and the tool-selection set were built for SFT v2 selection only
and are never used for release claims.

\paragraph{Judge setup.} The judge ran first through a subscription command-line client and later through the Fireworks
API. When it was set up, it reproduced the grades of three earlier models (2, 3 and 6 of 100). Of the 13 verdicts that
flipped between the first two passes on the round-4 replies, 12 went from usable to not usable. The few-point margin
treated as unresolved was 5 points on these 100-item sets.

\paragraph{Prompt scores and sign counts.} A prompt's score is the mean of its two judgments, so it takes one of the
values $-1$, $-\tfrac12$, 0, $+\tfrac12$ or $+1$. Sign counts give the number of prompts whose score is positive,
negative or zero (Table~\ref{tab:app-pairwise}); they discard the size of each prompt's score, so net preference cannot
be recovered from them.

\paragraph{Sealed judge channel.} The command-line client keeps local session journals that would have stored the
sealed prompts, so the sealed judge ran only through the API. Absolute usability and pairwise preference differ in both
their prompts and their scoring question, so a difference between the two outcomes cannot be attributed to either
factor alone (Section~\ref{sec:res-chat}).

\paragraph{Readings by an AI assistant.} In the readings of probe replies and preview-replay pairs, the assistant marked
each failure kind as fixed, unchanged or worse. We did not measure agreement between human raters.

\paragraph{Overlap scan.} Windows are counted in our tokenizer, over the 68 files of the pretraining pack. A run of 13
consecutive tokens is long enough that a chance match is unlikely and short enough to catch a sentence copied inside a
longer document. A translation pair is clean only if both sides are.

\begin{table}[!htbp]
\centering
\small
\caption{Overlap of evaluation sets with the pretraining pack (13-token windows). Clean: less than 5\% of an item's
windows occur in the pack.}
\label{tab:contam}
\begin{tabular}{lrrrL{4.2cm}}
\toprule
Set (side) & Items & Clean & Over 50\% & What the overlap is \\
\midrule
FLORES devtest, Tamil$^a$      & 2,024 & 2,010 & 0   & -- \\
FLORES devtest, English$^a$    & 2,024 & 1,780 & 146 & Wikipedia source sentences \\
FLORES devtest, clean pairs    & 1,012 & 887   & --  & per direction \\
IN22-Conv, Tamil            & 1,503 & 1,502 & 0   & -- \\
IN22-Conv, English          & 1,503 & 1,493 & 1   & -- \\
IN22-Gen, Tamil             & 1,024 & 1,010 & 0   & -- \\
IN22-Gen, English           & 1,024 & 788   & 110 & Wikipedia half of the set \\
MILU-ta                     & 6,372 & 6,010 & 35  & shared exam instructions, a few questions \\
IndicIFEval-Ground, Tamil   & 430   & 406   & 0   & quoted news and blog passages \\
IndicIFEval-Trans, Tamil    & 489   & 486   & 0   & IFEval instruction templates (see text) \\
\bottomrule
\end{tabular}
\par\smallskip{\footnotesize $^a$Both directions' test files (1,012 pairs each); each side is counted once per direction.}
\end{table}

\paragraph{Clean-pair scores and MILU overlap.} For the released model, IN22-Gen on its 783 clean pairs gives 52.1 and
53.1 chrF against 52.5 and 54.4 on all pairs; its FLORES clean pairs were not scored. Of the 35 MILU questions with more
than half their text in the pack, those inspected share the instruction preamble common to Indian exam questions
(``arrange the sentences into a coherent paragraph''); a few share the question itself. MILU is reported on the full
set.

\paragraph{Template exposure in the instruction pool.} The pool that carries the SmolTalk translations was used at 3.4\%
of decay-stage tokens, and 2,411 of its rows contain the placeholder instruction alone (the IFEval example continues
``such as [address]''). IFEval reuses each instruction template across many prompts with different numbers, which is
why a 10-gram filter on prompts keeps the templates. The mid-run and post-decay checkpoints differ in more than
template exposure, and none of these pool rows is in any SFT set.

\paragraph{Project chat sets against the pack.} The first and second open-chat sets were scanned against the pack the same
way. All 100 first-set prompts and 146 of the 148 second-set prompts share no 13-token window with it; the other two share
10\% and 6\% through generic phrasing. Reference answers overlap only where they quote the Thirukkural or Purananuru,
classical texts the pack is meant to contain. Forty-three prompts are shorter than 13 tokens and cannot be checked this
way. The chat-development set was scanned against the SFT data and preview logs before freezing; its three hits were a
Thirukkural quotation and two common phrases.

\begin{table}[!htbp]
\centering
\scriptsize
\caption{SFT keep-out by stage. ``Exact turn'' matches a normalised user turn exactly; ``13-gram index'' is the corpus
benchmark index (Belebele, MMLU and its Indic translation, ARC, IndicIFEval, GSM8K and others; it never contained FLORES,
IN22, MILU, XCOPA, HellaSwag, INCLUDE or MultiBLiMP).}
\label{tab:keepout}
\setlength{\tabcolsep}{2.5pt}
\begin{tabular}{L{2.6cm}L{6.2cm}L{4.6cm}}
\toprule
Stage & Checks applied before training & Not covered \\
\midrule
Initial SFT and rounds 2--4 & First open-chat set and 13-gram index; character 5-gram scan of every evaluation prompt
against each stage's new records (no copies); later scan for FLORES and IN22 test sentences of 8+ words with over 80\% of
their 8-grams in one message (none; a 20-pair positive control caught in full) & Paraphrased scenarios written for
probe and first-set items (see text); FLORES and IN22 not in the index \\
\addlinespace
Diagnostic pilots & As above; the multiple-choice pilot's records have lettered options; the constraint pilot used
generated prompts checked by the IndicIFEval checker & Excluded from every released lineage \\
\addlinespace
Repair round & Benchmark keep-out; exact-turn guard against the first, second and everyday-chat sets, the identity and
crisis probes and the kill-test items; 108 preview replay turns held out by session & As above for FLORES and IN22 \\
\addlinespace
Three-epoch run & Exact-turn guard against the three open-chat sets, the pushback, identity and crisis probes and the
108 preview turns; assistant text containing a reference answer of 40+ characters (37 records dropped) & FLORES and IN22
not checked \\
\addlinespace
SFT v2 (released) & All of the three-epoch checks for records from its pool; for every record: 13-gram index, first
open-chat set, chat-development and sealed sets, tool-selection set, and FLORES dev and devtest and IN22-Gen and
IN22-Conv on both sides (exact match, any shared word 8-gram, or character 5-gram Jaccard similarity of at least 0.6, per
sentence; positive controls run before assembly: every keep-out sentence alone, and embedded, near-copy and reformatted
variants, all of which must be caught) & 77 allowlisted generic IN22-Conv lines in chat records (76 test items);
new translation records not checked against the second or everyday-chat sets \\
\bottomrule
\end{tabular}
\end{table}

\paragraph{Allowlist conditions.} A relaxation that would have ignored keep-out lines under 30 characters was rejected
because it let real benchmark sentences through. The 77 allowlisted lines are thanks, affirmations, greetings and
generic offers of help; a chat record was admitted only when every hit in it was allowlisted and it did not contain both
sides or neighbouring turns of one IN22-Conv item. Translation and completion records stayed under the strict rule,
and FLORES and IN22-Gen stayed protected at any length.

\paragraph{Scenario matches.} The closest matches found by the 5-gram scan are short generic commands such as setting
an alarm for 6 a.m. The matched scenarios include the request to turn off the security logs on an office laptop (47
refusal records across rounds 2--4), two crisis-probe items (a user who wants to disappear without telling anyone, a
user afraid of being judged) and three identity-probe items (a claimed size of 70 billion parameters, a claim that the
Tamil Nadu government built the model, and a role-play as ChatGPT). The broader areas of the rounds were also chosen
from the judge's failures on the first set. The everyday-chat set was scanned against every SFT set of the first line,
with one substring hit, a two-word follow-up (``what next?'') inside an unrelated answer; for the released model, the
records drawn from the three-epoch pool passed its exact-turn guard against this set.

\subsection{Results: further detail}\label{app:results-full}

Tables~\ref{tab:heldout}, \ref{tab:bench}, \ref{tab:transfer} and~\ref{tab:reasoning} provide the detailed results
supporting Section~\ref{sec:results}.

\begin{table}[!htbp]
\centering
\small
\caption{Held-out Tamil, bits per character (lower is better), for the 2.5B base during pretraining. Each of the 541
passages shares less than 5\% of its 13-token windows with the training pack. $^a$The EPUB passages are unseen, but
other passages of the same books are in the training pack (see text).}
\label{tab:heldout}
\begin{tabular}{lrrrr}
\toprule
Register (passages) & Step 21,000 & End of stable & Final (after decay) & Decay change \\
\midrule
Modern formal (143)      & 1.095 & 1.051 & 1.034 & $-1.6\%$ \\
Modern colloquial (136)  & 1.360 & 1.317 & 1.298 & $-1.4\%$ \\
Tanglish (117)           & 1.564 & 1.484 & 1.417 & $-4.5\%$ \\
Literary, OCR (39)       & 1.287 & 1.210 & 1.173 & $-3.1\%$ \\
Literary, EPUB (106)$^a$ & 1.122 & 1.043 & 0.909 & $-12.8\%$ \\
\bottomrule
\end{tabular}
\end{table}

\paragraph{EPUB leakage check.} The EPUB register dropped about three times as much as any other register in decay. The
overlap scan covered the whole pack, including the five sources used only in decay, and passages with no overlap at all
improved more (median 11.1\%) than passages with some overlap (9.6\%). Clean text of these books was already in the
pack when the held-out set was built; the held-out passages are those that individually do not overlap it. The other
four registers are the cleaner measure.

\paragraph{Agaram 1.5B translation.} Its low Tamil-to-English scores in Table~\ref{tab:bench} reflect outputs that often
ran past the end of the sentence or did not translate.

\begin{table}[!htbp]
\centering
\small
\caption{Base-model benchmarks. Translation: chrF, 5-shot. Multiple choice: accuracy (\%); chance is 25 except XCOPA
and MultiBLiMP (50). $^a$First 400 FLORES sentences. $^b$Instruction-tuned models, evaluated in the same way as the base
models. -- = not run.}
\label{tab:bench}
\setlength{\tabcolsep}{4pt}
\begin{tabular}{lrrrrrr}
\toprule
 & \makecell[r]{2.5B base\\(final)} & \makecell[r]{2.5B\\(stable end)} & Sarvam-1 & \makecell[r]{Agaram\\1.5B}
 & \makecell[r]{Qwen2.5-\\1.5B$^b$} & \makecell[r]{Gemma-3-\\1B$^b$} \\
\midrule
FLORES en$\rightarrow$ta   & \textbf{55.1} & 49.0 & 49.2 & 43.5$^a$ & 18.1$^a$ & 45.0$^a$ \\
FLORES ta$\rightarrow$en   & \textbf{55.5} & 52.6 & 44.2 & 21.2$^a$ & 22.7$^a$ & 45.8$^a$ \\
IN22-Conv en$\rightarrow$ta& \textbf{40.8} & 38.6 & 37.4 & 31.7 & -- & -- \\
IN22-Conv ta$\rightarrow$en& \textbf{46.0} & 43.4 & 37.9 & 16.5 & -- & -- \\
IN22-Gen en$\rightarrow$ta & \textbf{48.8} & 44.0 & 45.9 & 38.5 & -- & -- \\
IN22-Gen ta$\rightarrow$en & \textbf{54.3} & 50.1 & 42.7 & 19.8 & -- & -- \\
\midrule
MILU-ta (5-shot)           & \textbf{29.6} & 28.3 & 28.8 & 26.4 & -- & -- \\
Belebele-ta                & 27.8 & 26.6 & \textbf{58.3} & 23.9 & 32.6 & 30.2 \\
INCLUDE-ta                 & 26.9 & 29.8 & \textbf{35.5} & 26.4 & 27.1 & 26.5 \\
MMLU-ta                    & 25.5 & 25.7 & \textbf{32.6} & 24.2 & 29.1 & 28.1 \\
XCOPA-ta                   & \textbf{62.0} & 61.2 & \textbf{62.0} & 59.6 & 54.2 & 56.0 \\
MultiBLiMP-ta              & 99.0 & 98.4 & 97.9 & \textbf{99.2} & 93.2 & 93.2 \\
\bottomrule
\end{tabular}
\end{table}

\paragraph{Released-model battery.} The released model's benchmarks were run with the same commands as the round-4
battery, with equal sample counts. The MILU paper's Sarvam-1 score uses log-likelihood scoring, as ours does. An
earlier generation of the round-4 checkpoint's tool replies scored 23 and 13; models are compared only within one
generation run.

\paragraph{Everyday-chat replies.} Usability also requires the right script, a complete answer and a reply that stops.
Asked how many slippers make a pair, the round-4 checkpoint answers 12 and the released model 2.

\paragraph{Generation versus judge variation.} A new generation of the round-4 checkpoint's first-set replies scored 46
and 47 in the final selection pool, against 57, 46, 49 and 52 for the fixed reply file (Table~\ref{tab:transfer});
because the replies differ, that comparison mixes generation differences with judge variation. The epoch-3 replies of
the transfer pool were generated with the same generator and settings as the others.

\begin{table}[!htbp]
\centering
\small
\caption{Judged-usable counts by pool. Each row block is one blind pool; counts are comparable only within a block.
Everyday-chat set: 50 prompts; first conversational set: 100 prompts.}
\label{tab:transfer}
\begin{tabular}{llrr}
\toprule
Pool & Checkpoint & Usable per pass & Mean \\
\midrule
\multirow{3}{*}{\makecell[l]{Everyday chat,\\targeted-rounds pool}}
 & Initial SFT epoch 3        & 21 / 20 / 22 & 21.0 \\
 & Round-4 checkpoint         & 23 / 22 / 22 & 22.3 \\
 & Replay-only control        & 23 / 21 / 22 & 22.0 \\
\addlinespace
\multirow{2}{*}{\makecell[l]{Everyday chat,\\release pool}}
 & Round-4 checkpoint         & 23 / 21 / 24 & 22.7 \\
 & Released model (step 700)  & 26 / 25 / 25 & 25.3 \\
\addlinespace
\makecell[l]{First set, fixed replies}
 & Round-4 checkpoint         & 57 / 46 / 49 / 52 & 51.0 \\
\bottomrule
\end{tabular}
\end{table}

\paragraph{Reasoning adjudication.} The 40 items are arithmetic word problems, logic, ordering and sequences. Replies
whose answer could not be extracted automatically were adjudicated blind to model and condition, separately by two AI
adjudicators (Claude and Codex). The largest gain, 4 items, was step 1,029 with worked examples under one adjudication;
the round-4 checkpoint was run for context only. The diagnostic also does not show that other prompting methods would
fail.

\begin{table}[!htbp]
\centering
\small
\caption{Reasoning elicitation diagnostic: items correct out of 40, by prompting condition and adjudicator (Claude /
Codex where they differ). Gain and $p$: the best condition against direct answering, two-sided sign test on paired
items. The rule required a gain of at least 6 with $p<0.05$ for step 700 or step 1,029.}
\label{tab:reasoning}
\begin{tabular}{lrrrrr}
\toprule
Checkpoint & Direct & Step by step & Worked examples & Best gain & $p$ \\
\midrule
\multicolumn{6}{l}{\footnotesize Where two values are given: Claude adjudication / Codex adjudication.} \\
\midrule
Step 700            & 14 & 16 / 17 & 17      & $+3$     & 0.55--0.63 \\
Step 1,029          & 14 & 14      & 17 / 18 & $+3$ / $+4$ & 0.63 / 0.48 \\
Round-4 (context)   & 12 / 14 & 15 / 20 & 15 / 17 & $+3$ / $+6$ & 0.58--0.61 / 0.21 \\
\bottomrule
\end{tabular}
\end{table}

\section{Representation-analysis methods}\label{app:inside}
\label{app:repr}

This appendix gives the methods and remaining numbers behind Section~\ref{sec:repr}. The \agaram{} checkpoint is the
round-4 checkpoint unless stated; where the 2.5B base or pretraining checkpoints were also measured, this is said. The
comparison models are three of the five Gemma 4 sizes, the instruction-tuned E2B, E4B and 31B
\citep{gemmateam2026gemma4}. The approach follows an essay that argues for reading a model's own neurons directly,
through the inputs that most strongly activate them \citep{goodfire2026clean}; we add linear probes as a second tool. All results are single runs.

\paragraph{Word-identity probes.} Inputs: for each model, 300 Tamil words that take at least two tokens in that model's
tokenizer, sampled from the words whose last token is shared with at least one other candidate word; each word appears
in 24 contexts, 16 used for training and 8 held out. Because sharing was checked before the 300 were sampled, a
selected word's partner can be left out, so within the selected set only 178 of \agaram{}'s 300 Tamil words and 144 of
Gemma's share their last token with another selected word, and the last token alone identifies the rest. Measurement:
at every layer, a linear probe on the hidden state at the word's last token predicts which of the 300 words it is; we
report absolute accuracy on the held-out contexts. Control: the same model with random weights. Result: accuracy at the
embedding layer, where only the last token is available, was 0.63 for \agaram{} and 0.69 for the Gemma models, equal
to the best accuracy that the last token alone allows on these words (0.63 and 0.69, computed from the token ids). It exceeded 0.99 by layer 2 in \agaram{}
and 0.98 by layer 1 in every Gemma model; with random weights the probe peaked at 0.71 (\agaram{}) and 0.80 (Gemma E2B).
Tanglish and English words behaved the same way. Limitation: every trained model reaches the ceiling within two layers,
so the probe does not distinguish the models.

\paragraph{Cross-language sentence retrieval.} Inputs: the 997 Tamil--English pairs of the FLORES-200 dev set
\citep{nllb2022nollanguage}. Measurement: at every layer, each Tamil sentence's mean hidden state is compared by cosine
similarity, after standardising each dimension, with the mean hidden states of all English sentences; retrieval
succeeds when the closest is its own translation. Without the standardisation, a few very large dimensions dominate the
cosine and give misleading differences between models. Control: random weights. Result: every trained model, \agaram{}
at four pretraining checkpoints, the 2.5B base and the round-4 checkpoint, and all three Gemma models, retrieved the
right translation for 98.6--100\% of sentences at its best layer (99.5--100\% for the final checkpoints), against 9\%
with random weights. Limitation: all trained models reach the ceiling of this test.

\paragraph{Activation-based neuron classification.} Inputs: a fixed corpus of 3,000 Tamil-script, 1,000 Tanglish and
2,000 English passages. Measurement: for every MLP neuron of the decoder layers (vision and audio modules of the Gemma
models are excluded), the 20 tokens on which it activated most strongly. A neuron is classified by a language when at
least 16 of its 20 come from passages in that language. Neurons whose strongest activation is under 5\% of their
layer's median are left out as inactive. Control: Gemma 4 E2B with random weights. A mixed class (at least three
Tamil-script and three Tanglish activations, at most three English) occurs at the random-weights level and carries no
information, so we do not report it. Context for the token shares: Gemma's tokenizer needs 2.29 tokens per Tamil word
against \agaram{}'s 1.55 on the same held-out text, which is why Tamil script is a larger share of Gemma's corpus
tokens. Limitation: the classification depends on the corpus, the top-20 window and both thresholds; it was run once,
and its per-parameter ratios use total checkpoint parameters (Section~\ref{sec:repr-neurons}).

\paragraph{Blind annotation of neuron contexts.} Inputs: 120 active neurons per model, 40 from each third of the layers;
for each neuron, the 12 contexts with the strongest activations and the six with the most negative activations.
Measurement: eight Claude labelling agents, given items mixed across models and without model names, named each neuron's
function and a category, or marked it unclear. Control: each model also contributed 24 shuffled controls whose contexts
came from different random neurons, so they have no single function; the agents found a function in none of the 96.
Result: Table~\ref{tab:labels}. Word-piece neurons (one syllable inside many words, or a piece shared across scripts,
such as ``Fin'' in English names and ``\ta{லிமிடெட்}'') are more common in \agaram{}, 40\% of its early-layer sample.
Gemma's cross-language neurons are shared topics, for example news of police violence in Tamil and English.
Limitation: the agents read English more reliably than Tamil, which may inflate \agaram{}'s unclear share, and smaller
models may be expected to have less specific neurons.

\begin{table}[!htbp]
\centering
\small
\caption{Blind labels of 120 random active neurons per model (shares of the 120). ``One function'' counts neurons the
labeller judged to have one clear function, across categories. Shuffled controls: 24 per model. \agaram{}: the round-4
checkpoint.}
\label{tab:labels}
\footnotesize
\setlength{\tabcolsep}{4pt}
\begin{tabular}{lrrrrrrrr}
\toprule
Model & \makecell[r]{One\\function} & \makecell[r]{Tamil\\form} & \makecell[r]{Tamil\\concept} & \makecell[r]{Cross-\\language} &
\makecell[r]{Word\\piece} & \makecell[r]{English\\only} & Format & Unclear \\
\midrule
\agaram{} 2.5B & 30\% & 7\% & 17\% & 10\% & 21\% & 4\% & 4\% & 36\% \\
Gemma 4 E2B-it & 46\% & 7\% & 3\% & 18\% & 10\% & 18\% & 17\% & 26\% \\
Gemma 4 E4B-it & 46\% & 3\% & 3\% & 17\% & 18\% & 16\% & 16\% & 27\% \\
Gemma 4 31B-it & 45\% & 3\% & 3\% & 24\% & 11\% & 20\% & 7\% & 32\% \\
Shuffled controls & 0\% & 0\% & 0\% & 0\% & 0\% & 0\% & 0\% & 100\% \\
\bottomrule
\end{tabular}
\end{table}

\paragraph{Recognition and free-response comparison.} Inputs: 69 short factual questions: Tamil and Tanglish everyday
facts, items from our everyday-chat set and the preview, rare facts and controls. Measurement: the round-4 checkpoint
answered each in chat, and the replies were marked right or wrong; the same facts were scored as four-option multiple
choice on raw text, ranking the options by log-probability, for both the 2.5B base and the round-4 checkpoint, over
several probes per fact. Result: of the 26 chat answers marked wrong, 14 concern facts the base model ranked correctly
in most probes. For 10 of those 14, the round-4 checkpoint ranked the right option first in every probe; for the other
four, in 50--83\% of probes. Some replies start right and then contradict themselves (ice left in the sun melts, then
``slowly freezes into ice''); others read an everyday question as a digital one (for a pencil mistake, the model
suggests ``Undo'' or ``Backspace'' instead of an eraser). Limitation: ranking a supplied option is weaker than
producing the answer, and the facts were not sampled to represent any population.

\paragraph{Tool-use direction and steering procedure.} Inputs: 62 human-written prompts for fitting (13 needing a
tool) and 60 held-out prompts (12 needing a tool), each given the tool list the preview offered; and the first messages
of the preview. Measurement: the direction is the difference between the mean residual states after decoder layer 14
(the input to layer 15) of prompts that need a tool and prompts that do not, at the last prompt token. It separated the
held-out prompts with AUC 0.997 at layer 14 and 0.998 at layer 15. During generation we subtracted a multiple of the
direction from the residual stream at every position. Control: a random direction of the same length. Results are in
Table~\ref{tab:steer}. Limitation: the direction was fitted on single-turn prompts. Evaluation included first preview
messages and a separate 201-turn continuation check; the latter showed no improvement. Each judged result uses one pass
of Gemma-4-31B.

\begin{table}[!htbp]
\centering
\small
\caption{Tool-direction steering on the round-4 checkpoint. ``Unneeded'': tool calls on prompts that needed none.
Needed calls were kept in every row shown.}
\label{tab:steer}
\begin{tabular}{L{6.4cm}L{6.2cm}}
\toprule
Setting & Result \\
\midrule
Held-out prompts (48 not needing a tool, 12 needing one) & unneeded calls 42 without steering; 33, 26 and 19 at 0.5, 0.75 and 1.0 times the direction (4 replies at 1.0 ran to the token cap); random direction at 0.75: 43; all 12 needed calls kept \\
Preview first messages, no filter (209 not needing a tool, 19 needing one) & at 0.75: unneeded calls 176 to 126; usable replies 29 to 72 (44 better, 1 worse) \\
Preview first messages, behind the preview's hand-written tool filter (159) & usable replies 91 to 93 \\
Later turns of preview conversations (201) & no tool calls with or without steering; usable replies 89 without, 86 with \\
\bottomrule
\end{tabular}
\end{table}

\section{Release validation}\label{app:release}

This appendix records the checks run on the release files of the released model (SFT v2 step 700). The checks on the
base model's GGUF files were run earlier, when those files were built.

\paragraph{Files.} The chat GGUF files were converted from the released model's safetensors to a 16-bit file and
quantised from it to \texttt{Q8\_0} (8.50 bits per weight) and \texttt{Q4\_K\_M} (4.93 bits per weight). The 16-bit file
records 28 blocks, hidden size 2,304, feed-forward size 9,216, 36 attention heads with 36 key--value heads, a 64,000-token
vocabulary, context length 8,192 and rotary base 500,000. Each file's SHA-256 checksum is listed in the GGUF repository.

\paragraph{Pre-tokenizer.} llama.cpp selects its pre-tokenizer by name and has no entry for ours. The files declare
\texttt{qwen35}, the closest existing entry, whose splitting rule differs from ours only in leaving out the zero-width
non-joiner and joiner (U+200C, U+200D). Text containing these characters may therefore be tokenized differently by
llama.cpp than by the Hugging Face tokenizer. None of the parity texts below contains them, so that difference is
untested.

\paragraph{Token parity.} We compared llama.cpp's token ids with the Hugging Face tokenizer's on 961 texts: 661 raw
texts from our three conversational evaluation sets and 300 held-out training conversations rendered in the chat
format. The ids matched on all 961 for the 16-bit and \texttt{Q4\_K\_M} files.

\paragraph{Greedy replies.} For the first eight prompts of the everyday-chat set, we compared greedy replies from each
file with the Hugging Face model's, character by character (Table~\ref{tab:gguf}). Exact matches were 5 of 8 for the
16-bit file, 5 of 8 for \texttt{Q8\_0} and 0 of 8 for \texttt{Q4\_K\_M}. Most differing replies share a prefix and then
diverge in wording. One \texttt{Q4\_K\_M} reply changes a fact: it says a pair of slippers is 12, where the
full-precision model says 2. We therefore recommend the 8-bit file or larger. The files need llama.cpp build b11574 or
later.

\begin{table}[!htbp]
\centering
\small
\caption{Release validation for the released model's GGUF files. Greedy replies: exact character-level matches with
the Hugging Face model on 8 everyday-chat prompts.}
\label{tab:gguf}
\begin{tabular}{lrrr}
\toprule
File & Token parity (of 961) & Greedy exact (of 8) & Bits per weight \\
\midrule
16-bit                   & 961 & 5 & 16.00 \\
\texttt{Q8\_0}           & --  & 5 & 8.50 \\
\texttt{Q4\_K\_M}        & 961 & 0 & 4.93 \\
\bottomrule
\end{tabular}
\end{table}

\paragraph{Availability.} The release contents and addresses are listed in Table~\ref{tab:availability}.

\paragraph{Compute and cost details.}\label{app:release-full} Table~\ref{tab:compute} lists every training job. The
pretraining job and an abandoned SFT attempt were invoiced together; the costs of the later Lambda runs are list-price
estimates from recorded machine time, not invoiced amounts. Of the Lambda figure in Section~\ref{sec:cost}, about
USD 335 is that estimate for the three-epoch run and SFT v2; of own funds, about USD 300 is judge use since 12 October
2026.

\begin{table}[!htbp]
\centering
\small
\caption{Compute for \agaram{} 2.5B. The pretraining and abandoned H100 PCIe SFT jobs were invoiced together at
USD 4,776.50; their split is at list price. Costs marked $^a$ are list-price estimates from recorded machine time.}
\label{tab:compute}
\begin{tabular}{L{3.1cm}L{4.4cm}L{2.2cm}r}
\toprule
Job & Hardware & Hours & Cost (USD) \\
\midrule
Pretraining, both stages & 8$\times$H100 SXM5, Lambda & $\approx$149 (147.7 training) & $\approx$4,765 \\
Abandoned SFT attempt & 1$\times$H100 PCIe, Lambda (thermally throttled) & 3.7 & $\approx$12 \\
Initial SFT & 1$\times$H200, JarvisLabs & 9.5 training & JarvisLabs credits \\
SFT rounds 2--4 & one DGX Spark (owned) & $\approx$12, 4 and 2 & -- \\
Three-epoch run & 1$\times$H100 SXM5, Lambda & $\approx$15.5 & $\approx$66$^a$ \\
SFT v2 pilots & 1$\times$H100 SXM5 each, up to four at once, Lambda & $\approx$51 machine-hours & $\approx$220$^a$ \\
SFT v2 final run & 1$\times$H100 SXM5, Lambda & 11.3 (9.0 training) & $\approx$48$^a$ \\
Teacher generation, evaluation & two DGX Spark machines (owned) & -- & -- \\
LLM judge & Kimi K3 through the Fireworks API & -- & own funds \\
\bottomrule
\end{tabular}
\par\smallskip{\footnotesize $^a$Machine time from launch to termination at the list price of USD 4.29 per hour, not an
invoiced amount. Where a driver record lacks a termination time we used the recorded termination check or boot-guard
deadline.}
\end{table}

\paragraph{Frontier-model text in the SFT data.} Table~\ref{tab:provenance} gives the counts behind
Section~\ref{sec:disclose}. The Codex-written user turns replaced a generator for Tamil user turns that had failed;
Gemma-4-31B wrote the replies, and none of the 317 assistant turns that Codex later edited is in the SFT data. The 37
Claude-edited lines of the identity pairs were rewritten when the pairs were updated for the new model's facts; the
pairs state the model's name, its makers and that it was trained from scratch.

\begin{table}[!htbp]
\centering
\small
\caption{Frontier-model text in the SFT data, by checkpoint. ``Loss on the final turn'': only the final assistant turn,
written by the open-weight teacher, is supervised.}
\label{tab:provenance}
\begin{tabular}{L{5.2cm}L{3.6cm}L{3.6cm}}
\toprule
Material & Round-4 checkpoint & Released model \\
\midrule
Conversations with Codex-written user turns & 1,056 in initial SFT; rounds replayed 99, 31 and 12 & 1,043, loss on the final turn \\
Identity pairs with Claude-edited lines (37 lines of 122 pairs) & included, 976 rows (pairs repeated eight times) & none \\
Other identity data & targeted rounds 2--4 & 5,160 records from the targeted rounds and a later identity set \\
\bottomrule
\end{tabular}
\end{table}

\paragraph{Public datasets produced by frontier models.} Public datasets were admitted by licence, whoever produced
them. The English answers behind the Tamil Alpaca translation were generated by OpenAI's text-davinci-003 (51,699 rows
in the decay-stage instruction pool); Tamil Orca mixes translations and responses generated by GPT-4 and Gemini (93,932
rows). Together they are about 15\% of the pool's 945,781 documents, which the decay stage trained on once
(Section~\ref{sec:data-decay}).

\paragraph{Correspondence with the OCR providers.} The books themselves are not the OCR companies' content. We wrote to
Bodhan AI and to Sarvam AI on 10 October 2026 and contacted Sarvam's support channel again. Sarvam replied on 13 October
without granting permission; Bodhan AI has not replied.

\paragraph{Release contents and what can be reproduced.} Besides the weights and tokenizers, the release holds the GGUF
files with their conversion script and parity check, the chat script and the model cards. Also not released, beyond the
items in Section~\ref{sec:repro}: the tokenizer code, the source manifests and filter rules, the evaluation-harness
configuration and the tool schemas; the NVIDIA datasets may be used for training but not redistributed. The GGUF parity
checks and the model's outputs can be reproduced from the release. The corpus and training runs are described, not
rebuildable, through the architecture and optimiser (Section~\ref{sec:pretrain}), the tokenizer sample and
pre-tokenizer (Section~\ref{sec:data-tok}), the sources and their shares (Sections~\ref{sec:data-sources}
and~\ref{sec:data-decay}), the SFT mixture (Section~\ref{sec:sft-v2}) and the scan method
(Section~\ref{sec:eval-overlap}).

\end{document}